%% file: main.tex
\documentclass{article}
\usepackage[english]{babel}
\usepackage{microtype}
\usepackage{graphicx}
\usepackage{booktabs}
\usepackage{hyperref}
\usepackage[preprint]{icml2026}
\usepackage{etoolbox}
\makeatletter
\renewcommand{\ICML@preprint}{\textit{Preprint. 2027.}}
\patchcmd{\printAffiliationsAndNotice}{Correspondence to: \icmlcorrespondingauthor@text}{Email: \authoremails}{}{\PackageError{main}{footnote patch failed}{}}
\makeatother
\newcommand{\authoremails}{Yuhang He \textless{}yuhanghe@stu.pku.edu.cn\textgreater{}, Si Wu \textless{}siwu@pku.edu.cn\textgreater{}}
\usepackage{amsmath,amssymb,amsthm,mathtools}
\usepackage{array,enumitem,placeins,url,xcolor}
\usepackage{xurl}
\usepackage{etoc,tabularx}
\input{macros.tex}
\input{numbers.tex}
\input{results_tables.tex}
\hypersetup{colorlinks=true,linkcolor=mydarkblue,citecolor=mydarkblue,urlcolor=mydarkblue,
pdftitle={How Finite Brains Generalize Without Bound: Spatial Memory as a Universal Computer},pdfauthor={Yuhang He and Si Wu},pdfsubject={}}
\icmltitlerunning{How Finite Brains Generalize Without Bound}

\begin{document}
\twocolumn[
\icmltitle{How Finite Brains Generalize Without Bound:\\Spatial Memory as a Universal Computer}
\icmlsetsymbol{corr}{*}
\begin{icmlauthorlist}
\icmlauthor{Yuhang He}{pku1,pku2,pku3,pku4,pku5}
\icmlauthor{Si Wu}{pku1,pku2,pku3,pku4,pku5,corr}
\end{icmlauthorlist}
\icmlaffiliation{pku1}{School of Psychological and Cognitive Sciences, Peking University, Beijing, China}
\icmlaffiliation{pku2}{IDG/McGovern Institute for Brain Research, Peking University, Beijing, China}
\icmlaffiliation{pku3}{Peking-Tsinghua Center for Life Sciences, Academy for Advanced Interdisciplinary Studies, Peking University, Beijing, China}
\icmlaffiliation{pku4}{Center of Quantitative Biology, Peking University, Beijing, China}
\icmlaffiliation{pku5}{Key Laboratory of Machine Perception (Ministry of Education), Peking University, Beijing, China}
\icmlcorrespondingauthor{Si Wu}{siwu@pku.edu.cn}
\icmlkeywords{length generalization, recursion, grid cells, hippocampus, place cells, Chinese remainder theorem, rule following, theoretical neuroscience}
\vskip 0.3in
]
\printAffiliationsAndNotice{\textsuperscript{*}Corresponding author }
\hypersetup{pdftitle={How Finite Brains Generalize Without Bound: Spatial Memory as a Universal Computer}}

\begin{abstract}
\input{sec_abstract.tex}
\end{abstract}

\begin{figure*}[t]
\centering
\includegraphics[width=0.86\textwidth]{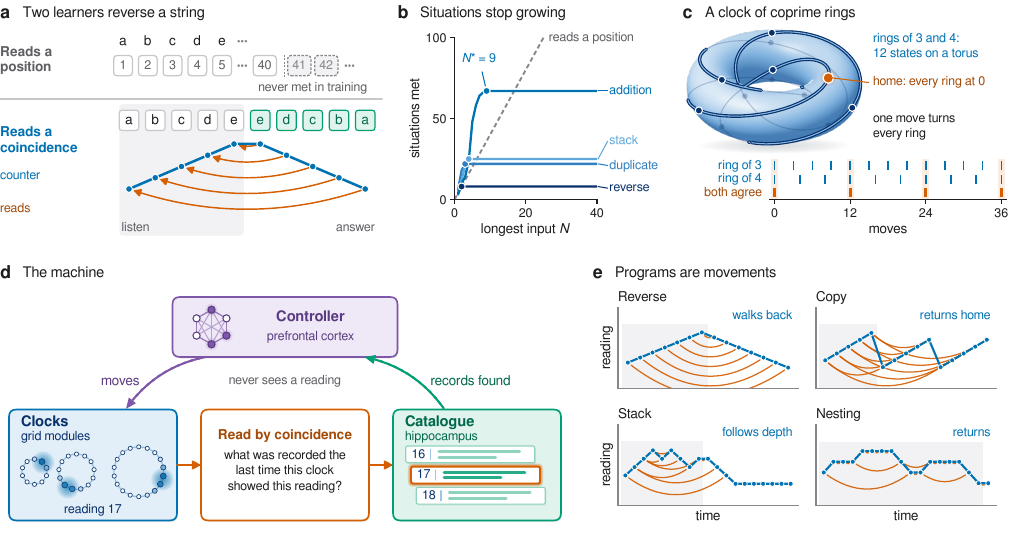}
\caption{\textbf{A learner that reads coincidences.} \textbf{(a)} Two learners reverse a string: one files each symbol under its position, and every position beyond its training is new to it; the other moves a counter up while it listens and down while it answers, and reads what it heard when the counter last showed its present value (arcs). \textbf{(b)} Situations met on inputs up to length $N$: new ones at every length for a learner that reads its position, none beyond a small $N^\ast$ for a procedure that reads coincidences. \textbf{(c)} A clock of coprime rings: one move turns every ring, rings of $3$ and $4$ form a cycle of $12$ states on a torus, and a cell tuned to one state of both fires where they agree. \textbf{(d)} The machine of Eq.~\eqref{eq:machine} and its parts in the brain. \textbf{(e)} Programs are movements of clocks (dots: readings; arcs: reads): a clock that walks back reverses, a clock sent \home{} copies, a clock that follows depth is a stack, and a clock that returns handles nesting.}
\label{fig:concept}
\end{figure*}

\etocdepthtag.toc{mtmain}
\input{sec_intro.tex}
\input{sec_theory.tex}

\begin{figure*}[t]
\centering
\includegraphics[width=0.86\textwidth]{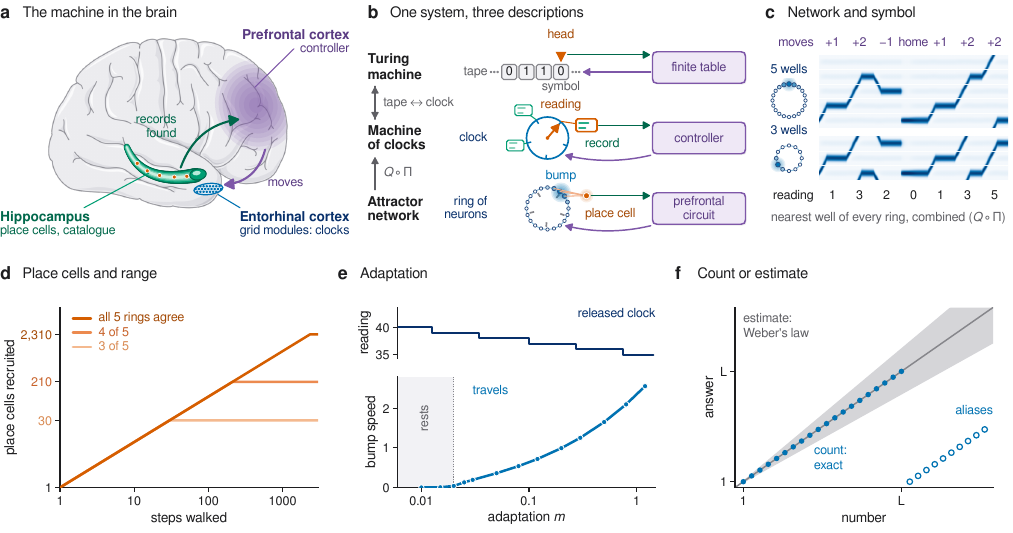}
\caption{\textbf{One system, three descriptions.} \textbf{(a)} The parts of the machine in the brain. \textbf{(b)} The same computation described three times: a Turing machine's tape is a clock whose cells are records and its head the reading (Theorem~\ref{thm:tapes}); a clock is a ring of neurons whose bump stands in one of $p$ wells, read by $\Pi$ and $Q$, and a place cell fires where the bump returns (Theorem~\ref{thm:neural}); the finite table, the controller and the prefrontal circuit are one rule. \textbf{(c)} Two rings of Eq.~\eqref{eq:cann}, with $5$ and $3$ wells, under a sequence of moves, and the reading $Q\circ\Pi$ after each. \textbf{(d)} Place cells recruited as a clock of rings $2,3,5,7,11$ advances: recruitment stops at $2{,}310$ cells when all rings must agree, at $210$ or $30$ when one or two may differ. \textbf{(e)} A clock released after $40$ moves runs back through its last readings by itself (top); above $\tau/\tau_v$ adaptation makes a bump travel (bottom). \textbf{(f)} A count is exact up to its clock's range and then repeats; an estimate reads a graded magnitude. Brain outline adapted from Servier Medical Art, CC BY 3.0.}
\label{fig:brain}
\end{figure*}

\input{sec_computation.tex}

\input{sec_brain.tex}

\begin{figure*}[t]
\centering
\includegraphics[width=0.90\textwidth]{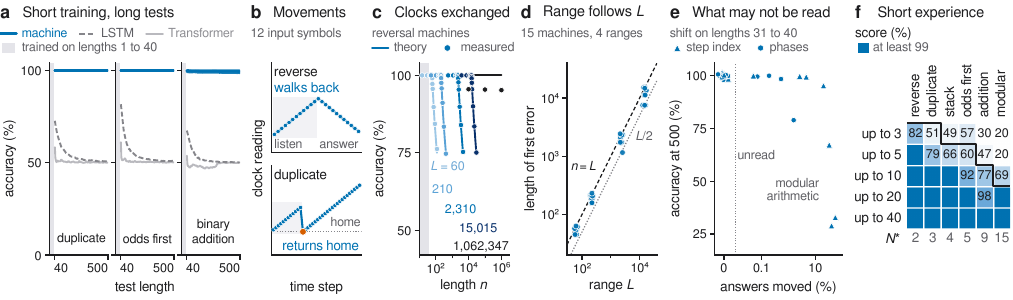}
\caption{\textbf{Rules learned on short inputs, tested at every length.} \textbf{(a)} Accuracy at every test length after training on lengths $1$ to $40$: the machine (three seeds), LSTMs and Transformers. \textbf{(b)} Clock readings of trained machines: in reversal a clock walks back, in duplication a clock returns \home{} and runs forward. \textbf{(c)} Reversal with clocks exchanged after training: dots are measured, lines the accuracy the theory predicts with no free parameter (Section~\ref{sec:range}); black, rings $11$ to $23$ on inputs of up to a million symbols. \textbf{(d)} The first error moves with the range $L$. \textbf{(e)} Leibniz's test: the share of answers on inputs of lengths $31$ to $40$ that change when the step index (triangles) or every reading (circles) is shifted by a common amount, against accuracy at length $500$, one point per run. \textbf{(f)} Score after training up to length $3$, $5$, $10$, $20$ or $40$, with the $N^\ast$ of each task; above the line, training was shorter than $N^\ast$.}
\label{fig:tests}
\end{figure*}

\input{sec_experiments.tex}

\input{sec_discussion.tex}

\section*{Impact Statement}
This paper presents work whose goal is to advance theoretical neuroscience and machine learning: a principle for how neural systems generalize learned procedures beyond the sizes they were trained on, tested with simulations and benchmarks of modest size. No human participants or animal experiments were involved. There are many potential societal consequences of our work, none of which we feel must be specifically highlighted here.

\section*{AI Use Statement}

\bibliography{references}
\bibliographystyle{icml2026}

\clearpage
\onecolumn
\appendix
\input{appendix.tex}
\end{document}

%% file: macros.tex
\definecolor{mydarkblue}{rgb}{0.1,0.2,0.55}

\newcommand{\Z}{\mathbb Z}

\newcommand{\home}{\textup{\textsc{home}}}
\theoremstyle{plain}
\newtheorem{theorem}{Theorem}
\newtheorem{proposition}{Proposition}

\newtheorem{lemma}{Lemma}
\theoremstyle{definition}
\newtheorem{definition}{Definition}
\newcommand{\setres}[2]{\expandafter\def\csname res@#1\endcsname{#2}}
\newcommand{\res}[1]{\ifcsname res@#1\endcsname\csname res@#1\endcsname\else\textcolor{gray}{[pending]}\fi}
\newcommand{\rest}[1]{\ifcsname res@#1\endcsname\csname res@#1\endcsname\else\textcolor{gray}{--}\fi}
\newcommand{\restable}[1]{\ifcsname restab@#1\endcsname\csname restab@#1\endcsname\else\begin{table}[ht]\centering\small\caption{\textcolor{gray}{Written by \texttt{code/make\_numbers.py} once the experiment has run.}}\label{#1}\end{table}\fi}
\newcommand{\resultfigure}[4]{\IfFileExists{figs/#1}{\includegraphics[width=#2]{figs/#1}}{\fbox{\parbox[c][\dimexpr#4-2\fboxsep-2\fboxrule\relax][c]{\dimexpr#2-2\fboxsep-2\fboxrule\relax}{\centering\small\textcolor{gray}{Figure produced by \texttt{\detokenize{#3}}}}}}}

%% file: numbers.tex
\setres{A.plain.att}{97}
\setres{A.plain.exact}{five}
\setres{A.plain.fit}{20}
\setres{A.plain.fitexact}{19}
\setres{A.plain.gap}{17}
\setres{A.plain.other}{modular arithmetic, seed 1, 98.5\%}
\setres{B.add.hi}{41}
\setres{B.add.lo}{27}
\setres{B.br.bare}{37, 83, 84}
\setres{B.br.full}{79, 90, 90}
\setres{B.past.hi}{53}
\setres{C.cap.fit}{14 of 15}
\setres{C.coord.mod.hi}{67}
\setres{C.coord.mod.lo}{29}
\setres{C.graded.fit}{all 15}
\setres{C.graded.hi}{71}
\setres{C.graded.lo}{48}
\setres{C.neural.mult.maxn}{12}
\setres{C.neural.tm}{all $40$}
\setres{C.think.acc0}{52.0}
\setres{C.think.maxn}{100}
\setres{C.think.steps}{5{,}716}
\setres{C.tm.beyond.k}{4}
\setres{C.tm.beyond.m}{623}
\setres{C.tm.n}{60}
\setres{C.tm.phrase}{all $1{,}200$}
\setres{C.tm.within}{all $2{,}977$}
\setres{D.pos}{46.6}
\setres{E0.C.binary_addition}{3}
\setres{E0.C.bucket_sort}{4}
\setres{E0.C.cycle_navigation}{0}
\setres{E0.C.duplicate_string}{1}
\setres{E0.C.even_pairs}{0}
\setres{E0.C.missing_duplicate_string}{3}
\setres{E0.C.modular_arithmetic}{0}
\setres{E0.C.modular_arithmetic_brackets}{1}
\setres{E0.C.odds_first}{2}
\setres{E0.C.parity_check}{0}
\setres{E0.C.reverse_string}{1}
\setres{E0.C.solve_equation}{1}
\setres{E0.C.stack_manipulation}{1}
\setres{E0.N.binary_addition}{9}
\setres{E0.N.bucket_sort}{6}
\setres{E0.N.cycle_navigation}{3}
\setres{E0.N.duplicate_string}{3}
\setres{E0.N.even_pairs}{3}
\setres{E0.N.missing_duplicate_string}{6}
\setres{E0.N.modular_arithmetic}{15}
\setres{E0.N.modular_arithmetic_brackets}{29}
\setres{E0.N.odds_first}{5}
\setres{E0.N.parity_check}{2}
\setres{E0.N.reverse_string}{2}
\setres{E0.N.solve_equation}{49}
\setres{E0.N.stack_manipulation}{4}
\setres{E0.R.binary_addition}{3}
\setres{E0.R.bucket_sort}{4}
\setres{E0.R.cycle_navigation}{0}
\setres{E0.R.duplicate_string}{2}
\setres{E0.R.even_pairs}{0}
\setres{E0.R.missing_duplicate_string}{2}
\setres{E0.R.modular_arithmetic}{0}
\setres{E0.R.modular_arithmetic_brackets}{1}
\setres{E0.R.odds_first}{2}
\setres{E0.R.parity_check}{0}
\setres{E0.R.reverse_string}{1}
\setres{E0.R.solve_equation}{1}
\setres{E0.R.stack_manipulation}{1}
\setres{E0.sit.binary_addition}{67}
\setres{E0.sit.bucket_sort}{221}
\setres{E0.sit.cycle_navigation}{20}
\setres{E0.sit.duplicate_string}{22}
\setres{E0.sit.even_pairs}{14}
\setres{E0.sit.missing_duplicate_string}{42}
\setres{E0.sit.modular_arithmetic}{505}
\setres{E0.sit.modular_arithmetic_brackets}{378}
\setres{E0.sit.odds_first}{22}
\setres{E0.sit.parity_check}{6}
\setres{E0.sit.reverse_string}{8}
\setres{E0.sit.solve_equation}{1003}
\setres{E0.sit.stack_manipulation}{25}
\setres{E1.binary_addition.max}{99.8}
\setres{E1.binary_addition.mean}{99.4$\pm$0.5}
\setres{E1.binary_multiplication.max}{54.3}
\setres{E1.binary_multiplication.mean}{53.9$\pm$0.4}
\setres{E1.bucket_sort.max}{70.8}
\setres{E1.bucket_sort.mean}{58.9$\pm$15.4}
\setres{E1.compute.base}{18}
\setres{E1.compute.jobh}{264}
\setres{E1.compute.jobs}{309}
\setres{E1.compute.median}{1.7}
\setres{E1.compute.nbase}{135}
\setres{E1.compute.total}{100}
\setres{E1.compute.wall}{13}
\setres{E1.compute_sqrt.max}{55.0}
\setres{E1.compute_sqrt.mean}{54.4$\pm$0.5}
\setres{E1.cycle_navigation.max}{100.0}
\setres{E1.cycle_navigation.mean}{89.5$\pm$18.1}
\setres{E1.drift.a}{95.6}
\setres{E1.drift.b}{50.1}
\setres{E1.drift.seed}{1}
\setres{E1.drift.task}{cycle navigation}
\setres{E1.duplicate_string.max}{100.0}
\setres{E1.duplicate_string.mean}{100.0$\pm$0.0}
\setres{E1.even_pairs.max}{100.0}
\setres{E1.even_pairs.mean}{100.0$\pm$0.0}
\setres{E1.fit.k}{27}
\setres{E1.fit.m}{30}
\setres{E1.fit.min}{96.9}
\setres{E1.lstm.k}{12}
\setres{E1.lstm.m}{19}
\setres{E1.missing_duplicate_string.max}{100.0}
\setres{E1.missing_duplicate_string.mean}{99.5$\pm$0.8}
\setres{E1.modular_arithmetic.max}{100.0}
\setres{E1.modular_arithmetic.mean}{99.9$\pm$0.1}
\setres{E1.modular_arithmetic_brackets.max}{28.6}
\setres{E1.modular_arithmetic_brackets.mean}{27.0$\pm$2.5}
\setres{E1.near.detail}{missing duplicate (seed 1) at 98.6\% and binary addition (seed 1) at 98.9\%}
\setres{E1.nexact}{nine}
\setres{E1.odds_first.max}{100.0}
\setres{E1.odds_first.mean}{100.0$\pm$0.0}
\setres{E1.parity_check.max}{100.0}
\setres{E1.parity_check.mean}{100.0$\pm$0.0}
\setres{E1.prev.binary_addition}{64.5}
\setres{E1.prev.duplicate_string}{75.1}
\setres{E1.prev.odds_first}{69.3}
\setres{E1.restarts.first}{23}
\setres{E1.restarts.later}{7}
\setres{E1.restarts.long}{8}
\setres{E1.restarts.n}{45}
\setres{E1.restarts.noisy}{6}
\setres{E1.restarts.none}{15}
\setres{E1.reverse_string.max}{100.0}
\setres{E1.reverse_string.mean}{100.0$\pm$0.0}
\setres{E1.solve_equation.max}{21.3}
\setres{E1.solve_equation.mean}{20.7$\pm$0.8}
\setres{E1.stack_manipulation.max}{100.0}
\setres{E1.stack_manipulation.mean}{100.0$\pm$0.0}
\setres{E1.tf.k}{none}
\setres{E1.tf.m}{19}
\setres{E2.abl.phrase}{all $810$}
\setres{E2.alias.phrase}{all $8{,}521$}
\setres{E2.big.dup.k}{two}
\setres{E2.big.dup.m}{three}
\setres{E2.big.rev.k}{three}
\setres{E2.big.rev.m}{three}
\setres{E2.ext.factor}{375}
\setres{E2.ext.revdup}{15{,}015}
\setres{E2.in.phrase}{all $2{,}562$}
\setres{E2.mega.L}{1{,}062{,}347}
\setres{E2.mega.k}{three}
\setres{E2.mega.t10.err}{346}
\setres{E2.mega.t10.k}{two}
\setres{E2.mega.t10.span}{3}
\setres{E2.never}{missing duplicate, seed 2}
\setres{E2.never.n}{30{,}030}
\setres{E2.never.span}{32}
\setres{E2.nmach}{15}
\setres{E2.ratio.hi}{1.10}
\setres{E2.ratio.lo}{0.50}
\setres{E2.rev.err}{0.27}
\setres{E2.rev.n}{81}
\setres{H.compute.jobs}{12}
\setres{H.compute.wall}{10.2}
\setres{H.exact.phrase}{one run of three for each $k$}
\setres{H.fail.fit}{none}
\setres{H.fail.maxval}{77}
\setres{H.fail.n}{fifteen}
\setres{H.other.hi}{95}
\setres{H.other.lo}{64}
\setres{K.c4.k}{one}
\setres{K.c4.n}{three}
\setres{K.mean.1}{33.7}
\setres{K.mean.2}{52.6}
\setres{K.mean.3}{58.9}
\setres{K.mean.4}{77.6}
\setres{K.val.1}{58}
\setres{K.val.2}{59}
\setres{K.val.3}{77}
\setres{L.clk.phrase}{at most 0.05\%}
\setres{L.mid.hi}{20}
\setres{L.mid.lo}{78.9}
\setres{L.mid.top}{99.9}
\setres{L.mod.hi}{55}
\setres{L.mod.lo}{33}
\setres{N.adapt.onset}{0.025}
\setres{N.adapt.threshold}{0.02}
\setres{N.broad.agree}{all $720$}
\setres{N.clock.firstrate}{3.0}
\setres{N.clock.firstring}{11}
\setres{N.clock.firstsig}{0.9}
\setres{N.clock.maxsig}{0.8}
\setres{N.clock.nowell0}{0.5}
\setres{N.clock.nowell08}{12.0}
\setres{N.clock.steps}{32{,}000}
\setres{N.compute.jobs}{148}
\setres{N.compute.wall}{0.5}
\setres{N.exact.ntasks}{ten}
\setres{N.exact.phrase}{all $1{,}920$}
\setres{N.noise06.agree}{all $720$}
\setres{N.noise08.agree}{99.9}
\setres{N.noise08.slips}{380}
\setres{N.nowells.hi}{86}
\setres{N.nowells.lo}{73}
\setres{N.nowells.n0}{41}
\setres{N.nowells.n1}{200}
\setres{N.partial1.below}{all $131$}
\setres{N.partial1.cells}{210}
\setres{N.partial2.below}{all $16$}
\setres{N.partial2.cells}{30}
\setres{N.release.dir}{back}
\setres{N.release.dwell}{91}
\setres{N.release.m}{0.3}
\setres{N.release.seq}{$40, 39, 38, \dots, 35$}
\setres{N.replay.dup}{1.00}
\setres{N.replay.rev}{$-$1.00}
\setres{Q.10.high.hi}{49}
\setres{Q.10.high.lo}{37}
\setres{Q.10.low.hi}{11}
\setres{Q.10.low.lo}{4}
\setres{Q.20.exact.lo}{99.2}
\setres{Q.40.exact.hi}{100.0}
\setres{Q.40.exact.lo}{70.7}
\setres{Q.bits}{100}
\setres{Q.n}{256}
\setres{T.10.add.hi}{83.2}
\setres{T.10.add.lo}{72.2}
\setres{T.10.odds.k}{two}
\setres{T.20.add.hi}{99.1}
\setres{T.20.add.lo}{97.5}
\setres{T.3.rev.hi}{98.2}
\setres{T.3.rev.lo}{50.7}
\setres{T.below.max}{71.5}
\setres{T.below.n}{21}
\setres{T.n}{72}
\setres{T.near.k}{five}
\setres{T.near.m}{18}

%% file: results_tables.tex
\expandafter\def\csname restab@tab:perseed\endcsname{%
\begin{table}[ht]\centering\small\caption{Benchmark runs of the machine: test score (\%, lengths 41--500) / best validation accuracy at training lengths (\%, lengths 31--40) / attempt kept.}\label{tab:perseed}
\begin{tabular}{@{}lrrr@{}}\toprule Task & seed 0 & seed 1 & seed 2\\\midrule
even pairs & 100.0 / 100.0 / 1 & 100.0 / 100.0 / 1 & 100.0 / 100.0 / 1\\
modular arithmetic & 99.8 / 100.0 / 1 & 99.8 / 100.0 / 1 & 100.0 / 100.0 / 1\\
parity & 100.0 / 100.0 / 1 & 100.0 / 100.0 / 1 & 100.0 / 100.0 / 1\\
cycle navigation & 99.9 / 100.0 / 1 & 68.6 / 100.0 / 1 & 100.0 / 100.0 / 2\\
stack manipulation & 100.0 / 100.0 / 1 & 100.0 / 100.0 / 1 & 100.0 / 100.0 / 2\\
reversal & 100.0 / 100.0 / 1 & 100.0 / 100.0 / 1 & 100.0 / 100.0 / 1\\
bracketed arithmetic & 24.1 / 84.0 / 5 & 28.6 / 91.8 / 1 & 28.4 / 93.0 / 3\\
equation solving & 20.9 / 56.2 / 5 & 19.8 / 28.5 / 1 & 21.3 / 64.8 / 5\\
duplication & 100.0 / 100.0 / 1 & 100.0 / 100.0 / 1 & 100.0 / 100.0 / 1\\
missing duplicate & 100.0 / 100.0 / 1 & 98.6 / 99.6 / 4 & 99.9 / 100.0 / 2\\
odds first & 100.0 / 100.0 / 1 & 100.0 / 100.0 / 1 & 100.0 / 100.0 / 1\\
binary addition & 99.8 / 100.0 / 5 & 98.9 / 99.2 / 2 & 99.5 / 99.9 / 3\\
binary multiplication & 53.7 / 70.6 / 1 & 54.3 / 72.5 / 5 & 53.6 / 69.8 / 1\\
square root & 54.0 / 71.3 / 5 & 54.1 / 72.9 / 5 & 55.0 / 75.9 / 1\\
bucket sort & 41.6 / 44.8 / 5 & 70.8 / 77.4 / 2 & 64.4 / 68.3 / 3\\
\bottomrule\end{tabular}\end{table}}
\expandafter\def\csname restab@tab:exchange\endcsname{%
\begin{table}[ht]\centering\small\caption{Clock exchange: the smallest sampled length with an output that differs from the unbounded clocks, for every benchmark machine that fits and extrapolates, with the cycles of $L=60$, $210$, $2{,}310$ and $15{,}015$ (--: no difference up to $2L$).}\label{tab:exchange}
\begin{tabular}{@{}lrrrrr@{}}\toprule Task & seed & $L=60$ & $L=210$ & $L=2{,}310$ & $L=15{,}015$\\\midrule
stack manipulation & 0 & 57 & 206 & 2{,}287 & 13{,}514\\
stack manipulation & 1 & 63 & 231 & 2{,}426 & 14{,}715\\
stack manipulation & 2 & 45 & 158 & 1{,}732 & 11{,}261\\
reversal & 0 & 61 & 212 & 2{,}333 & 15{,}165\\
reversal & 1 & 61 & 212 & 2{,}333 & 15{,}165\\
reversal & 2 & 61 & 212 & 2{,}333 & 15{,}165\\
duplication & 0 & 61 & 212 & 2{,}333 & 15{,}165\\
duplication & 1 & 45 & 158 & 1{,}732 & 7{,}508\\
duplication & 2 & 45 & 158 & 1{,}732 & 15{,}165\\
missing duplicate & 0 & 59 & 189 & 1{,}732 & 7{,}508\\
missing duplicate & 2 & -- & -- & -- & --\\
odds first & 0 & 57 & 158 & 1{,}732 & 7{,}508\\
odds first & 1 & 61 & 189 & 2{,}079 & 11{,}261\\
odds first & 2 & 54 & 158 & 1{,}732 & 7{,}508\\
binary addition & 0 & 45 & 158 & 1{,}732 & 11{,}261\\
binary addition & 2 & 54 & 158 & 1{,}155 & 7{,}508\\
\bottomrule\end{tabular}\end{table}}
\expandafter\def\csname restab@tab:plain\endcsname{%
\begin{table}[ht]\centering\small\caption{Plain relaxation (plain read, constant step size, no discrete phase, three attempts): test score (\%) / best validation accuracy at training lengths (\%).}\label{tab:plain}
\begin{tabular}{@{}lrrr@{}}\toprule Task & seed 0 & seed 1 & seed 2\\\midrule
even pairs & 89.9 / 89.5 & 100.0 / 100.0 & 100.0 / 100.0\\
modular arithmetic & 99.7 / 100.0 & 98.5 / 99.6 & 99.9 / 100.0\\
parity & 100.0 / 100.0 & 100.0 / 100.0 & 100.0 / 100.0\\
cycle navigation & 75.8 / 98.8 & 50.1 / 98.0 & 100.0 / 100.0\\
stack manipulation & 100.0 / 100.0 & 76.1 / 77.2 & 100.0 / 100.0\\
reversal & 100.0 / 100.0 & 100.0 / 100.0 & 100.0 / 100.0\\
bracketed arithmetic & 20.3 / 42.2 & 26.1 / 86.7 & 25.3 / 81.6\\
equation solving & 20.3 / 27.3 & 19.7 / 26.6 & 19.9 / 28.5\\
duplication & 100.0 / 100.0 & 100.0 / 100.0 & 100.0 / 100.0\\
missing duplicate & 51.5 / 69.1 & 56.4 / 79.7 & 74.6 / 82.0\\
odds first & 100.0 / 100.0 & 100.0 / 100.0 & 100.0 / 100.0\\
binary addition & 61.8 / 61.4 & 80.5 / 79.1 & 79.7 / 78.7\\
binary multiplication & 45.7 / 58.0 & 51.8 / 59.5 & 51.8 / 62.7\\
square root & 52.5 / 64.2 & 51.6 / 59.0 & 54.1 / 71.8\\
bucket sort & 34.3 / 44.3 & 51.0 / 58.7 & 47.3 / 50.5\\
\bottomrule\end{tabular}\end{table}}
\expandafter\def\csname restab@tab:teach\endcsname{%
\begin{table}[ht]\centering\footnotesize\setlength{\tabcolsep}{3pt}\caption{Training on short inputs: test score (\%, lengths 41--500) of machines trained on lengths $1$--$N$, seeds 0 / 1 / 2. $N^\ast$: size by which the reference procedure meets all its situations. $N=40$ is the benchmark.}\label{tab:teach}
\begin{tabular}{@{}lrrrrrr@{}}\toprule Task & $N^\ast$ & $N=3$ & $N=5$ & $N=10$ & $N=20$ & $N=40$\\\midrule
reversal & 2 & 98.2 / 97.3 / 50.7 & 100.0 / 100.0 / 100.0 & 100.0 / 100.0 / 100.0 & 100.0 / 100.0 / 100.0 & 100.0 / 100.0 / 100.0\\
duplication & 3 & 50.1 / 50.3 / 51.7 & 55.7 / 82.4 / 100.0 & 100.0 / 100.0 / 100.0 & 100.0 / 100.0 / 98.6 & 100.0 / 100.0 / 100.0\\
stack manipulation & 4 & 44.6 / 50.2 / 50.8 & 48.6 / 99.5 / 49.8 & 100.0 / 100.0 / 99.3 & 100.0 / 100.0 / 100.0 & 100.0 / 100.0 / 100.0\\
odds first & 5 & 59.0 / 61.1 / 50.7 & 50.8 / 70.6 / 59.5 & 75.5 / 100.0 / 100.0 & 100.0 / 100.0 / 100.0 & 100.0 / 100.0 / 100.0\\
binary addition & 9 & 25.6 / 28.1 / 37.2 & 44.1 / 47.9 / 48.8 & 83.2 / 72.2 / 76.5 & 99.1 / 97.5 / 98.6 & 99.8 / 98.9 / 99.5\\
modular arithmetic & 15 & 19.9 / 19.7 / 19.9 & 20.5 / 20.0 / 20.0 & 67.4 / 71.5 / 67.3 & 100.0 / 99.8 / 100.0 & 99.8 / 99.8 / 100.0\\
\bottomrule\end{tabular}\end{table}}
\expandafter\def\csname restab@tab:controls\endcsname{%
\begin{table}[ht]\centering\small\caption{Forbidden readings: accuracy (\%) on $256$ inputs of length $500$, seeds 0 / 1 / 2 (in parentheses: runs that do not fit their training lengths).}\label{tab:controls}
\begin{tabular}{@{}lrrrr@{}}\toprule Task & graded clocks & step index & phases & no noise\\\midrule
modular arithmetic & 100.0 / 99.2 / 100.0 & 32.8 / 67.2 / 28.9 & 78.9 / 98.4 / 99.2 & 100.0 / 99.2 / 99.2\\
stack manipulation & 59.1 / 61.7 / 57.6 & 100.0 / 100.0 / 100.0 & 100.0 / 100.0 / 100.0 & 100.0 / 99.9 / 100.0\\
reversal & 61.7 / 50.7 / 71.0 & 100.0 / 100.0 / 100.0 & 100.0 / 100.0 / 100.0 & 100.0 / 100.0 / 100.0\\
duplication & 50.1 / 53.4 / 50.2 & 100.0 / 100.0 / 100.0 & 100.0 / 99.8 / 100.0 & 100.0 / 100.0 / 100.0\\
binary addition & 49.0 / 49.4 / 48.2 & 95.3 / 99.7 / 99.5 & 100.0 / 99.6 / 98.4 & 99.4 / (50.2) / 99.5\\
\bottomrule\end{tabular}\end{table}}
\expandafter\def\csname restab@tab:shift\endcsname{%
\begin{table}[ht]\centering\small\caption{Leibniz's test on the runs given a coordinate: the share of answers (\%) on inputs of lengths $31$ to $40$ that change under the largest of four shifts of the step index or of every reading / accuracy (\%) on $256$ inputs of length $500$.}\label{tab:shift}
\begin{tabular}{@{}llrrr@{}}\toprule Task & given & seed 0 & seed 1 & seed 2\\\midrule
reversal & step index & 0.00 / 100.0 & 0.00 / 100.0 & 0.00 / 100.0\\
duplication & step index & 0.00 / 100.0 & 0.00 / 100.0 & 0.00 / 100.0\\
stack manipulation & step index & 0.05 / 100.0 & 0.00 / 100.0 & 0.00 / 100.0\\
binary addition & step index & 20.10 / 95.3 & 2.61 / 99.7 & 4.58 / 99.5\\
modular arithmetic & step index & 55.08 / 32.8 & 32.81 / 67.2 & 40.23 / 28.9\\
reversal & phases & 0.00 / 100.0 & 0.00 / 100.0 & 0.00 / 100.0\\
duplication & phases & 0.00 / 100.0 & 0.00 / 99.8 & 0.00 / 100.0\\
stack manipulation & phases & 0.00 / 100.0 & 0.00 / 100.0 & 0.00 / 100.0\\
binary addition & phases & 0.19 / 100.0 & 0.07 / 99.6 & 0.53 / 98.4\\
modular arithmetic & phases & 1.56 / 78.9 & 0.00 / 98.4 & 0.00 / 99.2\\
\bottomrule\end{tabular}\end{table}}
\expandafter\def\csname restab@tab:bare\endcsname{%
\begin{table}[ht]\centering\small\caption{What the machines read their past through: accuracy (\%) on inputs of the training lengths with the catalogue / with the catalogue removed, so that every head returns nothing.}\label{tab:bare}
\begin{tabular}{@{}lrrr@{}}\toprule Task & seed 0 & seed 1 & seed 2\\\midrule
even pairs & 100 / 51 & 100 / 51 & 100 / 51\\
modular arithmetic & 100 / 100 & 100 / 100 & 100 / 100\\
parity & 100 / 100 & 100 / 100 & 100 / 100\\
cycle navigation & 100 / 14 & 99 / 17 & 100 / 100\\
stack manipulation & 100 / 46 & 100 / 49 & 100 / 46\\
reversal & 100 / 50 & 100 / 53 & 100 / 51\\
bracketed arithmetic & 79 / 37 & 90 / 83 & 90 / 84\\
equation solving & 47 / 33 & 22 / 20 & 58 / 23\\
duplication & 100 / 50 & 100 / 50 & 100 / 50\\
missing duplicate & 100 / 100 & 99 / 46 & 100 / 54\\
odds first & 100 / 50 & 100 / 50 & 100 / 50\\
binary addition & 100 / 49 & 99 / 47 & 100 / 47\\
binary multiplication & 70 / 49 & 71 / 49 & 67 / 49\\
square root & 69 / 52 & 70 / 53 & 74 / 52\\
bucket sort & 45 / 21 & 77 / 27 & 69 / 22\\
binary addition, trained up to $10$ & 100 / 27 & 100 / 38 & 100 / 41\\
\bottomrule\end{tabular}\end{table}}
\expandafter\def\csname restab@tab:clocks\endcsname{%
\begin{table}[ht]\centering\small\caption{Bucket sort with $C$ clocks: test score (\%, lengths 41--497 for $C\neq3$, 41--500 for $C=3$) / best validation accuracy at training lengths (\%), seeds 0, 1, 2.}\label{tab:clocks}
\begin{tabular}{@{}crrr@{}}\toprule $C$ & seed 0 & seed 1 & seed 2\\\midrule
1 & 22.2 / 52.9 & 32.7 / 44.7 & 46.3 / 57.8\\
2 & 52.8 / 54.0 & 50.7 / 59.0 & 54.1 / 54.7\\
3 & 41.6 / 44.8 & 70.8 / 77.4 & 64.4 / 68.3\\
4 & 63.5 / 78.4 & 100.0 / 100.0 & 69.2 / 65.4\\
\bottomrule\end{tabular}\end{table}}
\expandafter\def\csname restab@tab:baselines\endcsname{%
\begin{table}[ht]\centering\small\caption{Networks trained under one protocol: test score (\%, lengths 41--500, mean over three seeds) and, in parentheses, the number of runs that fit their training lengths.}\label{tab:baselines}
\begin{tabular}{@{}lrrrr@{}}\toprule Task & LSTM & Transformer, rotary & Transformer, no positions & This work\\\midrule
even pairs & 100.0 (3) & 50.4 (3) & 50.2 (0) & 100.0 (3)\\
modular arithmetic & 99.7 (3) & 20.0 (0) & 20.2 (0) & 99.9 (3)\\
parity & 100.0 (3) & 49.9 (0) & 49.7 (0) & 100.0 (3)\\
cycle navigation & 50.2 (3) & 22.7 (3) & 24.1 (2) & 89.5 (3)\\
stack manipulation & 62.5 (0) & 49.0 (0) & 50.1 (0) & 100.0 (3)\\
reversal & 59.8 (0) & 50.8 (3) & 52.8 (0) & 100.0 (3)\\
bracketed arithmetic & 37.3 (1) & 19.9 (0) & 19.9 (0) & 27.0 (0)\\
equation solving & 28.4 (0) & 20.4 (0) & 20.4 (0) & 20.7 (0)\\
duplication & 54.0 (0) & 50.5 (0) & 52.8 (0) & 100.0 (3)\\
missing duplicate & 100.0 (3) & 52.2 (3) & 50.7 (0) & 99.5 (3)\\
odds first & 54.6 (0) & 50.7 (2) & 52.8 (0) & 100.0 (3)\\
binary addition & 52.4 (0) & 48.9 (0) & 45.3 (0) & 99.4 (3)\\
binary multiplication & 51.4 (0) & 49.8 (0) & 49.7 (0) & 53.9 (0)\\
square root & 54.4 (0) & 50.2 (0) & 50.2 (0) & 54.4 (0)\\
bucket sort & 90.6 (3) & 28.1 (3) & 23.7 (0) & 58.9 (0)\\
\bottomrule\end{tabular}\end{table}}
\expandafter\def\csname restab@tab:moves\endcsname{%
\begin{table}[ht]\centering\small\caption{How the clocks of trained machines move: for each clock (in order), the moves that make up at least $20\%$ of its moves while the input arrives, at the first answer step, and while the answer is given, most frequent first.}\label{tab:moves}
\begin{tabular}{@{}lllll@{}}\toprule Task & seed & input & first answer step & answer\\\midrule
stack manipulation & 0 & $+1$, $0$/\home, $+1$/$-1$ & $-1$, $-2$, $-1$ & $-1$, $-2$, $-1$\\
 & 1 & $0$/$+1$, $-1$/$+1$, $-1$/$-2$ & $0$, $+1$, \home & $-1$/$0$, $+1$, \home\\
 & 2 & $+1$/$-1$, $+1$/$-1$, \home/$0$/$+1$ & $+1$, $-1$, $+1$ & $+1$, $-1$, $+1$\\
reversal & 0 & $+1$, $+1$, $+1$ & $+1$, $+1$, $-1$ & $+1$, $-1$, $-1$\\
 & 1 & $+1$, $+1$, $+1$ & $-1$, $-1$, $+1$ & $-1$, $-1$, $-2$/$+1$\\
 & 2 & $0$/$-2$, $+1$, $+1$ & $-2$, \home, $-2$ & $-1$, \home, $-1$\\
duplication & 0 & $+1$, $+1$, $+1$ & \home, $+1$, $+1$ & $+1$, $+1$, $+1$\\
 & 1 & \home, $+1$, \home/$0$ & $+1$, \home, $+1$ & $+1$, \home/$+1$, $+1$\\
 & 2 & $-1$/$-2$, $+2$, $0$ & $+1$, $+2$, \home & $+1$, $+2$, $+2$\\
missing duplicate & 0 & $0$/$+2$, $-1$, $-2$/$+1$ & $0$, $+1$, $+1$ & $0$/$+1$, $+1$, $+1$\\
 & 1 & \home/$-1$/$+2$, \home/$+1$/$-2$, $+1$/$+2$/$-1$/$-2$ & $-1$, $+2$, $-2$/$0$ & $-1$, $+2$, $-2$\\
 & 2 & $0$/$+1$/$-1$, \home/$-1$, $-1$/\home & $-1$, \home, \home & $-1$, \home, \home\\
odds first & 0 & $+1$/\home, $+1$, $-1$ & \home, $+1$, \home & $-2$, $-2$, $-2$\\
 & 1 & $+1$, $+1$, \home/$0$ & $+2$, \home, $+1$ & $+2$, $+2$, $+2$\\
 & 2 & \home, $+1$, $+1$ & $+2$, $+1$, \home & $+2$, $+2$, $+2$\\
binary addition & 0 & $0$/$+2$, \home, $+1$ & \home, $+1$, $+1$ & $+1$, $+1$, $+1$\\
 & 1 & $+1$, $+1$, $0$ & $+2$, $+1$/$0$, $+1$ & $+2$, $-2$/$+2$, $+1$\\
 & 2 & $+1$, $+1$, \home/$+1$ & $+1$, $+1$, $+1$ & $+1$, $+2$/$+1$, $+1$\\
\bottomrule\end{tabular}\end{table}}
\expandafter\def\csname restab@tab:neural\endcsname{%
\begin{table}[ht]\centering\small\caption{Trained controllers on neural clocks and place cells (no noise): agreement of the output symbols with the symbolic machine (\%) / accuracy of the neural machine (\%), mean over the runs that fit, by length.}\label{tab:neural}
\begin{tabular}{@{}lrrrr@{}}\toprule Task & $n=41$ & $n=100$ & $n=200$ & $n=500$\\\midrule
even pairs & 100.0 / 100.0 & 100.0 / 100.0 & 100.0 / 100.0 & 100.0 / 100.0\\
modular arithmetic & 100.0 / 100.0 & 100.0 / 100.0 & 100.0 / 100.0 & 100.0 / 100.0\\
parity & 100.0 / 100.0 & 100.0 / 100.0 & 100.0 / 100.0 & 100.0 / 100.0\\
cycle navigation & 100.0 / 100.0 & 100.0 / 97.9 & 100.0 / 91.7 & 100.0 / 87.5\\
stack manipulation & 100.0 / 100.0 & 100.0 / 99.9 & 100.0 / 100.0 & 100.0 / 100.0\\
reversal & 100.0 / 100.0 & 100.0 / 100.0 & 100.0 / 100.0 & 100.0 / 100.0\\
duplication & 100.0 / 100.0 & 100.0 / 100.0 & 100.0 / 100.0 & 100.0 / 100.0\\
missing duplicate & 100.0 / 100.0 & 100.0 / 100.0 & 100.0 / 100.0 & 100.0 / 100.0\\
odds first & 100.0 / 100.0 & 100.0 / 100.0 & 100.0 / 100.0 & 100.0 / 100.0\\
binary addition & 100.0 / 99.8 & 100.0 / 99.5 & 100.0 / 99.4 & 100.0 / 99.4\\
\bottomrule\end{tabular}\end{table}}
\expandafter\def\csname restab@tab:neuralvar\endcsname{%
\begin{table}[ht]\centering\small\caption{Variants of the neural machine: accuracy (\%) by length, mean over three seeds. noise: noise of strength $\sigma$ on every neuron; partial$j$: place cells that fire when all but $j$ rings agree; nowells: no wells, $\sigma=0.3$; broad: recurrent kernel twice as wide.}\label{tab:neuralvar}
\begin{tabular}{@{}lll@{}}\toprule Variant & Task & accuracy at each length\\\midrule
broad & stack manipulation & 41: 100.0, 100: 99.9, 200: 100.0\\
broad & reversal & 41: 100.0, 100: 100.0, 200: 100.0\\
broad & duplication & 41: 100.0, 100: 100.0, 200: 100.0\\
broad & odds first & 41: 100.0, 100: 100.0, 200: 100.0\\
broad & binary addition & 41: 99.8, 100: 99.5, 200: 99.4\\
noise0.6 & stack manipulation & 41: 100.0, 100: 99.9, 200: 100.0\\
noise0.6 & reversal & 41: 100.0, 100: 100.0, 200: 100.0\\
noise0.6 & duplication & 41: 100.0, 100: 100.0, 200: 100.0\\
noise0.6 & odds first & 41: 100.0, 100: 100.0, 200: 100.0\\
noise0.6 & binary addition & 41: 99.8, 100: 99.5, 200: 99.4\\
noise0.8 & stack manipulation & 41: 100.0, 100: 99.9, 200: 100.0\\
noise0.8 & reversal & 41: 100.0, 100: 100.0, 200: 100.0\\
noise0.8 & duplication & 41: 99.3, 100: 100.0, 200: 100.0\\
noise0.8 & odds first & 41: 99.7, 100: 100.0, 200: 100.0\\
noise0.8 & binary addition & 41: 99.8, 100: 99.5, 200: 99.4\\
nowells & stack manipulation & 41: 94.7, 100: 84.3, 200: 72.5\\
nowells & reversal & 41: 93.9, 100: 84.9, 200: 80.2\\
nowells & duplication & 41: 85.3, 100: 78.1, 200: 70.9\\
nowells & odds first & 41: 61.6, 100: 55.2, 200: 52.6\\
nowells & binary addition & 41: 92.8, 100: 84.9, 200: 77.1\\
partial1 & stack manipulation & 100: 100.0, 180: 99.7, 200: 99.6, 240: 98.8, 300: 94.3\\
partial1 & reversal & 100: 100.0, 180: 100.0, 200: 100.0, 240: 93.9, 300: 85.3\\
partial1 & duplication & 100: 93.2, 180: 74.6, 200: 71.1, 240: 58.9, 300: 57.4\\
partial1 & odds first & 100: 100.0, 180: 91.0, 200: 89.1, 240: 79.5, 300: 67.9\\
partial1 & binary addition & 100: 99.6, 180: 99.4, 200: 99.1, 240: 93.3, 300: 80.9\\
partial2 & stack manipulation & 20: 99.7, 28: 98.0, 36: 95.6, 60: 80.4\\
partial2 & reversal & 20: 100.0, 28: 99.0, 36: 91.2, 60: 70.6\\
partial2 & duplication & 20: 81.1, 28: 73.5, 36: 63.9, 60: 58.3\\
partial2 & odds first & 20: 95.1, 28: 90.0, 36: 72.4, 60: 57.5\\
partial2 & binary addition & 20: 99.5, 28: 98.8, 36: 88.7, 60: 61.6\\
\bottomrule\end{tabular}\end{table}}
\expandafter\def\csname restab@tab:clockslips\endcsname{%
\begin{table}[ht]\centering\small\caption{Neural clocks alone: slips per step under random moves, by noise strength $\sigma$ ($16$ clocks of rings $2,3,5,7,11$, $2{,}000$ steps each).}\label{tab:clockslips}
\begin{tabular}{@{}lrrrrrrrrrrr@{}}\toprule Clocks & 0 & 0.3 & 0.4 & 0.5 & 0.6 & 0.7 & 0.8 & 0.9 & 1 & 1.1 & 1.2\\\midrule
wells & 0 & 0 & 0 & 0 & 0 & 0 & 0 & 3.0e-02 & 5.7e-01 & 8.3e-01 & 8.7e-01\\
no wells & 5.0e-03 & 3.7e-02 & 5.0e-02 & 6.3e-02 & 7.9e-02 & 1.0e-01 & 1.2e-01 & 1.4e-01 & 1.7e-01 & 2.4e-01 & 6.7e-01\\
\bottomrule\end{tabular}\end{table}}
\expandafter\def\csname restab@tab:adapt\endcsname{%
\begin{table}[ht]\centering\small\caption{Adaptation in one ring of $11$ wells ($\tau/\tau_v=0.02$). Left: speed (wells per $\tau$; the sign gives the direction) of a bump released with a trail of adaptation, by adaptation $m$ and well depth $\epsilon$ ($^\dagger$: the bump dies). Right: lead (wells; negative: lag) of a bump without wells that follows an input of strength $A$ moving at $v$ neurons per $\tau$, for $A, v$ as in the header.}\label{tab:adapt}
\begin{tabular}{@{}rrrrrrrr@{}}\toprule $m$ & $\epsilon=0$ & $\epsilon=0.5$ & $\epsilon=1$ & 0.3, 0.05 & 0.3, 0.1 & 1, 0.05 & 1, 0.1\\\midrule
0 & $0.0000$ & $0.0000$ & $0.0000$ & $-0.118$ & $-2.280$ & $-0.034$ & $-0.070$\\
0.01 & $0.0000$ & $0.0000$ & $0.0000$ & $-0.084$ & $-1.923$ & $-0.026$ & $-0.061$\\
0.015 & $0.0001$ & $0.0000$ & $0.0000$ & -- & -- & -- & --\\
0.02 & $0.0004$ & $0.0000$ & $0.0000$ & $-0.056$ & $-1.076$ & $-0.017$ & $-0.053$\\
0.025 & $0.0013$ & $0.0000$ & $0.0000$ & -- & -- & -- & --\\
0.03 & $0.0019$ & $0.0000$ & $0.0000$ & $-0.030$ & $-0.197$ & $-0.009$ & $-0.046$\\
0.05 & $0.0036$ & $0.0000$ & $0.0000$ & $+0.018$ & $-0.105$ & $+0.007$ & $-0.030$\\
0.08 & $0.0054$ & $0.0000$ & $0.0000$ & $+0.096$ & $-0.031$ & $+0.030$ & $-0.009$\\
0.12 & $0.0072$ & $0.0000$ & $0.0000$ & -- & -- & -- & --\\
0.2 & $0.0099$ & $0.0001$ & $0.0000$ & -- & -- & -- & --\\
0.3 & $0.0125$ & $-0.0088$ & $0.0001$ & -- & -- & -- & --\\
0.5 & $0.0165$ & $0.0151$ & $0.0168$$^\dagger$ & -- & -- & -- & --\\
0.8 & $0.0210$ & $0.0205$ & $-0.0001$$^\dagger$ & -- & -- & -- & --\\
1.2 & $0.0256$ & $0.0252$ & $0.0000$$^\dagger$ & -- & -- & -- & --\\
\bottomrule\end{tabular}\end{table}}
\expandafter\def\csname restab@tab:tmexact\endcsname{%
\begin{table}[ht]\centering\small\caption{Compiled random Turing machines ($20$ machines per number of tapes, $8$ states, $3$ symbols; inputs of lengths $10$ to $1{,}000$): runs that agree with direct simulation, with unbounded clocks and, with bounded clocks, among the inputs whose head range is below or at least $L$.}\label{tab:tmexact}
\begin{tabular}{@{}lrrrrrrr@{}}\toprule Tapes & unbounded & $L{=}60$, in & beyond & $L{=}210$, in & beyond & $L{=}2310$, in & beyond\\\midrule
1 & 400/400 & 284/284 & 2/116 & 358/358 & 1/42 & 400/400 & --\\
2 & 400/400 & 243/243 & 0/157 & 330/330 & 0/70 & 400/400 & --\\
3 & 400/400 & 241/241 & 0/159 & 321/321 & 1/79 & 400/400 & --\\
\bottomrule\end{tabular}\end{table}}
\expandafter\def\csname restab@tab:think\endcsname{%
\begin{table}[ht]\centering\small\caption{Schoolbook multiplication compiled from a three-tape Turing machine: accuracy (\%) on the benchmark inputs of length $n$ with $(n-1)^2/2+8n+16$ steps to think and with none.}\label{tab:think}
\begin{tabular}{@{}rrrr@{}}\toprule $n$ & steps to think & with time & without\\\midrule
5 & 64 & 100.0 & 20.6\\
10 & 136 & 100.0 & 40.2\\
20 & 356 & 100.0 & 47.5\\
40 & 1{,}096 & 100.0 & 46.7\\
64 & 2{,}512 & 100.0 & 52.0\\
100 & 5{,}716 & 100.0 & 51.1\\
\bottomrule\end{tabular}\end{table}}
\expandafter\def\csname restab@tab:hier\endcsname{%
\begin{table}[ht]\centering\small\caption{Bucket sort over $k$ kinds of symbol with $C$ clocks: test score (\%, lengths 41--500) / best validation accuracy (\%), seeds 0 / 1 / 2.}\label{tab:hier}
\begin{tabular}{@{}llrrr@{}}\toprule $k$ & $C$ & seed 0 & seed 1 & seed 2\\\midrule
3 & 1 & 64.6 / 67.7 & 70.2 / 71.1 & 61.5 / 69.5\\
3 & 2 & 80.8 / 87.3 & 95.1 / 87.1 & 100.0 / 100.0\\
4 & 2 & 55.6 / 77.4 & 51.5 / 48.9 & 53.8 / 57.1\\
4 & 3 & 65.3 / 73.7 & 100.0 / 100.0 & 92.7 / 89.5\\
\bottomrule\end{tabular}\end{table}}

%% file: sec_abstract.tex
A child who has learned to add small numbers adds numbers of any length. LSTMs and Transformers trained on short sequences fail on longer ones. We trace the difference to what a learner reads when it decides. A position, a count or a time keeps taking new values as the input grows: it is a coordinate, and a learner that reads one meets in every long input values its experience never covered. Whether two things are the same has the same two answers at every size: it is a coincidence, and a learner that reads only coincidences finds in a long input nothing that short inputs lacked, so finite experience fixes its rule for every size. The memory carries the size of the input and the rule stays finite, and this determines the memory a generalizing brain can have: a memory that grows while what the learner reads of it stays finite has to look the same from every place, so it is made of cycles, and the past is reached by asking what happened when the cycles last stood where they stand now. The cycles are the grid modules, the cells that detect their return the place cells, the records the hippocampal index; a prefrontal controller moves the cycles and never sees them. Reading only coincidences costs no power: what finite experience fixes in such a learner is what Turing machines compute in step with their input, which is why a circuit that evolved to find the way home can learn arithmetic and grammar. In a network of neurons this machine of cycles is the dynamics seen through its symmetry, and symbols are its invariants. Trained on strings of up to 40 symbols, the machine solves five benchmark tasks that no LSTM or Transformer solves, among them duplication and binary addition, and beyond the reach of its cycles it fails on the curve the theory predicts. Given larger cycles after training, a machine that has seen no string longer than five symbols reverses a string of a million without error. Its controllers run unchanged on neurons.

%% file: sec_intro.tex
\section{Introduction}
\label{sec:intro}

Language makes infinite use of finite means \citep{humboldt1836,chomsky1965aspects}. A listener who has heard only short sentences understands long ones, a child who has added small numbers adds numbers of any length, and monkeys and crows trained on short nested sequences carry the nesting to longer ones \citep{ferrigno2020recursive,liao2022recursive}. LSTMs and Transformers do not: trained on strings of up to 40 symbols they fail at lengths 41 to 500 on tasks as plain as duplicating a string \citep{deletang2023neural,ruoss2023randomized}, and a Transformer that reverses strings of 40 symbols without error is at chance by length 50, the hole of Theorem~\ref{thm:hole} in its plainest form (Appendix~\ref{app:bench}).

Hume asked why the future should resemble the past \citep{hume1748enquiry}; for a learner of rules the question is sharp, because every finite set of examples fits infinitely many rules. Kripke's \emph{quus} agrees with plus whenever both numbers are below 57 and answers 5 otherwise \citep{kripke1982wittgenstein}; Goodman's \emph{grue} agrees with green on everything examined before a certain date \citep{goodman1955fact}. A child who has added only numbers below 57 has no evidence against quus, and she goes on with plus. Something in her brain fixes the rule for numbers she has never met. This paper says what it is, derives from it the memory a generalizing brain has to have, and finds that memory in the hippocampal system.

What fixes the rule is what the learner reads when it decides. Two learners that reverse a string show the difference (Figure~\ref{fig:concept}a). The first stores the $k$th symbol under the label ``position $k$'' and, when it answers, looks up the position it needs; every label beyond 40 is new to it, and nothing in its training says what to do there. The second moves a counter one step up for each symbol it hears; for each symbol it says, it asks what it heard the last time the counter showed its present value, and then moves the counter one step down. Its counter grows with the input too, but the learner never looks at the value. We call such a counter a \emph{clock} and the value it shows its \emph{reading}. Everything a learner's next action depends on is its \emph{situation}. The second learner has eight (whether it is listening or answering, the symbol it hears, the symbol it finds), meets them all by length 2, and at length 500 meets no others (Figure~\ref{fig:concept}b). The memory carries the size of the input; the rule stays finite.

The first learner reads a \emph{coordinate}: a quantity, such as a position, a count or a time, that keeps taking new values as the input grows. The second reads a \emph{coincidence}: whether two things are the same, which has the same two answers at every size. Einstein met the distinction in the hole argument. Laws that keep their form under every change of coordinates seemed unable to fix the gravitational field inside a region, since relabelling the points inside turns one solution into another that agrees with it outside. The two solutions agree on every point coincidence, every meeting of two world lines; measurement records nothing else, so they describe one world \citep{einstein1916foundation,stachel1989einstein,norton1993general}. A coordinate is a name; a coincidence is an event. A learner that reads a coordinate has a hole of the same kind, the values beyond its experience. Einstein closed the hole by seeing that measurement records only coincidences; a learner closes its own by reading nothing else.

The principle determines the memory. The memory has to grow with the input, and the learner has to meet only finitely many situations in it; a memory that looks the same from every place does both. The learner moves through such a memory, and may read only what its moves leave unchanged near where it stands: whether two places are the same. A finite memory looks the same from every place only by closing on itself, so it is made of cycles, and a few cycles whose periods share no factor cover the most moments with the fewest neurons. These cycles are the modules of the entorhinal grid code \citep{hafting2005microstructure,fiete2008what}. A place cell fires where the modules that drive it agree, and so detects a coincidence \citep{okeefe1971hippocampus,solstad2006grid}; the hippocampus keeps an index of every moment \citep{teyler1986hippocampal}, filed here under the readings the cycles had; prefrontal cortex, gated by the basal ganglia \citep{oreilly2006making}, moves the cycles without seeing them. A program is a movement of the cycles, and a return from a call is the movement back; replay, in which place cells fire in sequence while the animal rests, is a clock walked over stored readings, a program running without input. Tolman's cognitive map \citep{tolman1948cognitive,okeefe1978hippocampus} is this memory read while the body moves the cycles; the theory derives it from generalization, and finding the way home is one of its uses.

Reading only coincidences costs no power. Turing's machine never reads where its head is: a step depends on the state and the symbol under the head \citep{turing1937computable}, and that is why one finite table serves tapes of every length. A tape is a clock whose cells are records, and what finite experience fixes in the machine of cycles is what multitape Turing machines compute in step with their input (Section~\ref{sec:computation}): the architecture of spatial memory is a universal computer. In attractor networks of neurons the machine is the dynamics seen through its symmetry, and symbols are what the symmetry leaves unchanged (Section~\ref{sec:brain}). In the experiments the memory is built from the theory and only the controller is learned: trained on strings of up to 40 symbols, it solves tasks that no LSTM or Transformer solves; trained on strings of at most five symbols and given larger cycles, it reverses a string of a million without error; and it runs unchanged on neurons (Section~\ref{sec:experiments}). Kant's form of intuition, given before experience, becomes the group of moves that turns the cycles (Section~\ref{sec:discussion}).

%% file: sec_theory.tex
\section{From a Principle to a Machine}
\label{sec:theory}

\subsection{What a learner reads}
\label{sec:reads}
A learner meets its input one step at a time and acts at each step: it emits a symbol, changes its state, writes a record to its memory. Everything its action depends on is its \emph{situation}, and the learner \emph{reads} what the situation contains.

\begin{definition}
A quantity in a learner's situation is a \emph{coordinate} if longer inputs keep giving it values that shorter ones never gave: a position, a count, a time, a distance, an activity that rises with length. It is a \emph{coincidence} if it says whether two events agree, as two symbols or two readings of a clock do or do not; a coincidence has the same few values at every size.
\end{definition}

\begin{theorem}[The hole]
\label{thm:hole}
(i) A learner that reads a coordinate while it answers has, for every training size $N$, a twin that acts as it does in every situation the training inputs produce and answers differently on some longer input. (ii) A learner whose situation holds only coincidences, the present symbol, a state from a finite set and records from a finite set is fixed by finite experience: for every task there is a size $N_{\rm teach}$ such that every such learner correct on all inputs up to $N_{\rm teach}$ is correct on every input its memory can hold.
\end{theorem}

For (i), change the learner's answers only where the coordinate takes a value never met on inputs of size $N$ or less: no example tells the twins apart, and a large network can be either. This is Goodman's riddle made exact, since grue reads the date, and Kripke's quus needs a state that adding small numbers never visits. Kripke's sceptic held that no finite disposition settles plus against quus beyond the numbers met; a disposition that reads only coincidences does, being a finite table over situations that small numbers visit. For (ii), every situation on a long input is one of finitely many; finitely many situations allow finitely many rules; each wrong rule fails on some input of finite size; and $N_{\rm teach}$ is the largest of these sizes (Appendix~\ref{app:hole}). One procedure is fixed by far less experience than the whole class of learners needs. By some size $N^\ast$ it has met every situation it will ever meet, and a learner that acts as the procedure on every input up to $N^\ast$, moves and writes included, acts as it at every size its memory can hold (Lemma~\ref{lem:closure}, Appendix~\ref{app:hole}). Reversal meets its eight situations on inputs of length $2$, so $N^\ast=2$; binary addition, with its carries, has $N^\ast=9$ (Figure~\ref{fig:concept}b). Five symbols can fix a rule for a million.

\subsection{Memory that looks the same everywhere: clocks are cycles}
\label{sec:clocks}
\label{sec:symmetry}
A learner's memory has to grow with its input, and the learner has to meet only finitely many situations in it. Both hold if the memory looks the same from every place. If the learner's moves through the memory form a group, so that every move can be undone and two moves combine into a move, and the group carries any place to any other, a learner that reads only what lies within a few moves of where it stands meets no more situations than there are ways to fill those few places (Appendix~\ref{app:homogeneity}). The address of a place is a coordinate, and so is the distance between two places, which grows with the memory; whether two places are the same, or a few moves apart, is a coincidence. Klein defined a geometry by a group of transformations and its content by what the group leaves unchanged \citep{klein1893comparative}. A learner that generalizes reads what the group of its memory leaves unchanged within a few moves of where it stands.

Two distant moments can be linked by a coincidence only through something that changes between them and can take an earlier value again: a clock. Four things are asked of a clock. It has finitely many states, as every physical memory has. Each move can be undone, so that a past state can be visited again. Moves commute, so that where the clock ends up depends only on how far it has moved; if they did not, the way back would have to undo them in reverse order, and remembering that order is already a memory. And one elementary move, repeated, visits every state.

\begin{proposition}[Clocks are cycles]
\label{prop:cycles}
A clock with these properties is a cycle of $L$ states that the elementary move advances by one, and every other move advances by some fixed number of states. By the Chinese remainder theorem a cycle whose period $L$ is a product of numbers no two of which share a factor is the same as one cycle of each of those periods, all advanced together: cycles of $3$ and $4$ driven by one move are a cycle of $12$ (Figure~\ref{fig:concept}c). Several clocks, each with its own move, form a torus, a cycle of cycles.
\end{proposition}

\label{sec:representation}
A cycle of period $p$ is held by a ring of neurons carrying one bump of activity, a ring attractor \citep{zhang1996representation,seelig2015neural}. Cycles whose periods $p_1,\dots,p_M$ share no factor turn together without carries, and together they tell apart $L=p_1\cdots p_M$ moments at a cost of $p_1+\dots+p_M$ units: the periods $2,3,5,7,11$ tell apart $2{,}310$ moments with $28$ units. The largest $L$ that a given number of units can reach is Landau's function, which grows faster than any power of that number (Appendix~\ref{app:cost}). The entorhinal grid code has this form: a few discrete modules, each a torus \citep{gardner2022toroidal}, which respond independently to a change of environment yet move together \citep{stensola2012entorhinal,waaga2022grid,fiete2008what}. The cycles follow from generalization, and the economy of the grid code from the cycles.

\subsection{Coincidence, catalogue and machine}
\label{sec:catalogue}
A neuron tuned to one reading on several cycles fires only where all of them stand as it requires: it detects a coincidence of readings, which is how place fields arise from grid cells \citep{solstad2006grid}.

Let every moment be recorded with the readings its clocks had, and let memory be searched with one question: \emph{what was recorded the last time this clock showed this reading?} Such a catalogue is a memory written and read by address, as computation requires \citep{gallistel2009memory,gershman2025key}, and it is the index the hippocampus keeps \citep{teyler1986hippocampal}. The machine (Figure~\ref{fig:concept}d) has a controller, clocks $g^1,\dots,g^C$ and heads $1,\dots,R$, each of which looks up the past. Head $i$ compares the present reading of its query clock $q_i$, plus an offset $\delta_i$, with the past readings of its key clock $c_i$, which may be the same clock:
\begin{equation}
\begin{aligned}
&\text{move:}\;\; g^c_t = g^c_{t-1}+\Delta^c_t \ (\mathrm{mod}\ L),\quad \Delta_t=\mu(h_{t-1},x_t),\\
&\text{find:}\;\; s^i_t = \max\{s<t:\ g^{c_i}_s=g^{q_i}_t+\delta_i,\ \gamma_i(w_s)=1\},\\
&\text{act:}\;\; (h_t,y_t) = \Phi(h_{t-1},x_t,w_{s^1_t},\dots,w_{s^R_t}),\\
&\text{record:}\;\; w_t=\omega(h_t,x_t).
\end{aligned}
\label{eq:machine}
\end{equation}
From its state $h_{t-1}$ and the present symbol $x_t$ the controller moves every clock by at most two readings, $\Delta^c_t\in\{-2,\dots,2\}$, or resets it \home{} to zero, $g^c_t=0$. Head $i$ then searches the records its gate $\gamma_i$ accepts, a learned bit written with each record, and returns the latest one made when its key clock showed the reading its query clock shows now, plus the offset; if there is none it returns nothing. The controller, with finitely many states, updates its state from the symbol and the records found, emits $y_t$ and records the moment as $w_t$, one of finitely many records; of its clocks it learns only whether a match exists and what was recorded there. This is attention with exact matching, where the latest match wins; a graded attention \citep{graves2014neural,weston2015memory} would report a distance.

\label{sec:range}
\textbf{The range.} Moments $L$ apart show the same reading, so a head that compares readings $L$ or more apart can find the wrong moment. The \emph{span} of an input, for a given machine, is the largest distance between a reading a head asks for and the reading of any record its gate accepts, measured on clocks that never wrap: how far back a head ever has to look. The reversal procedure of Figure~\ref{fig:concept}e has span $n-1$ on $n$ symbols. An input is \emph{within range} if its span is below $L$, the \emph{range}.

\begin{theorem}[The range belongs to the clocks]
\label{thm:range}
(i) On every input within range the machine runs exactly as the same controller with unbounded clocks, whatever pairwise coprime cycles make up $L$. (ii) Beyond the range, the first departure from that run is a lookup that finds a record a nonzero multiple of $L$ readings away, counted on clocks that never wrap; by then the span so far has reached $L$.
\end{theorem}

The controller is learned; the range belongs to the clocks. Cycles of another product $L'$ move the range to $L'$ and change no run within both ranges, so a learned rule can be given a longer memory as a Turing machine is given a longer tape. The same confusion of moments $L$ apart predicts, with no free parameter, how the reversal procedure of Figure~\ref{fig:concept}e fails beyond its range: the first $L$ answers find their own symbols, each later one finds a symbol a multiple of $L$ places back, right half the time on random bits, so the accuracy at length $n>L$ is $(L+\tfrac12(n-L))/n$ (Appendix~\ref{app:range}). Transformers whose positional code is a set of rotations with periods $5,7,11,13$ copy strings up to length $5{,}005$, the product \citep{guo2026copying}, where the theorem puts the range of such cycles.

\label{sec:forbids}
\textbf{What leaves the rule open.} Three things let a learner fit its data without the data fixing its rule. A blurred match is strong between near readings and weak between far ones, so it reports a distance, a coordinate; clocks and controller have to sit in discrete attractor states \citep{hopfield1982neural}. A controller with states to spare can fit short inputs by holding them, or by counting them and changing its answers past the count, a quus of its own, so that $N_{\rm teach}$ grows with its states (Lemma~\ref{lem:room}); a small working memory leaves no such room and pushes the procedure into the catalogue \citep{newport1990maturational,elman1993learning}. And the readings themselves keep taking new values up to the range, so a controller that sees them has a rule the data leave open: the machine moves its clocks blind. Adding one common amount to every clock, and to the reading \home{} returns to, changes no run, since the controller sees only equalities between readings, which a common shift preserves. This is Leibniz's argument that a world shifted as a whole, or created a year earlier, is the same world \citep{alexander1956leibniz}: to the controller, as space and time were to Leibniz, the memory is nothing beyond the relations in it. The argument is also a test that needs only short inputs: a learner whose answers change when every reading, or the count of its steps, is shifted by one common amount reads a coordinate (Section~\ref{sec:exp_forbid}).

\subsection{Programs are movements, recursion is return}
\label{sec:disciplines}
A \emph{stack} is a clock moved forward by each push and back by each pop; a \emph{queue} is two clocks, one writing and one reading; a \emph{copy} is a clock sent \home{} and run again over the same readings. With one clock and a head at offset $0$, Eq.~\eqref{eq:machine} runs in step with its input any deterministic pushdown automaton that reads one symbol per step, and with two clocks and a head at offset $+1$ any queue automaton (Proposition~\ref{prop:disciplines} of Appendix~\ref{app:programs}). To call a procedure is to move forward and leave the pending context in the record of that moment; to return is to move back and find it there. Recursion is a return. Procedures compose as movements compose, since every move can be undone, and when moves commute the return depends only on the net displacement, as a desert ant after a winding search heads nearly straight home \citep{muller1988path}. The levels of the Chomsky hierarchy become movements: regular tasks need no clock, nesting needs a clock that returns, copying a clock sent \home{}, and crossed dependencies, as in Dutch verb clusters, are copies, which listeners find easier than nesting \citep{bach1986crossed} (Figure~\ref{fig:concept}e; Appendix~\ref{app:reference}).

%% file: sec_computation.tex
\section{What a Finite Learner Can Compute}
\label{sec:computation}

Turing asked what a person computes with pencil and paper, and bounded what enters a step: finitely many states of mind and finitely many observed squares, on a tape that looks the same everywhere \citep{turing1937computable}. His conditions came from the limits of a human computer; here they come from generalization, and they are the same conditions: the bound that lets one finite table serve every tape is the bound that lets finite experience fix the table. A tape is a clock, and its cells are records.

\begin{theorem}[Clocks are tapes]
\label{thm:tapes}
(i) A Turing machine with $C$ work tapes that reads one input symbol per step is, step for step, a machine of Eq.~\eqref{eq:machine} with $C$ unbounded clocks, each read by one head at offset $0$: the reading of a clock is the position of a head, and the record found under it is the symbol last written there. (ii) Conversely, a machine of Eq.~\eqref{eq:machine} with $C$ unbounded clocks, each read by one head at offset $0$, whose clocks move by at most one reading, by an amount fixed by the controller's state alone, is a Turing machine with $C$ tapes whose symbols are records. (iii) Every machine of Eq.~\eqref{eq:machine} with unbounded clocks is simulated in real time by a multitape Turing machine.
\end{theorem}

Given steps without input, one clock runs any Turing machine whose tape fits in its range, and an unbounded clock runs them all: a controller that never reads a coordinate computes everything a computer computes. At one step per input symbol, the machine with unbounded clocks computes what multitape Turing machines compute in real time \citep{rabin1963real,hartmanis1965computational}. In real time every added tape adds power \citep{aanderaa1974k,paul1981information}, so such clocks are a resource no controller replaces. Multiplication bit by bit as the factors arrive is beyond any number of tapes \citep{cook1969minimum,paterson1974improved}; given time to think, the machine multiplies, as a chain of thought extends what a Transformer computes \citep{merrill2024expressive}. With Theorem~\ref{thm:hole}(ii) the same class is what finite experience fixes: a learner of Eq.~\eqref{eq:machine} with a bounded controller and unbounded clocks has a size $N_{\rm teach}$ such that a fit up to $N_{\rm teach}$ is correct at every length, and the rules it can hold are exactly the procedures that multitape Turing machines run in real time. What a finite mind can learn from finite experience and what it can compute in step with the world are the same procedures.

Networks have been proven universal by keeping the unbounded part of a computation in a precision or an address \citep{siegelmann1995computational,perez2021attention,graves2014neural}, stores that are coordinates; the machine keeps it in the positions of its clocks, which its controller never reads, so it is universal and finite experience fixes its rule. A memory with the form of a space, read by coincidence, is a tape, and a tape carries every computation: the circuit that evolved to find the way home learns arithmetic and grammar because the abilities share a form.

%% file: sec_brain.tex
\section{The Machine in the Brain}
\label{sec:brain}

\subsection{One system, two descriptions}
\label{sec:cann}
The brain's circuit for a cycle is a continuous attractor network: neurons on a ring excite their neighbours and share inhibition, so that a bump of activity rests at any position, moves with velocity inputs, and with firing rate adaptation moves by itself \citep{amari1977dynamics,zhang1996representation,mcnaughton2006path,burak2009accurate,chu2024firing}. For neurons at positions $\theta$ on a ring, with synaptic input $U$, firing rate $r$, adaptation $V$ and a bias of amplitude $\epsilon$ that carves $p$ resting places for the bump, the wells,
\begin{equation}
\begin{gathered}
\tau\,\partial_t U = -U + \rho\,J\!*\!r - V + \epsilon\cos p\theta + I,\\
\tau_v\,\partial_t V = -V + m\,U,\qquad r = [U]_+^2\big/\big(1+k\rho\!\textstyle\int\! [U]_+^2\big),
\end{gathered}
\label{eq:cann}
\end{equation}
with $J*r$ the convolution of the firing rate with a Gaussian kernel, a density $\rho$ of neurons, global inhibition $k$, adaptation of strength $m$ and time constant $\tau_v$, and an input $I$ that carries the moves, the reset \home{} and noise; a move is a pulse that adds a shifted copy of the recurrent input and pushes the bump on by whole wells \citep{zhang1996representation}. A clock is $M$ rings of periods $p_1,\dots,p_M$, no two sharing a factor, driven by the same pulses; a ring has as many wells as its period has steps, so one pulse turns every ring by one well, as one velocity signal turns every grid module. Two maps join network and machine: $\Pi$ takes each ring to the angle at which its bump stands, and $Q$ rounds it to the nearest well, so that $Q\circ\Pi$ sends the network to the wells of its rings, which the Chinese remainder theorem reads as one of the $L=p_1\cdots p_M$ readings of the clock (Figure~\ref{fig:brain}c).

\begin{theorem}[One system, two descriptions]
\label{thm:neural}
(i) Turning every ring by whole wells, and with it the well \home{} returns to, commutes with Eq.~\eqref{eq:cann}: the turns of the clock, which form its group, are a symmetry of the network. (ii) Without noise, under the conditions of Appendix~\ref{app:neural} (a weak bias, suitable pulses, adaptation strength $m$ below $\tau/\tau_v$), each ring has $p$ stable bumps, one in each well, a pulse of $n$ wells carries every bump $n$ wells on, and on these states the network is the clock: $Q\circ\Pi\circ F_\Delta=\Delta\circ Q\circ\Pi$, with $F_\Delta$ one step of Eq.~\eqref{eq:cann} under the move $\Delta$. (iii) A place cell that fires only when every ring stands where it stood when the cell was recruited detects the coincidence of the present reading with the recorded one; a cell that tolerates $j$ disagreeing rings fires again after the product of the $M-j$ smallest periods, which is then its range.
\end{theorem}

The proof reduces the network to the motion of its bumps \citep[Appendix~\ref{app:neural};][]{wu2008dynamics,fung2010moving}. From the same reduction, noise carries a bump out of its well at a rate exponentially small in the well's depth \citep{kramers1940brownian}, and adaptation above $\tau/\tau_v$, or above a higher value when there are wells to cross, makes a bump lead a slow input or travel by itself \citep{mi2014spike,chu2024firing}. Turning every ring by one well moves every bump and leaves every record, every controller state and every equality between readings as it was. These quantities, which the symmetry leaves unchanged, are the symbols. The symbolic machine is the network's dynamics on its stable states, written in terms of what its symmetry leaves unchanged, and under the conditions of Theorem~\ref{thm:neural} every run of the machine is a run of the network. With Theorem~\ref{thm:tapes} the chain closes: a Turing machine is a machine of clocks, and a machine of clocks is an attractor network (Figure~\ref{fig:brain}b). Marr held the three levels of description to be only loosely related \citep{marr1982vision}; here they are one system described three times, with the symmetry as the map from the network to the clocks and Theorem~\ref{thm:tapes} as the map from the clocks to the tape.

A record is data, written in one trial and never trained; the rule is the finite table of the controller, learned, and unable to depend on how many records there are, since that number is a coordinate. The fast store of episodes and the slow learner of rules \citep{marr1971simple,mcclelland1995why} are two kinds of change in one machine, kept apart by the principle, and a learner without the catalogue keeps its rules, as a patient without a hippocampus learns a skill and keeps no episode \citep{corkin1968acquisition}.

\subsection{The parts in the brain}
\label{sec:anatomy}
A grid module is a torus \citep{gardner2022toroidal} of two rings, one for each of two directions of its lattice. Grid cells track space, time and pitch \citep{kraus2015during,aronov2017mapping}, and a grid-like code organizes abstract spaces \citep{constantinescu2016organizing}. The moves are speed and heading signals \citep{kropff2015speed}, counted steps in ants \citep{wittlinger2006ant} and copies of the controller's commands \citep{crapse2008corollary}; in mental navigation the entorhinal cortex of macaques is periodic while the body is still \citep{neupane2024mental}. \home{} is the reset of a module at a wall, which corrects the ring perpendicular to it \citep{hardcastle2015environmental}, at every turn of a hairpin maze, where the grid map starts again \citep{derdikman2009fragmentation}, and at the boundary between events \citep{baldassano2017discovering}. Place cells are the coincidence detectors \citep{okeefe1971hippocampus,solstad2006grid} of whatever clocks reach them, grid modules the best known. Time cells are the same detectors on a clock turned by time \citep{macdonald2011hippocampal,eichenbaum2014time}. The catalogue is the hippocampal index \citep{teyler1986hippocampal}, built as in \citet{chandra2025episodic} from conjunctions of grid states, where a cell is recruited when none answers the present reading: a plateau potential potentiates the inputs active around it \citep{bittner2017behavioral}, so the cell comes to fire where the clocks stand as they stood then, and the record is written in one trial. The controller is the prefrontal circuit that holds the pending items of a procedure, few and discrete \citep{cowan2001magical}, gated by the basal ganglia \citep{frank2001interactions,oreilly2006making}; it moves the clocks by commands and reads only what the heads return (Figure~\ref{fig:brain}a).

\textbf{What the theory explains} (Appendix~\ref{app:evidence}). \emph{The clocks are given.} Toroidal structure is present in the entorhinal cortex of rat pups at ten days of age, before their eyes open and before they explore \citep{guardamagna2026toroidal}, as the head direction ring is before the eyes open \citep{bassett2018self}, and grid fields mature last \citep{langston2010development,wills2010development}: the form precedes its content. \emph{Coordinates move, coincidences stay.} Between rooms the grid realigns when place cells remap \citep{fyhn2007hippocampal}, the relative phases within a module stay fixed \citep{yoon2013specific}, and each module shifts by its own amount \citep{lykken2025functional,monaco2011modular}: every place gets a new combination of readings, under which the old coincidences are not met, and a new catalogue opens on the same clocks. \emph{Large spaces give many fields} (Figure~\ref{fig:brain}d). A cell that needs only some of its modules to agree fires wherever that agreement recurs (Theorem~\ref{thm:neural}(iii)); in a $200$\,m tunnel nearly nine place cells in ten have several fields \citep{eliav2021multiscale}. \emph{Replay is a clock in motion} (Figure~\ref{fig:brain}e). A clock run over stored readings returns their records in order, backward at the end of a run and forward before it \citep{foster2006reverse,diba2007forward}, hopping from one position to the next \citep{pfeiffer2015autoassociative}, along paths never travelled \citep{gupta2010hippocampal}, and in the learned order of an abstract task \citep{liu2019human}. Section~\ref{sec:computation} found procedures that need steps without input; rest and sleep supply them, and replay is the machine running while nothing arrives. A head set to ask a few readings ahead reads ahead, as place and grid cells sweep ahead of the animal within each cycle of the theta rhythm \citep{skaggs1996theta,vollan2025left,chu2024firing}. \emph{Order is kept by coincidence.} Serial recall exchanges items that held the same position in different groups \citep{henson1998short}, and an event boundary, a reset \home{}, breaks the order of items across it \citep{zheng2022neurons}. \emph{Counting and estimating obey different laws} (Figure~\ref{fig:brain}f). An estimate reads a graded magnitude, a coordinate, with errors proportional to the number \citep{nieder2003coding}; a count is a clock read by coincidence, exact until a slip, with errors that grow as the square root of the count \citep{cordes2001variability}.

\textbf{Predictions} (Appendix~\ref{app:predictions}). Exact sequence memory fails where the joint phase of the grid modules in play repeats, and lesions of ventral medial entorhinal cortex, which holds the modules of largest period, shorten it most; place cells made tolerant of $j$ disagreeing modules, by weakening the inhibition that sharpens their fields \citep{grienberger2017inhibitory}, cap exact memory at the product of the $M-j$ smallest periods; a replay during which the grid stands still would refute the identification; memory for the order of events rests on the periodic time code of the medial entorhinal cortex \citep{kraus2015during,heys2018evidence} and memory for how long ago on the drifting code of the lateral \citep{tsao2018integrating}; and the prefrontal code of the pending items of a procedure is unchanged when a new room shifts every grid module and the place cells remap.

%% file: sec_experiments.tex
\section{Experiments}
\label{sec:experiments}

Six experiments test six predictions: a rule fitted on strings of up to $40$ symbols holds to length $500$, with clocks that move as the programs of Section~\ref{sec:disciplines} (Section~\ref{sec:exp_bench}); exchanged clocks carry the rule as far as their range, and beyond it the rule fails on the curve of Section~\ref{sec:range} (Section~\ref{sec:exp_range}); a learner that reads a coordinate fails beyond its training data, and Leibniz's shift shows on those data which learners read one (Section~\ref{sec:exp_forbid}); experience shorter than $N^\ast$ fixes no rule, five symbols can fix one for a million, and each independent count takes a clock (Section~\ref{sec:exp_teach}); trained controllers run unchanged on neurons (Section~\ref{sec:exp_neural}); and Turing machines compiled onto clocks run as on tapes (Section~\ref{sec:exp_tm}).

\begin{figure*}[t]
\centering
\includegraphics[width=0.89\textwidth]{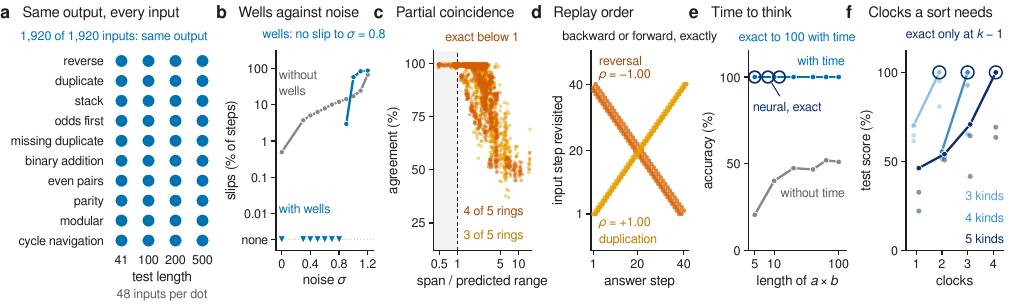}
\caption{\textbf{The machine as neural dynamics and as a computer.} \textbf{(a)} Trained controllers run on neurons: agreement with the symbolic output on every test input. \textbf{(b)} Clock slips against noise, with and without wells. \textbf{(c)} Place cells that tolerate one or two disagreeing rings: agreement against the span divided by the predicted range. \textbf{(d)} Replay order in reversal and duplication. \textbf{(e)} A schoolbook multiplier written as a three-tape machine, run on clocks with and without time to think; circles: the same machine on neurons. \textbf{(f)} Sorting $k$ kinds of symbol, every run: no run with fewer than $k-1$ clocks fits, and with $k-1$ clocks training finds a machine that sorts exactly (circled).}
\label{fig:neural}
\end{figure*}

\label{sec:exp_setup}
\textbf{The machine and the protocol.} The controller is a gated recurrent unit of $256$ units. It moves three clocks of rings $2,3,5,7,11$ ($L=2{,}310$) and reads the catalogue through $36$ gated heads, one for each ordered pair of clocks, the same clock included, and each offset in $\{-1,0,1,2\}$; at every step it writes a record of $16$ numbers, and it receives the present symbol and the records found, nothing else. The controller of Eq.~\eqref{eq:machine} has finitely many states; this one has a continuous state, pushed toward a few discrete values by Gaussian noise added in training. In training each clock holds a distribution over readings, so that gradients flow, sharpened to a single reading as training proceeds; every test runs the discrete machine (Appendix~\ref{app:methods}). A run is trained on lengths $1$ to $40$ and tested on $32$ random inputs at every length from $41$ to $500$. It \emph{fits} if it is at least $99\%$ correct on validation inputs of lengths $31$ to $40$, with up to six attempts and test lengths never entering selection; it \emph{extrapolates} if its mean score over the test lengths is at least $99\%$; and a task is \emph{solved} if the mean over its three seeds is. LSTMs of $256$ units and Transformers of five layers and width $64$, with rotary positions or none, are trained on the same data with the same selection; code, results and trained machines accompany the paper (Appendix~\ref{app:code}).

\subsection{Short training, long tests}
\label{sec:exp_bench}
\label{sec:exp_dynamics}
The fifteen tasks of \citet{deletang2023neural} span the Chomsky hierarchy (Appendix~\ref{app:tasks}). Four need no memory beyond a finite state, among them parity and an expression evaluated modulo five. The others read their past: reverse a string ($100111\to111001$), duplicate it, write its second, fourth and sixth symbols and then the rest ($000111\to011001$), run pushes and pops on a stack and read it out, add binary numbers lowest bit first, find the symbol hidden in a string written twice, evaluate bracketed arithmetic and equations, sort, multiply, take a square root. The theory says in advance which of them three clocks can carry out in this format: twelve have a procedure that keeps pace with the input in three clocks (Appendix~\ref{app:reference}); multiplication has none bit by bit as the factors arrive (Section~\ref{sec:computation}), none is known for square root, and sorting five kinds of symbol needs four counts, one clock each (Section~\ref{sec:exp_teach}).

Of the \res{E1.fit.m} runs of the machine that fit their training lengths, \res{E1.fit.k} extrapolate, with the same accuracy at length $500$ as at $41$ (Figure~\ref{fig:tests}a); of the \res{E1.lstm.m} LSTM runs that fit, \res{E1.lstm.k} do, all on tasks that a finite state solves, and of the \res{E1.tf.m} Transformer runs, \res{E1.tf.k}. The machine solves \res{E1.nexact} tasks, five of which no LSTM or Transformer solves: reversal, duplication, stack manipulation, odds first and binary addition, on which the best published LSTM and Transformer runs reach \res{E1.prev.binary_addition} to $95.1\%$ (Table~\ref{tab:bench}). The networks of the benchmark that extrapolate on them were given a memory, a stack or a tape, and a tape is a clock. No run fits the three tasks without a procedure in three clocks, nor bracketed arithmetic and equation solving, which no network solves, and on cycle navigation two seeds of three extrapolate (Appendix~\ref{app:bench}).

The clocks show what was learned. In every reversal machine a clock advances one reading per input symbol and walks back one per answer, so that each answer finds the symbol recorded at the present reading; in every duplication machine a clock goes \home{} at the first answer and runs forward over the readings the input was filed under (Figure~\ref{fig:tests}b). These are the stack and the copy of Section~\ref{sec:disciplines}, reverse and forward replay, found from inputs and outputs alone.

\subsection{The range belongs to the clocks}
\label{sec:exp_range}
The controller never sees its clocks, so they can be exchanged after training; by Theorem~\ref{thm:range} the range is then the product of the new periods, and beyond it reversal should score $(L+\frac12(n-L))/n$ (Section~\ref{sec:range}). We run the trained machines of eight tasks, without retraining, with rings $3,4,5$ ($L=60$), $2,3,5,7$ ($210$), $2,3,5,7,11$ ($2{,}310$) and $3,5,7,11,13$ ($15{,}015$), and the reversal machines also with rings $11,13,17,19,23$ ($L=\res{E2.mega.L}$) on strings of $10^4$, $10^5$ and $10^6$ symbols.

Every input agrees with Theorem~\ref{thm:range}: on \res{E2.in.phrase} inputs within range the output is that of unbounded clocks, and on \res{E2.alias.phrase} inputs with a misread the first misread comes after the span has reached $L$. A rule learned from strings of $40$ symbols therefore runs as far as the clocks reach: with rings $11$ to $23$ every reversal machine reverses a string of a million symbols without error. Beyond the range, reversal follows the predicted curve to within \res{E2.rev.err} percentage points on average (Figure~\ref{fig:tests}c), and for all \res{E2.nmach} machines that err the first error moves out with $L$, at \res{E2.ratio.lo} to \res{E2.ratio.hi} times $L$ (Figure~\ref{fig:tests}d).

\subsection{What may not be read}
\label{sec:exp_forbid}
A coordinate leaves the rule open beyond the training lengths (Theorem~\ref{thm:hole}(i)), and a blurred coincidence is a coordinate, since it reports a distance (Section~\ref{sec:forbids}). We retrain five tasks, three seeds each, with \emph{graded clocks}, the soft readings of training kept at test, so that a head averages near matches; with the \emph{step index} $t/40$ given to the controller; and with the \emph{phases} of all its clocks given to the controller, which together are the reading (Appendix~\ref{app:controls}). Leibniz's test (Section~\ref{sec:forbids}) tells whether a run reads the coordinate it is offered: on inputs of $31$ to $40$ symbols we shift the step index by up to $460$ steps, which gives a short input the index of a long one, or every clock and \home{} by up to $1{,}155$ readings, and count the answers that change.

With graded clocks every run fits, and at length $500$ every task that reads its past falls to between \res{C.graded.lo} and \res{C.graded.hi}\%, while modular arithmetic, which needs no clock, stays exact (Table~\ref{tab:controls}). Given the step index or the phases, the runs on reversal, duplication and stack manipulation change \res{L.clk.phrase} of their answers under the shift and are exact at length $500$: training left the coordinate unread. Every run on modular arithmetic given the step index changes \res{L.mod.lo} to \res{L.mod.hi}\% of its answers when the index takes the values of length $500$, and there it falls to \res{C.coord.mod.lo} to \res{C.coord.mod.hi}\% (Figure~\ref{fig:tests}e). What a learner reads shows on its training data and decides how it goes on.

\subsection{How much experience, how many clocks}
\label{sec:exp_teach}
$N^\ast$ is the length by which a procedure has met every situation it will ever meet: shorter experience cannot fix the rule, and a learner that acts as the procedure up to $N^\ast$ is fixed (Lemma~\ref{lem:closure}). A learner can also fit its experience with another rule, which gives the task's answers on every input it has met and departs from them on longer ones; Theorem~\ref{thm:hole}(ii) rules these out only from the size $N_{\rm teach}$ on. We retrain six tasks on lengths $1$ to $N$, for $N=3$, $5$, $10$ and $20$, three seeds each. Each independent count takes a clock: to sort $113311$ into $111133$ in step with the input, a learner must know where each kind of symbol past the first begins, $k-1$ counts for $k$ kinds; we train sorting over $k=3$, $4$ and $5$ kinds with fewer than $k-1$ clocks and with $k-1$.

All \res{T.n} runs on short inputs fit (Figure~\ref{fig:tests}f); none of the \res{T.below.n} trained on fewer symbols than $N^\ast$ extrapolates. Five symbols fix reversal: every machine trained on at most five extrapolates, and with the cycles $11$ to $23$ reverses a string of a million symbols without error. Near $N^\ast$ the controller still has room for a quus (Lemma~\ref{lem:room}), and training finds such rules, which longer experience removes: from ten symbols, every run on reversal, duplication and stack manipulation extrapolates, modular arithmetic ($N^\ast=15$) holds from twenty, and odds first ($N^\ast=5$) from twenty, and from ten in \res{T.10.odds.k} of three. Binary addition ($N^\ast=9$) shows such a rule: from sums of up to ten symbols training finds a quus, which agrees with addition on the sums it met and departs from it on longer ones, the more often the higher the bit (Figure~\ref{fig:quus}). From sums of up to twenty symbols the departure all but vanishes: these machines add \res{Q.20.exact.lo} to $100\%$ of the sums of two numbers of a hundred bits exactly and score \res{T.20.add.lo} to \res{T.20.add.hi}\% at lengths $41$ to $500$. In sorting, no machine with fewer than $k-1$ clocks fits its training lengths; with $k-1$ clocks training finds, for each $k$, a machine that sorts exactly at every tested length (Figure~\ref{fig:neural}f).

\subsection{The same machines in neural dynamics}
\label{sec:exp_neural}
If the network of Eq.~\eqref{eq:cann} is the clock on its stable states (Theorem~\ref{thm:neural}), a controller trained on symbolic clocks runs unchanged on neurons. We replace the clocks and the catalogue of the trained machines by rings of Eq.~\eqref{eq:cann} with $16$ neurons per well; place cells, each recruited for one joint reading of the rings; and records stored as synaptic weights strengthened once (Appendix~\ref{app:neural}).

Without noise, the neural machines give the symbolic output on \res{N.exact.phrase} test inputs of ten tasks, and no bump leaves its well (Figure~\ref{fig:neural}a); with noise of strength $0.6$ on every neuron the output of \res{N.noise06.agree} inputs of five tasks is unchanged. With wells the clocks never slip in \res{N.clock.steps} steps at noise up to $0.8$; without wells they slip in \res{N.clock.nowell08}\% of steps at $0.8$ (Figure~\ref{fig:neural}b). Place cells that tolerate one or two disagreeing rings shorten the range to $210$ and $30$, the products of the rings that must agree (Figure~\ref{fig:neural}c). As they answer, reversal machines revisit the input backward and duplication machines forward (rank correlations \res{N.replay.rev} and \res{N.replay.dup}; Figure~\ref{fig:neural}d). The controller cannot tell the network from the symbolic machine.

\subsection{Turing machines and time}
\label{sec:exp_tm}
By Theorem~\ref{thm:tapes}, a Turing machine with $C$ tapes that reads one symbol per step is a machine of Eq.~\eqref{eq:machine} with $C$ clocks. By that construction we compile \res{C.tm.n} random Turing machines with one to three tapes into controllers and run them on inputs of up to $1{,}000$ symbols, with unbounded and with bounded clocks, comparing every step with a direct simulation; a schoolbook multiplier written as a three-tape machine gets about $n^2/2$ steps without input to think, or none (Appendix~\ref{app:tm_exp}).

The machines on clocks match the simulation in \res{C.tm.phrase} runs with unbounded clocks and in \res{C.tm.within} runs with bounded clocks within range. Given time to think, the multiplier is exact up to length \res{C.think.maxn}; without it, it scores at most \res{C.think.acc0}\%, as the trained machines do (Figure~\ref{fig:neural}e).

%% file: sec_discussion.tex
\section{Discussion}
\label{sec:discussion}

Hume's question has an answer for a learner of procedures: if it reads only coincidences, each wrong rule betrays itself at some finite size (Theorem~\ref{thm:hole}), and a learner correct up to that size is correct at every size its memory can hold; the step from short inputs to long ones, an induction for a learner of coordinates, is for it a deduction. Wittgenstein's pupil, who writes $1000$, $1004$, $1008$ and holds that he goes on in the same way \citep{wittgenstein1953philosophical}, reads the size of his numbers; a learner that reads only coincidences has no state in which to go wrong. The same answer explains why brains learn from so little: a record is written in one trial, because it is data and the catalogue files it, and a rule is fixed by a few short examples, because the situations it has to cover are few. Both follow from one fact: the memory carries the size of the input, and the rule stays finite.

The memory that carries the size has a form, the group of the clocks. Kant held the forms of intuition to be given before experience and conditions of its possibility \citep{kant1998critique}; Poincar\'e found the origin of geometry in the group of displacements, the changes our movements can undo \citep{poincare1905science}. The group of the clocks is both, present in a pup's entorhinal cortex before it has explored \citep{guardamagna2026toroidal} and a group of the animal's own displacements, and the theory says why a learner needs such a group: of a memory that grows it can read only what the group leaves unchanged. Kant's two forms are one form under two drives: a clock turned by steps is space, and the same clock turned by seconds is time \citep{kraus2015during}. The hippocampus holds space, episodes, sequences and abstract spaces \citep{okeefe1978hippocampus,teyler1986hippocampal,eichenbaum2014time,buzsaki2013memory,behrens2018what}; one organ serves all because a clock does not know what turns it, and the catalogue files any content under its readings.

Fodor and Pylyshyn held that systematic thought needs a classical architecture of combined symbols \citep{fodor1988connectionism}; Smolensky, that symbols describe a network's dynamics only approximately \citep{smolensky1988proper}; van Gelder, that cognition is dynamics and computation a misdescription of it \citep{vangelder1998dynamical}. In the machine symbols and dynamics describe one system exactly: the group of the clocks is the symmetry of the network, the position of a bump is a coordinate, and what the group leaves unchanged, the records, the controller states and the equalities between readings, are the symbols. A network has symbols where its dynamics has a symmetry: a symbolic computation is a dynamical system described by its invariants, dynamicism with the symbols kept. Systematicity \citep{marcus2001algebraic,lake2018generalization,hupkes2020compositionality} is the composition of movements, since to run one procedure inside another is to move away and return; productivity is the rule staying finite while the memory grows. Computation that generalizes across scale rests on relations that a group of movements leaves unchanged, and for machine learning this is a design rule \citep{cohen2016group,bronstein2021geometric}: let the memory carry the size, read only what takes finitely many values, and reach the past by coincidence.

Hegel's image of the bad infinite is an endless line, and of the true infinite a circle that returns into itself \citep{hegel1969science}. A finite learner meets a stretch of a line and the whole of a circle. Brains build the line from circles.

%% file: appendix.tex
\makeatletter
\def\addcontentsline#1#2#3{\addtocontents{#1}{\protect\contentsline{#2}{#3}{\thepage}{\@currentHref}\protected@file@percent}}
\makeatother
\etocdepthtag.toc{mtappendix}
\etocsettagdepth{mtmain}{none}
\etocsettagdepth{mtappendix}{subsection}
\etocsettocstyle{\section*{Appendix}\noindent Appendices~\ref{app:theory} to~\ref{app:reference} give the details of the theory: the statements and proofs, the machine in neural dynamics, and the reference controllers. Appendix~\ref{app:methods} describes the trained machine and its training, and Appendix~\ref{app:experiments} gives one subsection to each experiment of Section~\ref{sec:experiments}, with why it was run, its design, its results and what they mean. Appendices~\ref{app:evidence} to~\ref{app:code} give the neural evidence, the related work and the code.\par\medskip}{\medskip}
{\setlength{\parskip}{0.5pt}\small\tableofcontents}
\clearpage
\addtocontents{toc}{\protect\medskip\protect\noindent\textbf{Theory}\protect\par}
\input{app_theory.tex}
\FloatBarrier
\input{app_neural.tex}
\FloatBarrier
\input{app_reference.tex}
\FloatBarrier
\addtocontents{toc}{\protect\medskip\protect\noindent\textbf{Experiments}\protect\par}
\input{app_methods.tex}
\input{app_experiments.tex}
\FloatBarrier
\addtocontents{toc}{\protect\medskip\protect\noindent\textbf{Evidence and context}\protect\par}
\input{app_evidence.tex}
\FloatBarrier
\input{app_related.tex}
\FloatBarrier
\input{app_code.tex}
\FloatBarrier

%% file: app_theory.tex
\section{Theory: Statements and Proofs}
\label{app:theory}
\label{app:proofs}

This appendix proves the statements of Sections~\ref{sec:theory} and~\ref{sec:computation}: Theorems~\ref{thm:hole} to~\ref{thm:tapes} and Proposition~\ref{prop:cycles}, which the main text states, and Lemmas~\ref{lem:closure} and~\ref{lem:room} and Proposition~\ref{prop:disciplines}, which the main text uses and which are stated here. Appendix~\ref{app:neural} treats the neural dynamics of Section~\ref{sec:brain}. Table~\ref{tab:theorymap} lists each statement, what it says in a few words, and where its proof is.

\begin{table}[ht]
\caption{\textbf{The formal results and where they are proved.}}
\label{tab:theorymap}
\centering\small
\begin{tabular}{@{}lll@{}}
\toprule
Statement & What it says & Proof\\
\midrule
Theorem~\ref{thm:hole}, the hole & a coordinate leaves the rule open; coincidences fix it & Appendix~\ref{app:hole}\\
Lemma~\ref{lem:closure} & acting as the procedure up to $N^\ast$ fixes it within range & Appendix~\ref{app:hole}\\
Lemma~\ref{lem:room}, room for a quus & spare controller states let short experience fit another rule & Appendix~\ref{app:hole}\\
Proposition~\ref{prop:cycles}, clocks are cycles & a finite, reversible, commuting clock is a cycle or a torus & Appendix~\ref{app:cycles}\\
Characters and coincidence & commuting linear moves give independent periodic components & Appendix~\ref{app:characters}\\
Homogeneity & a memory that looks the same everywhere gives finitely many situations & Appendix~\ref{app:homogeneity}\\
Cost of coprime cycles & the range grows faster than any power of the number of units & Appendix~\ref{app:cost}\\
Theorem~\ref{thm:range}, the range & the range belongs to the clocks; aliasing of reversal & Appendix~\ref{app:range}\\
Proposition~\ref{prop:disciplines}, programs & stacks and queues as movements of clocks & Appendix~\ref{app:programs}\\
Theorem~\ref{thm:tapes}, clocks are tapes & the machine and multitape Turing machines in real time & Appendix~\ref{app:tapes}\\
Theorem~\ref{thm:neural}, two descriptions & the attractor network is the machine on its stable states & Appendix~\ref{app:neural_proof}\\
\bottomrule
\end{tabular}
\end{table}

\subsection{Setting}
\label{app:setting}
An input is a string $x\in\Sigma^n$ followed by one placeholder for every output symbol; $n$ is its size, and the steps at which the placeholders arrive are the answering steps. A learner runs step by step. At step $t$ it meets a situation $\varsigma_t\in\mathcal S$, which contains the present input symbol, and takes an action $a_t=\lambda(\varsigma_t)\in\mathcal A$, which includes the symbol it emits and every change it makes to its own state and memory; the situation at step $t+1$ is a function of the input and of the actions taken so far. A class is a set of decision tables $\lambda:\mathcal S\to\mathcal A$. It is \emph{rich} if, with any table, it contains every table that differs from it only in the symbol emitted in some set of situations met at answering steps. A learner is correct on $x$ if the symbols it emits at the answering steps are the target.

\subsection{The hole (Theorem 1) and the size at which experience closes}
\label{app:hole}
\begin{proof}[Proof of Theorem~\ref{thm:hole}]
(i) Let $\kappa$ be a coordinate of the situations that a learner $\lambda$ meets at answering steps, and let $\lambda$ be correct on all inputs of size at most $N$. Since $N$ is below the range of $\kappa$, there is an input $x$ of size $n>N$ on which some answering step of $\lambda$ meets a value $\kappa^\ast$ that no answering step of any input of size at most $N$ meets. Define $\lambda'$ equal to $\lambda$ except that in every situation met at an answering step with $\kappa=\kappa^\ast$ it emits a different symbol. On inputs of size at most $N$ the two tables never meet such a situation, so their runs coincide step by step and $\lambda'$ is correct on all of them. On $x$ the runs coincide up to the first answering step with $\kappa=\kappa^\ast$, where they emit different symbols. So $\lambda'$ differs from $\lambda$ only in situations that inputs of size at most $N$ never produce, is correct on all of them and answers differently on $x$: it is the twin of Theorem~\ref{thm:hole}(i). A rich class contains $\lambda'$, so no training size below the range fixes the behaviour of a rich class beyond it.

(ii) If $\mathcal S$ and $\mathcal A$ are finite, the class contains at most $|\mathcal A|^{|\mathcal S|}$ tables. Let every table $\lambda$ come with a set $D_\lambda$ of admissible inputs, the inputs within its range. For each table that is incorrect on some input in $D_\lambda$, let $N(\lambda)$ be the smallest size of such an input, and let $N_{\rm teach}$ be the largest of these finitely many sizes. A table that is correct on all inputs up to $N_{\rm teach}$ is none of them, so it is correct on all of $D_\lambda$. For the machine of Eq.~\eqref{eq:machine}, a situation consists of the controller state, the input symbol and, for each head, a record or nothing; a record $\omega(h,x)$ is a function of a controller state and a symbol, so with finitely many controller states the situations form a finite set, and $D_\lambda$ is the set of inputs whose span is below $L$ (Theorem~\ref{thm:range}).
\end{proof}

The bound $N_{\rm teach}$ is uniform over the class and can be large. For one correct table $\lambda$ a smaller size matters. Let $\mathcal S_\lambda(N)$ be the set of situations that $\lambda$ meets on inputs of size at most $N$ in $D_\lambda$, and let the \emph{closure size} $N^\ast(\lambda)$ be the smallest $N$ with $\mathcal S_\lambda(N')=\mathcal S_\lambda(N)$ for every $N'\geq N$.

\begin{lemma}
\label{lem:closure}
If $\lambda$ is correct on every input in $D_\lambda$ and $\lambda'$ agrees with $\lambda$ on $\mathcal S_\lambda(N^\ast)$, then $\lambda'$ runs as $\lambda$ on every input in $D_\lambda$, and in particular is correct.
\end{lemma}
\begin{proof}
On any such input the runs of $\lambda$ and $\lambda'$ coincide by induction over steps: if they coincide up to step $t-1$, they meet the same situation at step $t$; it lies in $\mathcal S_\lambda(N^\ast)$, where the two tables agree, so they take the same action. The run of $\lambda'$ is the run of $\lambda$, moves, writes and answers included, and the answers are correct.
\end{proof}

The lemma separates two ways of being taught. A learner that acts as the procedure $\lambda$ on every input up to $N^\ast$, in its moves and writes as well as in its answers, has met every situation of that procedure, and these inputs fix its behaviour within range. A learner that fits the same inputs by another route, for instance by holding short inputs in its controller instead of reading them back through the clocks, meets other situations, and the lemma says nothing about it; only the uniform bound $N_{\rm teach}$ applies. How much larger $N_{\rm teach}$ is depends on the room the controller leaves.

\begin{lemma}[Room for a quus]
\label{lem:room}
Let $\lambda$ be a procedure with $k$ controller states that is correct on every input within range, and let $m\geq1$. A controller with $(m+2)k$ states, on the same clocks, heads and records, acts as $\lambda$ on every input of size at most $m$ and is incorrect on some input of size $m+1$. In the class of learners whose controllers have $K$ states, $N_{\rm teach}\geq\lfloor K/k\rfloor-1$.
\end{lemma}
\begin{proof}
Let the new controller hold a pair $(q,c)$: a state $q$ of $\lambda$, and the number $c\in\{0,\dots,m+1\}$ of input symbols read so far, stopped at $m+1$. It moves its clocks, writes its records, sets its gates and updates $q$ as $\lambda$ does in state $q$, and it emits what $\lambda$ emits, except at answering steps with $c=m+1$, where it emits another symbol. On an input of size at most $m$ the count never reaches $m+1$, so the run is the run of $\lambda$, which is correct. On an input of size $m+1$ every answering step has $c=m+1$, and the first one emits a wrong symbol. With $K\geq(m+2)k$ this controller is in the class and correct on every input up to size $m=\lfloor K/k\rfloor-2$.
\end{proof}

The count is one way to use spare states; holding a whole short input, and answering from it without the clocks, is another, and needs about $|\Sigma|^m$ states for inputs of size $m$. Either way a controller with room to spare can fit short experience with a rule of its own, and experience fixes its rule only once the inputs outrun what it can hold. A controller with fewer states is one with more states left unused, so $N_{\rm teach}$ never falls as states are added, and by the lemma it grows without bound. The smaller the working memory, the shorter the experience that fixes a rule: this makes exact Newport's \emph{less is more} \citep{newport1990maturational} and Elman's advantage of starting small \citep{elman1993learning}. This is the pattern of the experiment of Section~\ref{sec:exp_teach}: machines trained on short inputs extrapolate wherever they find the procedure, and the shortest training lengths leave room for other routes. For the reference controllers $N^\ast$ is measured by running each on $200$ random inputs at every length up to $60$ and recording where the count of distinct situations stops growing (Appendix~\ref{app:reference}; Figure~\ref{fig:concept}b).

\subsection{Clocks are cycles (Proposition 1)}
\label{app:cycles}
\begin{proof}[Proof of Proposition~\ref{prop:cycles}]
Let $\Gamma$ be the finite set of clock states and $G$ the group of permutations of $\Gamma$ generated by the moves; since each move can be undone, the moves are permutations, and since they commute, $G$ is abelian. Since repeating the elementary move $\iota$ visits every state, $\iota$ has a single orbit, so $\langle\iota\rangle$, and hence $G$, acts transitively. In a transitive action of an abelian group all points have the same stabilizer $H$ (stabilizers of points in one orbit are conjugate, and conjugation is trivial), and $H$ fixes every point, so it acts trivially; $G/H$ acts simply transitively and $\Gamma$ is a copy of $G/H$. Since $\iota$ generates a transitive cyclic subgroup, $G/H=\langle\iota H\rangle\cong\Z_L$ with $L=|\Gamma|$. The Chinese remainder theorem gives $\Z_L\cong\prod_i\Z_{p_i^{a_i}}$ for $L=\prod_i p_i^{a_i}$, with $\iota\mapsto(1,\dots,1)$: one move advances every factor by one, and two states are equal exactly when all their components are. If $d$ elementary moves act on factors of their own, each visiting every state of its factor, and commute, the same argument applied to each factor gives a product of $d$ cycles on which the moves act simply transitively, a torus.
\end{proof}

A finite set on which one move acts transitively and invertibly is a single orbit of a permutation, that is a cycle: a finite memory can be homogeneous under one kind of step only by closing on itself. Dropping commutativity leaves the group of moves free to be nonabelian, and then returning to an earlier state requires undoing the moves in reverse order; the record of that order is a stack, so a clock whose moves do not commute presupposes the memory it was meant to provide. The move \home{} of Eq.~\eqref{eq:machine}, which resets a clock, is neither invertible nor commuting and lies outside the group; it is a return to a fixed state.

\subsection{The representation: characters and coincidence}
\label{app:characters}
Section~\ref{sec:catalogue} says that a neuron tuned to one reading on several cycles detects a coincidence of readings; this section shows that a population which the moves shift linearly splits into independent cycles, and that a unit pooling them detects equality. Let the clock be held by a population whose activity is a vector, and let each move act by a fixed invertible linear map. Commuting linear maps of finite order are simultaneously diagonalizable over $\mathbb C$, and the joint eigenvectors carry the characters $\chi_k(g)=e^{2\pi ikg/L}$, $k\in\Z_L$: a move multiplies each component by a phase, and no component depends on another. Under the Chinese remainder isomorphism every character of $\Z_L$ is a product of characters of the factors, so the components group into independent cycles, one per maximal prime power. The orthogonality of characters,
\begin{equation*}
\sum_{k\in\Z_L}\chi_k(g-g')=L\,[g=g'],\qquad
\prod_m\sum_{k\in\Z_{p_m}}\chi_k(g-g')=\prod_m p_m\,[g\equiv g' \ (\mathrm{mod}\ p_m)],
\end{equation*}
says that equality of two states is the agreement of every cycle, and that a unit pooling the cycles and firing only when all agree detects it. A unit tuned to one phase of a cycle of period $p$ fires at every $p$th step along the clock, as a grid cell fires at the vertices of its lattice; a unit pooling several cycles fires where they all agree, as a place cell fed by several grid modules does \citep{solstad2006grid}. When the pooled cycles do not cover the whole clock, their agreement recurs at the least common multiple of their periods, and the unit has several fields, as place cells in large spaces have (Section~\ref{sec:anatomy}; Appendix~\ref{app:evidence}).

\subsection{Homogeneity and finitely many situations}
\label{app:homogeneity}
Let a memory be a set of locations on which a group $G$ with a finite generating set acts transitively, with contents from a finite alphabet $A$. Let a learner read, at each step, the contents at the locations $g\cdot f$ for a fixed finite set $F\subset G$ of offsets $f$ from its current location $g$, together with its state, from a finite set $Q$, and the input symbol, from $\Sigma$. Its situations take at most $|Q|\,|\Sigma|\,(|A|+1)^{|F|}$ values, whatever the size of the memory or of the input, and translating the whole memory by an element of $G$ leaves them unchanged. A read that tells locations apart, such as the label of the current location, takes as many values as there are locations visited, grows with the input, and is a coordinate, to which Theorem~\ref{thm:hole}(i) applies.

\subsection{The cost of coprime cycles}
\label{app:cost}
A cycle of $p$ states is held by a ring of $p$ units carrying one bump. A product of cycles with pairwise coprime periods needs as many units as the sum of the periods, instead of the $L$ units of one ring of the same range. The largest range for a given number of units is Landau's function, attained by periods that are powers of distinct primes; with the first $k$ primes the range grows as $e^{(1+o(1))k\log k}$ while the number $S$ of units grows as $(\tfrac12+o(1))k^2\log k$, so the range is $e^{(1+o(1))\sqrt{S\log S}}$, faster than any power of $S$. The training clocks, with periods $2,3,5,7,11$, address $2{,}310$ states with $28$ units; the clocks with periods $3,5,7,11,13$ address $15{,}015$ states with $39$ units.

\subsection{The range (Theorem 2) and the aliasing of reversal}
\label{app:range}
For a run of the machine and a head $i$, the \emph{compared readings} at step $t$ are the query reading $g^{q_i}_t+\delta_i$ and the readings $g^{c_i}_s$ of all earlier records that the gate of the head accepts; the \emph{span} of an input is the largest difference between the query reading and such a key reading, over heads and steps, computed on the run with unbounded integer clocks.

\emph{Shift invariance} (Section~\ref{sec:range}). The controller receives readings only through the equalities $g^{c_i}_s=g^{q_i}_t+\delta_i$ that select records. Adding one element $c\in\Z_L$ to every clock, and taking $c$ as the state that \home{} sets, commutes with every move and preserves every such equality, also between different clocks, and hence preserves the run. Shifting one clock alone does not, when a head compares two clocks.

\begin{proof}[Proof of Theorem~\ref{thm:range}]
(i) Let $\bar g$ be the run of the same controller with clocks in $\Z$, and $g=\bar g\bmod L$. Suppose the two runs agree up to step $t-1$; then their moves at step $t$, which depend on $h_{t-1}$ and $x_t$, agree, and $g_t=\bar g_t\bmod L$. A head compares $d=\bar g^{c_i}_s-\bar g^{q_i}_t-\delta_i$ with zero for every earlier record its gate accepts. If the span is below $L$, every such $d$ lies strictly between $-L$ and $L$, so $d\equiv0\pmod L$ exactly when $d=0$; both machines select the same record, meet the same situation and take the same action. By induction the runs agree throughout; the argument uses only the product $L$, whatever pairwise coprime cycles make it up. (ii) At the first step at which the runs differ, some head finds, in the modular run, a record with $d$ a nonzero multiple of $L$ that is more recent than the record the integer run selects, or the integer run selects none. Then $|d|\geq L$, so the span has reached $L$. The departure changes what the controller receives, and it causes an error only if the record found carries a different symbol.
\end{proof}

\emph{Other cycles} (Section~\ref{sec:range}). The controller is the same function whatever the modulus. By (i), with cycles of product $L$ and with cycles of product $L'$ the machine runs as the unbounded one on every input whose span is below both, and the inputs within range of the new machine are those whose span is below $L'$.

\emph{Reversal.} The procedure of Figure~\ref{fig:concept}e writes input $j$ at reading $j$, its gate accepts input records only, and output $k$ looks up reading $n-k+1$. For $n\leq L$ every target is the only input record with its residue. For $n>L$ the most recent input with residue $n-k+1$ is $n-k+1+L\lfloor (k-1)/L\rfloor$, which is the target exactly when $k\leq L$. Outputs $k\leq L$ are correct, and each of the other $n-L$ outputs copies a symbol that is independent of its target, correct with probability $\tfrac12$ when the input bits are independent and uniform. The expected accuracy at length $n>L$ is therefore $(L+\tfrac12(n-L))/n$, with no free parameter; at $n=L$ every output is correct. A trained machine that walks its clock back at one reading per step obeys the same formula (Figure~\ref{fig:tests}c).

\subsection{Programs (Proposition 2) and the number of clocks}
\label{app:programs}
\begin{proposition}[Programs]
\label{prop:disciplines}
With one clock and a head at offset $0$, Eq.~\eqref{eq:machine} simulates in step with its input any deterministic pushdown automaton that reads one symbol per step and pushes or pops at most one symbol per step; with two clocks and a head at offset $+1$ whose key is the writing clock, any queue automaton that reads one symbol per step and enqueues or dequeues at most one symbol per step.
\end{proposition}
\begin{proof}
\emph{Pushdown.} Let the automaton read one symbol per step and push one symbol, pop one, replace the top or leave its stack. The controller holds the automaton's state and the current top symbol, every record carries the top symbol at its moment, and the first moment carries a bottom marker at reading $0$. A push moves the depth clock by $+1$, and the new top is known. A replacement leaves the clock where it is and changes the top that the next record carries. A pop moves the clock by $-1$, and the head at offset $0$ returns the last record made at the new reading. That record was made after the last push to this level, and the clock has stayed above it since, so the levels below are untouched and the record carries the new top; a bottom marker tells the controller that the stack is empty. \emph{Queue.} A write clock advances with every enqueue and a read clock with every dequeue; the gate of the head accepts the records of enqueues, and the head whose query is the read clock at offset $+1$ and whose key is the write clock returns the symbol behind the front, which the controller holds as the next front before it decides on the next dequeue. The range is set by the number of enqueues. Tapes are treated in Appendix~\ref{app:tapes}. The reference controllers of Appendix~\ref{app:reference} instantiate these procedures for the benchmark tasks.
\end{proof}

\emph{The number of clocks.} Sorting $k$ kinds of symbol in step with the input means that the $k-1$ boundaries between kinds, the numbers $S_c$ of symbols at most $c$, are at hand when the output begins. Each $S_c$ grows with the input, and the $k-1$ counts vary independently. The procedure of Appendix~\ref{app:reference} keeps each in a clock of its own, which advances on every symbol at most $c$ and reads $S_c$ at the end of the input. A machine that places $C<k-1$ boundaries exactly and guesses the others loses about a $k$th of the output for every boundary it misses, and scores about $(C+1)/k$ on uniformly random strings. Section~\ref{sec:exp_teach} tests whether training finds a machine with fewer clocks that keeps pace.

\emph{What cannot keep pace.} Under the protocol of \citet{deletang2023neural} the whole input arrives before the first output, and each output step allows a bounded number of lookups. The $k$th bit of a product of two $n$-bit numbers depends on all pairs $(a_i,b_{k-i})$ and on a carry that grows with the input. No procedure that computes it with a bounded number of lookups per step is known, and multiplication that must emit each bit before the next input bit arrives is beyond every machine of Eq.~\eqref{eq:machine} (Appendix~\ref{app:tapes}). Square roots are similar.

\subsection{Clocks are tapes (Theorem 3) and what the machine computes}
\label{app:tapes}
A multitape Turing machine runs in \emph{real time} if it reads one input symbol at every step and emits the output of that step at once \citep{yamada1962real,rabin1963real,rosenberg1967real}. With $C$ work tapes it applies at every step a finite table
\begin{equation*}
(z',\,b^1,\dots,b^C,\,d^1,\dots,d^C,\,y)=\Psi(z,\,x,\,a^1,\dots,a^C),\qquad d^c\in\{-1,0,1\},
\end{equation*}
where $a^c$ is the symbol under head $c$, which the machine replaces by $b^c$ before it moves the head by $d^c$. In this section the clocks of Eq.~\eqref{eq:machine} are unbounded, $g^c_t\in\Z$.

\begin{proof}[Proof of Theorem~\ref{thm:tapes}]
(i) Given the Turing machine, let the controller hold $h_t=(z_t,d_t,b_t)$, the machine's state, the moves $d_t=(d^1_t,\dots,d^C_t)$ it decided and the symbols $b_t=(b^1_t,\dots,b^C_t)$ it wrote at step $t$, with $d_0=0$; let the record of step $t$ be the tuple of symbols written, $w_t=b_t$, which is a function $\omega(h_t,x_t)$ of the controller state as Eq.~\eqref{eq:machine} requires; let clock $c$ carry one head with key and query clock $c$, offset $0$ and a gate that accepts every record; and let the clocks move by $\Delta_t=d_{t-1}$, which the state $h_{t-1}$ fixes. Write $p^c_t$ for the position of head $c$ after step $t$. Then $g^c_t=p^c_{t-1}$: both are $0$ at $t=1$, and $g^c_{t+1}=g^c_t+d^c_t=p^c_{t-1}+d^c_t=p^c_t$. At the start of step $t$ the cell $p^c_{t-1}$ holds the symbol written at the last step $s<t$ with $p^c_{s-1}=p^c_{t-1}$, that is with $g^c_s=g^c_t$, or the blank if there is none; this is component $c$ of the record that head $c$ returns. The controller applies $\Psi$ to its state, the symbol and these components, and so it emits what the machine emits, writes what it writes and moves as it moves, step for step. The clock is the machine's head one step late: the machine reads, writes and then moves, and the clock moves at the start of the next step.

(ii) Given a machine of Eq.~\eqref{eq:machine} with unbounded clocks, each moved by at most one reading per step by moves $\mu(h_{t-1})$ that the controller's state fixes, and each read by one head with key and query clock $c$, offset $0$ and gate $\gamma_c$, let tape $c$ hold in cell $p$ the last record accepted by $\gamma_c$ that was made when clock $c$ showed $p$, or the blank. A Turing machine with state $h$ simulates it: at step $t$ head $c$ stands at $g^c_t$, where the move of the previous step put it, and reads the cell, which is what head $c$ of Eq.~\eqref{eq:machine} returns; the machine computes $h_t$, $y_t$ and $w_t$ as $\Phi$ and $\omega$ do, writes $w_t$ on every tape whose gate accepts it and rewrites the cell it read on the others, and moves head $c$ by $\mu^c(h_t)$. Its tape symbols are records.

(iii) Let the machine have clocks $1,\dots,C$ and heads $1,\dots,R$. For every clock $c$ take a tape $T_c$ with one track for each head $i$ whose key clock $c_i$ is $c$; square $y$ of that track is to hold the last record accepted by $\gamma_i$ that was made when clock $c$ showed $y$. On $T_c$ place five writing heads at $g^c_{t-1}+e$, $e\in\{-2,\dots,2\}$; for every head $i$ with $c_i=c$, five reading heads at $g^{q_i}_{t-1}+\delta_i+e$; and parked heads that never move, at the squares $\delta+e$ for every offset $\delta$ in use and for $\delta=0$. At step $t$ the control reads the input symbol and computes the moves $\Delta_t=\mu(h_{t-1},x_t)$. For every head $i$ it takes the cell under the reading head with $e=\Delta^{q_i}_t$, or, if clock $q_i$ returns \home{}, under the parked head at $\delta_i$. These cells hold every record made before step $t$, so they are what head $i$ returns in Eq.~\eqref{eq:machine}. The control computes $h_t$, $y_t$, $w_t$ and the gates, writes $w_t$ into the track of every head whose gate accepts it, with the writing head of clock $c_i$ at $e=\Delta^{c_i}_t$ (the parked head at $0$ after \home{}), and finally moves every head of clock $c$ by $\Delta^c_t$, or, after \home{}, lets it jump to the parked head at $\delta+e$. One head writes on each tape per step. The result is a machine with $C$ tapes, each carrying a fixed number of heads that move at most two squares per step and may jump to the square of another head; storing two cells per square turns moves of two squares into moves of one, and heads that start a bounded distance from the origin are placed during the first steps by the usual means. A tape with several heads and jumps between them is simulated in real time by multitape Turing machines \citep{kosaraju1979real}, and without jumps already by \citet{fischer1972real}; the power of machines in real time is unchanged by placing all heads on one tape with jumps between them \citep{savitch1984power}.
\end{proof}

The dictionary of (i) and (ii) is implemented in \texttt{code/orrery/tm.py}, and Appendix~\ref{app:tm_exp} checks it on random machines. Its consequences follow.

\emph{Universality given time.} Let the input be followed by blank symbols, during which the controller keeps moving. One clock then runs any Turing machine with one tape on which its input is written: while the input arrives the controller records each symbol, a record $\omega(h_t,x_t)$ that copies $x_t$, and advances the clock; then it returns the clock \home{} and runs the machine of (i), one step per blank symbol, holding the state, the move and the symbol written in $h_t$ as in (i). With an unbounded clock this is every Turing machine; with a clock of range $L$ it is every Turing machine whose head stays within $L$ squares (Theorem~\ref{thm:range}).

\emph{The class computed in real time.} By (i) and (iii), the functions that machines of Eq.~\eqref{eq:machine} with unbounded clocks compute one step per input symbol are exactly those that multitape Turing machines compute in real time; for languages this is the class of \citet{rosenberg1967real}. With clocks of range $L$ a machine computes the same function on every input within range (Theorem~\ref{thm:range}).

\emph{Every tape counts.} For every $k$ some language is recognized in real time by a Turing machine with $k+1$ tapes and by none with $k$ \citep{aanderaa1974k}; the proof was reduced to its essentials with Kolmogorov complexity \citep{paul1981information}, and simulating $k+1$ tapes online with $k$ takes more than linear time \citep{paul1982online,paturi1990milking}. Through (i) and (ii) the hierarchy holds for clocks moved one reading at a time and read at offset $0$: clocks are a resource that no controller replaces. Whether sorting $k$ kinds of symbol in step with the input is an instance is open; no trained machine with fewer than $k-1$ clocks fits it (Section~\ref{sec:exp_teach}).

\emph{Online multiplication.} A machine multiplies online if it emits the $k$th bit of the product, lowest first, before it reads the $(k+1)$th bits of the factors. Every multitape Turing machine that does so takes time at least of order $N\log N$ on factors of $N$ bits \citep{cook1969minimum,paterson1974improved}, more than real time. By (iii), no machine of Eq.~\eqref{eq:machine} multiplies online one step per input symbol, whatever its number of clocks. Given steps to think, the compiled schoolbook multiplier computes every product it was tested on exactly, up to length~\res{C.think.maxn} (Appendix~\ref{app:tm_exp}).

\emph{What bounded experience fixes.} Fix bounds on the number of controller states and of distinct records. By Theorem~\ref{thm:hole}(ii) there is a size $N_{\rm teach}$ such that every machine of Eq.~\eqref{eq:machine} within these bounds, with unbounded clocks, that is correct on all inputs up to $N_{\rm teach}$ is correct on all inputs. By (i) every multitape machine that runs in real time is such a machine for suitable bounds, so bounded experience fixes in a learner of this kind every procedure computed in real time, and by (iii) such a learner computes nothing else.

\emph{Precision and symmetry.} The universal recurrent network of \citet{siegelmann1995computational} keeps a stack in one rational activation, $q=\sum_i a_i4^{-i}$ with $a_i\in\{1,3\}$, and pushes and pops by affine maps; each stored symbol costs about two bits of precision. The universal Transformer of \citet{perez2021attention} uses positional encodings that contain $n$, $1/n$ and $1/n^2$ at arbitrary precision; with fixed precision neither construction is universal. Constructions with bounded precision keep the unbounded part in a memory addressed by position, as the neural Turing machine does with its graded addresses \citep{graves2014neural} and a looped Transformer with binary addresses of fixed width \citep{giannou2023looped}. In every case the unbounded part of the computation is a quantity that takes more values on longer inputs, a precision or an address, that is a coordinate, and Theorem~\ref{thm:hole}(i) applies to any learner that reads it. In Eq.~\eqref{eq:machine} the unbounded part is the position of a clock along the orbit of its group, which the controller never reads.

%% file: app_neural.tex
\section{The Machine in Neural Dynamics}
\label{app:neural}

This appendix supports Section~\ref{sec:cann} and the experiment of Section~\ref{sec:exp_neural}. It states the network of Eq.~\eqref{eq:cann} in full, gives the maps that join it to the machine, proves Theorem~\ref{thm:neural}, and describes how place cells store records and how a trained controller runs on the network (\texttt{code/orrery/neural.py}).

\subsection{The network}
\label{app:neural_model}
A cycle of period $p$ is a ring of $N=Kp$ rate neurons, $i\in\Z_N$, with $K$ neurons for each state of the cycle, at positions $\theta_i=2\pi i/N$. Its dynamics is Eq.~\eqref{eq:cann}, the continuous attractor network with firing rate adaptation of \citet{mi2014spike} and \citet{chu2024firing}, with divisive normalization as in \citet{wu2008dynamics}, written for discrete neurons ($u_i=U(\theta_i)$, $v_i=V(\theta_i)$, the density absorbed into $J$) and with the external input made explicit:
\begin{equation}
\begin{aligned}
\tau\dot u_i &= -u_i + \textstyle\sum_j J(i-j)\,r_j + \nu(t)\sum_j J'(i-j)\,r_j - v_i + \epsilon\cos\tfrac{2\pi i}{K} + A(t)\,\eta_i + \sigma\xi_i(t),\\
\tau_v\dot v_i &= -v_i + m\,u_i,\qquad r_i = \frac{[u_i]_+^2}{1+k\sum_j [u_j]_+^2},
\end{aligned}
\label{eq:cann_full}
\end{equation}
with the translation invariant kernel $J(d)=\frac{J_0}{\sqrt{2\pi}a}e^{-d^2/2a^2}$ and its asymmetric companion $J'(d)=\frac{d}{a^2}J(d)$, both cut at $|d|\le8$, and independent white noise $\xi_i$. Since $\sum_j J'(i-j)r_j\approx -J_0\,\partial_i r$, a positive gain $\nu$ adds to the recurrent input a copy of the activity shifted forward, and the bump moves forward at a speed proportional to $\nu$ \citep{zhang1996representation}. A move of $n$ wells is a pulse $\nu(t)=\nu_0\,\mathrm{sign}(n)$ for $0\le t<|n|T_1$, after which the network settles until the step ends at $2T_1+T_s$. The reset \home{} is the input $A(t)=A_{\rm home}$ for $t<T_{\rm home}$, with profile $\eta_i=e^{-d(i,0)^2/2a^2}$ centred on well $0$, where $d$ is the distance on the ring. The bias $\epsilon\cos(2\pi i/K)$ has period $K$, so a ring has $p$ wells, at the neurons $i=Kb$, $b\in\Z_p$. A clock is $M$ rings of periods $p_1,\dots,p_M$ that receive the same pulses and resets. At the start every ring holds a bump at well $0$ and $v=0$. The equations are integrated by the Euler method with step $0.1\tau$, so that one step of the machine is $140$ steps of integration. Table~\ref{tab:ringparams} lists the parameters.

\begin{table}[ht]
\caption{\textbf{Parameters of the network.} The inhibition $k_c=J_0^2/(8\sqrt{2\pi}a)$ is the largest that sustains a bump without bias \citep{wu2008dynamics}. A pulse moves every bump to the intended well for every gain $\nu_0$ from about $4.2$ to $4.8$; the value used lies inside this range.}
\label{tab:ringparams}
\centering\small
\begin{tabular}{@{}llll@{}}
\toprule
Symbol & Meaning & Value & Code\\
\midrule
$K$ & neurons per well & $16$ & \texttt{K}\\
$a$ & width of the recurrent kernel (neurons) & $2$ & \texttt{width}\\
$J_0$ & recurrent strength & $1$ & \texttt{J0}\\
$k$ & global inhibition & $0.5\,k_c$ & \texttt{k\_frac}\\
$\epsilon$ & depth of the bias that makes the wells & $1$ & \texttt{eps}\\
$\nu_0$, $T_1$ & gain of a move pulse, duration per well & $4.6$, $4\tau$ & \texttt{speed}, \texttt{T1}\\
$T_s$ & settling time after the longest pulse & $6\tau$ & \texttt{Ts}\\
$A_{\rm home}$, $T_{\rm home}$ & strength and duration of the reset & $30$, $4\tau$ & \texttt{A\_home}, \texttt{T\_home}\\
$\tau_v$, $m$ & time constant and strength of adaptation & $50\tau$, $0$ & \texttt{tau\_v}, \texttt{m}\\
$\sigma$ & noise on every neuron (per $\sqrt\tau$) & $0$ & \texttt{sigma}\\
\bottomrule
\end{tabular}
\end{table}

\subsection{The maps between the two descriptions}
\label{app:neural_maps}
Two maps and the Chinese remainder theorem send a state of the network to a state of the machine. The phase of ring $m$ is the angle of its population vector, in units of wells,
\begin{equation*}
\Pi_m(u)=\frac{p_m}{2\pi}\arg\sum_{i\in\Z_{N_m}} r_i\,e^{2\pi\mathrm{i}\, i/N_m}\in[0,p_m),
\end{equation*}
$Q$ rounds it to the nearest well, $Q(\phi)=\lfloor\phi+\tfrac12\rfloor \bmod p_m$, and the Chinese remainder theorem sends the wells $(b_1,\dots,b_M)$ of a clock to the reading $g\in\Z_L$ with $g\equiv b_m \pmod{p_m}$ for every $m$. Place cells read a fourth map, the activity pooled over each well, $P_m(u)_b=\sum_{|i-Kb|<K/2} r_i$, normalized on each ring and concatenated over the rings into a vector $\hat P(u)$ of dimension $p_1+\dots+p_M$. The group acts on these vectors by rotating every ring: $(R_d\hat P)_{m,b}=\hat P_{m,b-d}$.

\subsection{Proof of Theorem 4}
\label{app:neural_proof}
The proof uses the standard reduction of a continuous attractor network to the motion of its bump along the direction in which the network is neutral \citep{amari1977dynamics,wu2008dynamics,fung2010moving}; part (ii) and the two further properties stated after the proof (noise and wells; adaptation and travelling bumps) rest on the analyses cited there, and the experiments of Appendix~\ref{app:neural_exp} check every part in the network used.

\begin{proof}[Proof of Theorem~\ref{thm:neural} (sketch)]
(i) Let $R$ rotate a ring by $K$ neurons, $(Ru)_i=u_{i-K}$. The kernels $J$ and $J'$ depend only on $i-j$; the normalization depends on $\sum_j[u_j]_+^2$, which $R$ preserves; the bias depends on $i$ modulo $K$; adaptation is local; and the noise has the same law after rotation. So the vector field of Eq.~\eqref{eq:cann_full} commutes with $R$ for every gain $\nu(t)$. Rotating ring $m$ by $g$ wells modulo $p_m$, for every $m$, is the action of $g\in\Z_L\cong\Z_{p_1}\times\cdots\times\Z_{p_M}$, and it commutes with the driven network because the pulses reach every ring alike. The maps are equivariant: $\Pi_m(Ru)=\Pi_m(u)+1$ modulo $p_m$, and $\hat P(Ru)=R_1\hat P(u)$. The reset is centred on well $0$; rotating its profile $\eta$ with the rings, so that \home{} sets well $g$ instead of well $0$, keeps every term equivariant, as the shift invariance of Section~\ref{sec:range} (Appendix~\ref{app:range}) shifts the state that \home{} sets with the clocks. With the reset held at well $0$, \home{} is a return to a fixed state and lies outside the group (Appendix~\ref{app:cycles}).

(ii) Without bias and with $0<k<k_c$, a ring has a continuous family of stable bumps, one at every position, and translation is its only neutral direction \citep{amari1977dynamics,wu2008dynamics}; adaptation below $\tau/\tau_v$ keeps the resting bump stable \citep{mi2014spike,li2025dynamics}. A weak bias acts on a bump at position $z$ through $\sum_i\epsilon\cos(2\pi i/K)\,r_i(z)$, which by translation is proportional to $\cos(2\pi z/K)$. Along the neutral direction the bump therefore moves in a periodic potential $U(z)=-c\,\epsilon\cos(2\pi z/K)+O(\epsilon^2)$, where $c>0$ is the Fourier coefficient of the bump at the period of the bias. Its minima are the $p$ wells, and the basin of each well is the interval of width $K$ around it. A pulse of gain $\nu_0$ and duration $|n|T_1$ carries the bump by about $nK$ neurons; any landing point inside the basin of the well $n$ places away relaxes to that well during the settling time, so the move is exact over an interval of gains (Table~\ref{tab:ringparams}). The strong reset input carries every bump into well $0$. On its stable states the network is therefore conjugate to the clock: after a step $F_\Delta$ with move $\Delta$, the reading of the network is $\Delta$ applied to the reading before, $Q\circ\Pi\circ F_\Delta=\Delta\circ Q\circ\Pi$, with the wells of all rings read as one element of $\Z_L$ through the Chinese remainder theorem.

(iii) A place cell $\ell$ recruited at moment $s$ stores $\hat P_\ell=\hat P(s)$ and receives $o_\ell(t)=\langle\hat P(t),\hat P_\ell\rangle=\sum_m\langle\hat P_m(t),\hat P_m(s)\rangle$. A settled bump of width $a=2$ lies within the $K=16$ neurons of its well, so $\hat P_m$ is close to the indicator of that well, and each term is close to $1$ when ring $m$ stands in the same well at $t$ and at $s$ and close to $0$ otherwise. The cell fires when $o_\ell>M-j-\tfrac12$, that is when at least $M-j$ rings agree. Rotations are orthogonal, so $\langle R_g\hat P(t),R_g\hat P_\ell\rangle=o_\ell(t)$: the response depends only on the difference between the present and the recorded reading, and a head at offset $d$, which presents $R_d\hat P$ of its query clock, finds the cell recruited where the key clock stood $d$ wells ahead. With $j=0$ two readings activate the same cell exactly when all rings agree, that is when they are equal in $\Z_L$. With $j>0$, readings $t$ and $t'$ activate the same cell when they agree on some set $S$ of $M-j$ rings, that is when $t\equiv t'$ modulo $\prod_{m\in S}p_m$. On a clock that advances one well per step the first such coincidence comes after the product of the $M-j$ smallest periods, which is then the number of cells recruited and the range.

\end{proof}

\textbf{Two further properties} (Section~\ref{sec:cann}).
\emph{Noise and wells.} With noise of strength $\sigma$ on every neuron, the projection on the neutral direction gives the bump position an overdamped Langevin motion in $U$ with a diffusion constant proportional to $\sigma^2$, the other directions being stable \citep{fung2010moving}. The rate of escape over a barrier $\Delta U$ is of order $e^{-\Delta U/D}$ with $D\propto\sigma^2$ \citep{kramers1940brownian}, and an escape is a slip of one well. Without bias $U$ is flat and every position is a resting state. The position then diffuses, $\langle(z_t-z_0)^2\rangle=2Dt$, and the reading becomes wrong once $|z_t-z_0|$ exceeds $K/2$, after a time of order $K^2/D\propto\sigma^{-2}$; a move whose displacement differs from $K$ by a small amount leaves an error that nothing corrects and the next moves add to, so a clock without wells errs even without noise.

\emph{Adaptation and travelling bumps.} Without input and without bias, the resting bump loses stability at $m=\tau/\tau_v$ and a travelling bump appears, with speed $v_{\rm int}=\frac{2a}{\tau_v}\big(\frac{m\tau_v}{\tau}-\sqrt{\frac{m\tau_v}{\tau}}\big)^{1/2}$, which rises from zero as the square root of $m-\tau/\tau_v$ \citep[Eq.~10]{mi2014spike}. A bump that follows an input moving slowly at speed $v_{\rm ext}$ is displaced from it by approximately $A_u\tau_v v_{\rm ext}(m-\tau/\tau_v)/\alpha$, where $A_u$ is the bump's amplitude and $\alpha$ the input's strength \citep[Sec.~4.1]{mi2014spike}: a lag below $\tau/\tau_v$ that adaptation shortens, and above it a lead whose time is nearly independent of the input's speed, as in the anticipatory head direction cells of the anterior thalamus \citep{blair1995anticipatory}. With a bias, a travelling bump crosses the barriers of $U$ only if the drive that adaptation exerts exceeds their slope, so travel needs adaptation above a value that grows with the depth of the wells.

\subsection{Place cells and records}
\label{app:neural_catalogue}
The catalogue has a pool of place cells shared by the clocks and, for every head $i$, a Hebbian memory $A_i$ from place cells to records with an occupancy trace $O_i$. When a moment is to be recorded, every clock whose pattern $\hat P$ no cell answers recruits one cell, whose afferent weights become that pattern; this is plasticity in one trial, on the time scale of behaviour \citep{bittner2017behavioral,milstein2021bidirectional}. Let $\alpha(x)$ be the place code of a pattern $x$: the indicator of the cells that fire, divided by the square root of their number. The record $w_t$ is written for every head whose gate accepts it, under the present pattern $\hat P^{c_i}_t$ of the head's key clock, by the delta rule
\begin{equation*}
A_i\leftarrow A_i+\alpha\,(w_t-\alpha^\top A_i)^\top,\qquad O_i\leftarrow O_i+\alpha\,(1-\alpha^\top O_i),\qquad \alpha=\alpha(\hat P^{c_i}_t).
\end{equation*}
A read by head $i$ presents $R_{\delta_i}\hat P^{q_i}_t$ and returns $\alpha^\top A_i$, and whether $\alpha^\top O_i>\tfrac12$. With exact coincidence one cell fires, a write replaces what that cell held, and a read returns the last record made at the matching reading: the catalogue of Eq.~\eqref{eq:machine}. Records of the trained machines are vectors of $16$ numbers; records of the reference controllers are stored as random unit vectors of dimension $32$ and decoded as the nearest such vector.

\subsection{The neural executor}
\label{app:neural_executor}
A trained controller runs on the network in the order of the symbolic machine. At step $t$ it chooses the moves of every clock from its state and the present symbol; the clocks are driven for one step; for $t>0$ every head reads the catalogue; the controller updates its state from the symbol, the records and the found signals, emits its output, and computes a record and the gates; cells are recruited for the present patterns; and the record is written. The controller receives no reading, phase or activity of the clocks, only records and whether one was found. Symbolic clocks driven by the same moves are kept alongside, and a step at which $\mathrm{CRT}\circ Q\circ\Pi$ differs from them counts as a slip. A step of a clock of rings $2,3,5,7,11$ costs $140$ steps of integration of $448$ neurons.

%% file: app_reference.tex
\section{Reference Controllers}
\label{app:reference}

This appendix gives the controllers we wrote by hand for Eq.~\eqref{eq:machine}, one for every benchmark task whose procedure keeps pace with the input (\texttt{code/orrery/reference.py}); they are the programs of Section~\ref{sec:disciplines} written out, and Section~\ref{sec:exp_bench} compares the trained machines with them. These controllers show that each procedure exists inside the machine (Proposition~\ref{prop:disciplines}), they give the number of clocks our procedure uses for each task (Table~\ref{tab:bench}) and the closure size $N^\ast$ (Lemma~\ref{lem:closure}), and their runs are drawn in Figure~\ref{fig:concept}b and~e. Each is exact: on $50$ random inputs at each of the lengths $1$, $2$, $3$, $5$, $8$, $13$, $20$, $40$, $41$, $100$, $300$ and on $8$ inputs of length $1{,}000$ every output symbol is correct (\texttt{code/experiments/e0\_reference.py}). The controllers use moves in $\{-2,\dots,2,\home\}$ and heads at offsets in $\{-1,0,1,2\}$; a head's gate selects records by their kind (an input symbol, an output step, a pushed symbol, a boundary). A controller holds a few discrete variables and never receives a reading. Table~\ref{tab:reference} lists the clocks, the heads, the number of distinct situations each controller meets, and the size at which that number stops growing.

\begin{table}[h]
\caption{\textbf{Reference controllers.} $C$: clocks; $R$: heads; situations: distinct combinations of controller state, input symbol and records found, met on $200$ random inputs at every length up to $60$; $N^\ast$: the length after which no new situation appears.}
\label{tab:reference}
\centering\small
\begin{tabular}{@{}lccrrl@{}}
\toprule
Task & $C$ & $R$ & situations & $N^\ast$ & procedure\\
\midrule
Even pairs & \res{E0.C.even_pairs} & \res{E0.R.even_pairs} & \res{E0.sit.even_pairs} & \res{E0.N.even_pairs} & compare the first and the last symbol\\
Modular arithmetic & \res{E0.C.modular_arithmetic} & \res{E0.R.modular_arithmetic} & \res{E0.sit.modular_arithmetic} & \res{E0.N.modular_arithmetic} & running value, product and sign\\
Parity & \res{E0.C.parity_check} & \res{E0.R.parity_check} & \res{E0.sit.parity_check} & \res{E0.N.parity_check} & running parity\\
Cycle navigation & \res{E0.C.cycle_navigation} & \res{E0.R.cycle_navigation} & \res{E0.sit.cycle_navigation} & \res{E0.N.cycle_navigation} & running position on a cycle of five\\
Stack manipulation & \res{E0.C.stack_manipulation} & \res{E0.R.stack_manipulation} & \res{E0.sit.stack_manipulation} & \res{E0.N.stack_manipulation} & a clock follows the depth; walk down to answer\\
Reverse string & \res{E0.C.reverse_string} & \res{E0.R.reverse_string} & \res{E0.sit.reverse_string} & \res{E0.N.reverse_string} & forward while listening, back while answering\\
Bracketed arithmetic & \res{E0.C.modular_arithmetic_brackets} & \res{E0.R.modular_arithmetic_brackets} & \res{E0.sit.modular_arithmetic_brackets} & \res{E0.N.modular_arithmetic_brackets} & a clock follows the depth; a return finds the context\\
Solve equation & \res{E0.C.solve_equation} & \res{E0.R.solve_equation} & \res{E0.sit.solve_equation} & \res{E0.N.solve_equation} & as brackets, with values linear in $x$\\
Duplicate string & \res{E0.C.duplicate_string} & \res{E0.R.duplicate_string} & \res{E0.sit.duplicate_string} & \res{E0.N.duplicate_string} & \home{} and replay, twice\\
Missing duplicate & \res{E0.C.missing_duplicate_string} & \res{E0.R.missing_duplicate_string} & \res{E0.sit.missing_duplicate_string} & \res{E0.N.missing_duplicate_string} & two clocks meet the partner of the gap\\
Odds first & \res{E0.C.odds_first} & \res{E0.R.odds_first} & \res{E0.sit.odds_first} & \res{E0.N.odds_first} & strides of two over odd, then even positions\\
Binary addition & \res{E0.C.binary_addition} & \res{E0.R.binary_addition} & \res{E0.sit.binary_addition} & \res{E0.N.binary_addition} & aligned digit clocks; a clock of ones finds the end\\
Bucket sort & \res{E0.C.bucket_sort} & \res{E0.R.bucket_sort} & \res{E0.sit.bucket_sort} & \res{E0.N.bucket_sort} & one counting clock per boundary between symbols\\
\bottomrule
\end{tabular}
\end{table}

\textbf{No clock.} Parity, even pairs, cycle navigation and modular arithmetic are regular: the controller keeps a running value (a parity bit, the first symbol, a position on a cycle of five, or the value, pending product and sign of an expression modulo $5$) and needs no memory of the past. Modular arithmetic has the largest controller of these and the largest $N^\ast$, $15$, because the combinations of value, product, sign and operator are met only on inputs of some length.

\textbf{Reversal.} One clock advances on every input symbol and walks back on every output step. The head with offset $+1$ returns, at output $k$, the input recorded at reading $n-k+1$. The controller holds whether the input has ended; its whole repertoire is met by inputs of length $2$.

\textbf{Duplication.} One clock advances on every input symbol and goes \home{} at the first output step, then advances again, so that output $k$ reads the input recorded at reading $k$. A second head looks two readings ahead; when it finds nothing, the copy has reached the end of the input, and the clock goes \home{} again for the second copy.

\textbf{Stack manipulation.} The input is an initial stack followed by \textsc{push} and \textsc{pop} actions, and the answer is the final stack read from the top. A clock follows the depth: every pushed symbol moves it forward, and a \textsc{pop} moves it back unless the stack is empty, which the controller learns from the head at offset $0$ finding no pushed symbol at the present reading. To answer, the clock walks down and reads the symbol last pushed at each depth; an empty read ends the answer.

\textbf{Bracketed arithmetic and equations.} A clock follows the nesting depth: an opening bracket moves it forward, a closing bracket back. The controller holds the value of the innermost open expression with its pending operator and sign, and records it at every step. At a closing bracket the clock returns to the enclosing depth, and the head at offset $0$ returns the last record made there, the moment just before the matching opening bracket, which holds the state of the enclosing expression; the controller combines it with the value just completed. This is a call and its return (Section~\ref{sec:disciplines}). For equations the values are linear forms $ax+b$ modulo $5$, and the controller solves for $x$ at the end. Many combinations of value, operator and depth make $N^\ast$ large, $29$ and $49$.

\textbf{Odds first.} One clock counts the input. At the first output step both clocks go \home{}; one advances by two per step and reads the symbols at odd positions, and when it finds nothing the other advances by two from the first position and reads the symbols at even positions.

\textbf{Missing duplicate.} The input is a string $ww$ with one symbol replaced by a gap at position $m$, and the answer is the missing symbol. Over a binary alphabet the missing symbol is the parity of the visible ones, and a running parity solves the task without a clock, which is why the LSTM and the Transformer solve it (Table~\ref{tab:bench}). The controller of Table~\ref{tab:reference} reads the partner of the gap, which works over any alphabet. Clock~$0$ counts the input. Clock~$1$ moves by $z_t-\eta_t$ and clock~$2$ by $z_t+\eta_t$, where $z_t=1$ up to and including the gap and $\eta_t=1$ on every second symbol. At the end of an input of length $2h$ clock~$1$ reads $m-h$ and clock~$2$ reads $m+h$, the positions of the partner of the gap in either half, and exactly one of the two heads keyed by clock~$0$ finds a symbol there. The controller holds whether the gap has been seen and one parity bit. The readings it compares span about $3n/2$, so its range ends near length $2L/3$.

\textbf{Binary addition.} The input is $a{+}b$, each number written lowest bit first, and the answer is $a+b$ followed by an end mark. Clock~$A$ counts the digits of $a$, clock~$B$ those of $b$, and clock~$W$ the ones in both. At the first output step $A$ and $B$ go \home{} and the moment is recorded as the boundary, keyed by the final reading of $W$. From then on $A$ counts output positions and reads $a_k$ and $b_k$ through heads keyed by $A$ and by $B$, while $B$ counts the ones consumed so far; the head with query $B$ and key $W$ finds the boundary once all the ones have been consumed. The sum ends at the first position where all the ones have been consumed and nothing is carried. The controller holds the carry, the number of ones read at the last step and one flag.

\textbf{Bucket sort.} The input is a string over $k=5$ symbols and the answer is the same string sorted. Clock $c<k-1$ advances on every symbol at most $c$, so that at the end of the input it reads the prefix count $S_c$, the first output position of a symbol larger than $c$. At the first output step the clocks stay and the moment is recorded as the boundary; then clock~$0$ goes \home{} and counts output positions, and head $c$ finds the boundary when the count meets $S_c$. The symbol emitted is the number of boundaries passed. The $k-1$ boundaries are independent functions of the counts, and each is carried by one clock (Appendix~\ref{app:programs}).

%% file: app_methods.tex
\section{The Trained Machine and Its Training}
\label{app:methods}

This appendix describes the machine that Section~\ref{sec:experiments} trains: its clocks, heads, records and controller, the discrete machine run at test, the relaxation that gives training its gradients, the schedule, the data and the restart ladder, the baselines and the compute. Every experiment of Appendix~\ref{app:experiments} uses this machine and this protocol unless its design says otherwise.

\subsection{Architecture}
\label{app:architecture}
\textbf{Clocks.} Each of the $C=3$ clocks is a set of rings of periods $2,3,5,7,11$, each ring carrying one bump: $28$ units per clock and $L=2{,}310$ states. A move is one of $\{-2,-1,0,+1,+2,\home\}$; it rolls every ring of the clock by the same step, or sets every ring to phase $0$.

\textbf{Heads, gates and records.} There is one head for every ordered pair of a query clock and a key clock and every offset in $\{-1,0,1,2\}$, $36$ heads in all. At every step the controller writes a record of $16$ numbers, a linear function of its new state and the embedding of the present symbol, keyed by the readings of all clocks after the move of that step. Each head has a gate, a linear function of the same quantities with one output per head, which decides whether the record may be returned by that head. Head $i$ returns the most recent earlier record whose key reading on clock $c_i$ equals the reading of clock $q_i$ shifted by $\delta_i$, and which its gate accepts; it also returns one bit saying whether such a record exists.

\textbf{Controller.} The controller is a gated recurrent cell with $256$ units. At step $t$ it (1) chooses the moves of all clocks, through a linear map of its state $h_{t-1}$ and the embedding of the symbol $x_t$, with an initial bias of $2$ toward the move $+1$, so that untrained clocks count time; (2) the clocks move; (3) the heads read; (4) the cell updates its state from the symbol embedding and the $36$ records and found bits, which enter scaled by $1/\sqrt{36}$ so that together they weigh as much as one input; (5) the output, read from the new state at every output step, and the record of the moment are emitted. The input is the string followed by one placeholder symbol for every output position, as in \citet{deletang2023neural}. Nothing in the controller's input encodes the position or the length.

\subsection{The discrete machine at test}
\label{app:executor}
At test every clock is in one state, moves are the most probable ones, a coincidence is exact on every ring, a gate accepts when its logit is positive, and the most recent accepted coincidence wins. The executor \texttt{code/orrery/fast.py} runs this machine with integer clocks modulo $L$ and, for every head, a dictionary from key readings to the most recent accepted record. It reproduces the test pass of the network exactly and costs time linear in the length, so that inputs of $15{,}000$ symbols are cheap. Exchanging the clocks changes the modulus and nothing else; unbounded clocks are integers without a modulus.

\subsection{The relaxation used in training}
\label{app:relaxation}
Discrete moves and exact matches have no gradients. Training uses a relaxation at temperature $T$ whose limit $T\to0$ is the discrete machine, and ends with the discrete machine itself.

\emph{Moves.} Each clock is a distribution over its states, held as one bump per ring, and is moved by the mixture $\sum_a p_a P_a$ of the six moves $a$, with probabilities $p=\mathrm{softmax}(\ell/T)$ of the move logits $\ell$.

\emph{Reads.} With $q$ the query bump shifted by the head's offset, $k_s$ the key bump of moment $s$, $G_s$ the head's gate logit for that record, and $M=5$ rings, the head returns a weighted average of the records $w_s$ of the past moments $s<t$, with weights
\begin{equation*}
\zeta_s=\mathrm{softmax}_s\Big(\big[\beta\,\big(\textstyle\sum_m\langle q_m,k_{s,m}\rangle-M\big)+\rho\,(s-t)\big]/T+4\log\sigma(G_s/T)\Big),\qquad \beta=8,\ \rho=0.05 .
\end{equation*}
The first term counts the rings on which the two bumps agree, the second favours recent moments, and the third is the gate. As $T\to0$ the bumps become single states, agreement becomes all or nothing, the gate becomes a threshold, and the most recent accepted coincidence takes all the weight. The found bit is the largest weight before normalization, without the recency term.

\emph{Calibration.} When no record coincides exactly, the weighted average still returns the nearest misses at full strength, whereas the discrete read returns nothing. The calibrated read multiplies the average by $\pi$, the probability that some accepted record coincides exactly, $\pi=1-\prod_s(1-\pi_sg_s)$ with $\pi_s=\prod_m\langle q_m,k_{s,m}\rangle$ and $g_s=\sigma(G_s/T)$; $\pi$ is then also the found bit. Training uses annealed calibration, with factor $\pi^{1-T}$ and the found bit interpolated in the same way: the read is plain at $T=1$, where it is easiest to learn, and near misses fade as the temperature falls.

\emph{Discrete phase.} In the last twelfth of training every move is the most probable one, applied exactly, with the gradient of the relaxed choice passed straight through \citep{bengio2013estimating}, and every read is calibrated. The machine trained is then the machine tested, up to the gates, which remain soft.

\subsection{Schedule, data and selection}
\label{app:schedule}
A run has $12{,}000$ steps. The temperature is $T=1$ up to step $8{,}000$, falls linearly to $0.05$ at step $10{,}750$, and the discrete phase runs from step $11{,}000$ to $12{,}000$. The optimizer is Adam with step size $10^{-3}\,T$, falling with the temperature, and $5\cdot10^{-5}$ in the discrete phase; the batch has $32$ inputs of one length drawn uniformly from $1$ to $40$; gradients are clipped to norm $1$; the loss is the cross entropy summed over output symbols. Every $500$ steps the discrete machine is run on $64$ validation inputs of each of the lengths $31$, $34$, $37$ and $40$; the checkpoint kept is the one with the best accuracy there, the later one on ties.

\textbf{Noise on the controller.} Gaussian noise is added to the controller's state at every step of training, after the outputs and the record of the step are computed. The controller can fail in two opposite ways. Without noise it can learn procedures with many states, but it can also fit short inputs by holding them in its state, which does not extend to long ones. With noise it cannot hold fine graded information, so it keeps its past in the records, which the noise does not touch, and it learns procedures with many states slowly. The standard schedule therefore starts without noise, so that the controller can find the procedure, and raises the standard deviation linearly from $0$ at step $7{,}000$ to $0.5$ at step $8{,}000$, before the temperature falls, so that only distinctions that survive noise remain.

\textbf{Restarts.} A run whose best validation accuracy is below $99\%$ is repeated with a new initialization and data order, up to six attempts, along a fixed ladder that is the same for every task (Table~\ref{tab:ladder}); the first attempt that reaches $99\%$ is kept, and otherwise the one with the best validation accuracy. Of the \res{E1.restarts.n} benchmark runs, \res{E1.restarts.first} fit at the first attempt, \res{E1.restarts.later} after restarts, and \res{E1.restarts.none} never; these are the \res{E1.restarts.none} runs on the five tasks on which no run fits (Table~\ref{tab:bench}). The attempt kept was noisy throughout in \res{E1.restarts.noisy} runs and used the long schedule in \res{E1.restarts.long}. All retrained machines of Section~\ref{sec:experiments} use the same ladder.

\begin{table}[h]
\caption{\textbf{The restart ladder.} The long schedule has $20{,}000$ steps: $T=1$ up to step $13{,}500$, falling to $0.05$ at step $18{,}500$, and the discrete phase from step $19{,}000$.}
\label{tab:ladder}
\centering\small
\begin{tabular}{@{}cll@{}}
\toprule
Attempt & Schedule & Noise on the controller\\
\midrule
1, 3 & standard, $12{,}000$ steps & none up to step $7{,}000$, rising to $0.5$ at step $8{,}000$\\
2, 4 & standard, $12{,}000$ steps & $0.5$ throughout\\
5 & long, $20{,}000$ steps & none up to step $8{,}000$, rising slowly to $0.5$ at step $13{,}500$\\
6 & long, $20{,}000$ steps & $0.5$ throughout\\
\bottomrule
\end{tabular}
\end{table}

\subsection{Training without calibration and without the discrete phase}
\label{app:plain}
The relaxation of Appendix~\ref{app:relaxation} hardens into the discrete machine by two means, the calibrated read and the discrete phase. Without them, with the plain read throughout, a constant step size, no discrete phase and three attempts, the relaxed machine and the discrete machine can come apart: in \res{A.plain.gap} of \res{A.plain.att} training attempts the relaxed machine is correct on at least $99\%$ of the symbols at length $40$ while the discrete machine is not. The discrete machines that fit still extrapolate (\res{A.plain.fitexact} of \res{A.plain.fit}; the exception is \res{A.plain.other}), and the machine trained this way solves \res{A.plain.exact} tasks (Table~\ref{tab:plain}).

\restable{tab:plain}

\subsection{Baselines}
\label{app:baselines}
Table~\ref{tab:bench} quotes the published results of \citet{deletang2023neural} for the LSTM, Stack-RNN and Tape-RNN (best of $10$ seeds and three learning rates) and the best Transformer of \citet{ruoss2023randomized} over the eleven positional encodings they compare; their randomized encodings draw training positions from $1$ to $2{,}048$, so that every test position has been seen in training. To compare training under one protocol, we also trained an LSTM with $256$ units (step size $10^{-3}$) and Transformer encoders with five layers, width $64$ and eight heads, with rotary positions or none (step size $3\cdot10^{-4}$), on the same data, with the same number of steps, batch, loss and selection on validation lengths $31$ to $40$, and with up to three attempts under the same validation rule (Table~\ref{tab:baselines}). These runs give the comparison of Section~\ref{sec:exp_bench}: of the runs that fit their training lengths, \res{E1.lstm.k} of \res{E1.lstm.m} LSTM runs and \res{E1.tf.k} of \res{E1.tf.m} Transformer runs extrapolate.

\subsection{Compute}
\label{app:compute}
The machines were trained on two NVIDIA A100 GPUs with up to $40$ jobs at a time, one CPU thread each; the machines trained on at most three and five symbols, the exact executor and the analyses ran on CPUs. A benchmark run of the machine, with its restarts, took a median of \res{E1.compute.median} hours, and the \res{E1.restarts.n} benchmark runs \res{E1.compute.total} hours in total; the \res{E1.compute.nbase} baseline runs took \res{E1.compute.base} hours. The \res{E1.compute.jobs} jobs of the machine experiments (benchmark, the experiments on range, readings, experience and clocks, and the analysis of dynamics) ran in \res{E1.compute.wall} hours of wall time and \res{E1.compute.jobh} job hours. The \res{N.compute.jobs} jobs of the experiments in neural dynamics ran on $24$ CPU threads in \res{N.compute.wall} hours, and the \res{H.compute.jobs} training runs on the number of clocks on the two GPUs in \res{H.compute.wall} hours.

%% file: app_experiments.tex
\section{The Experiments in Detail}
\label{app:experiments}

This appendix first describes the tasks and the measures. Each later subsection describes one experiment of Section~\ref{sec:experiments} in four paragraphs: why it was run, its design, its results, and what they mean for the theory. Appendices~\ref{app:bench} and~\ref{app:dynamics} expand Section~\ref{sec:exp_bench}, Appendix~\ref{app:range_exp} Section~\ref{sec:exp_range}, Appendix~\ref{app:controls} Section~\ref{sec:exp_forbid}, Appendices~\ref{app:teach} and~\ref{app:clocks_exp} Section~\ref{sec:exp_teach}, Appendix~\ref{app:neural_exp} Section~\ref{sec:exp_neural} and Appendix~\ref{app:tm_exp} Section~\ref{sec:exp_tm}. The machine and its training are described in Appendix~\ref{app:methods}.

\subsection{The tasks}
\label{app:tasks}
The fifteen tasks of \citet{deletang2023neural} are listed in Table~\ref{tab:tasks} with one example each, generated by \texttt{code/orrery/tasks.py}. An input is a string of symbols followed by one placeholder for every output symbol; the learner answers one symbol per placeholder. Tasks whose answer is a single symbol have one output step, and tasks whose answer has variable length end it with an end mark ($.$). The score of an output is the fraction of its symbols that are correct, counted over the positions the task marks as part of the answer.

\begin{table}[ht]
\caption{\textbf{The fifteen tasks}, grouped by the level of the Chomsky hierarchy at which \citet{deletang2023neural} place them, with one input and its answer. Numbers are written lowest bit first, except in the square root task. Positions in odds first are counted from zero.}
\label{tab:tasks}
\centering\footnotesize
\setlength{\tabcolsep}{3pt}
\begin{tabular}{@{}lllll@{}}
\toprule
Level & Task & Asked for & Input & Answer\\
\midrule
Regular & Even pairs & whether the first and last symbols differ & \texttt{100001} & \texttt{0}\\
 & Modular arithmetic & the value of an expression modulo $5$ & \texttt{4+2-0-0} & \texttt{1}\\
 & Parity & whether the number of ones is odd & \texttt{000110} & \texttt{0}\\
 & Cycle navigation & the end position on a cycle of five & \texttt{-1 0 0 +1 0 0} & \texttt{0}\\
\midrule
Context free & Stack manipulation & the final stack, read from the top & \texttt{110 PUSH1 PUSH1 PUSH1} & \texttt{111011.}\\
 & Reverse string & the string reversed & \texttt{100111} & \texttt{111001}\\
 & Bracketed arithmetic & the value of a nested expression modulo $5$ & \texttt{((4+(0))-4)} & \texttt{0}\\
 & Solve equation & the $x$ that solves a nested equation modulo $5$ & \texttt{(x+(1))=1} & \texttt{0}\\
\midrule
Context & Duplicate string & the string written twice & \texttt{110001} & \texttt{110001110001}\\
sensitive & Missing duplicate & the symbol hidden in a string written twice & \texttt{0001\_001} & \texttt{0}\\
 & Odds first & odd positions first, then even ones & \texttt{000111} & \texttt{011001}\\
 & Binary addition & the sum of two numbers & \texttt{11101+1} & \texttt{00011.}\\
 & Binary multiplication & the product of two numbers & \texttt{11*0011} & \texttt{001001.}\\
 & Square root & the integer part of the square root & \texttt{010100} & \texttt{100}\\
 & Bucket sort & the string sorted & \texttt{113311} & \texttt{111133}\\
\bottomrule
\end{tabular}
\end{table}

\textbf{Measures.} A run is trained on lengths $1$ to $40$ (or $1$ to $N$ in Section~\ref{sec:exp_teach}) and selected on validation inputs of lengths $31$ to $40$ (or $N-3$ to $N$). It \emph{fits} if the discrete machine is correct on at least $99\%$ of the output symbols of these validation inputs. The \emph{test score} is the mean, over the lengths $41$ to $500$, of the accuracy on $32$ random inputs at each length. A run \emph{extrapolates} if its test score is at least $99\%$, and a task is \emph{solved} if the mean over its three seeds is at least $99\%$.

\subsection{Learning from short inputs}
\label{app:bench}
\textbf{Why.} By Theorem~\ref{thm:hole}(ii), bounded experience fixes the rule of a learner whose decisions read only coincidences, the present symbol and a finite state. The size $N_{\rm teach}$ this needs can be large, but by Lemma~\ref{lem:closure} a run that fits by acting as the procedure on inputs up to its closure size $N^\ast$ extrapolates at every length within the range of its clocks, $2{,}310$; the experiment asks how often training on lengths $1$ to $40$ finds the procedure (Section~\ref{sec:exp_bench}). By Theorem~\ref{thm:hole}(i), a network that reads a coordinate, such as a position or a recurrent state that drifts with length, can fit the same inputs and fail beyond them.

\textbf{Design.} The fifteen tasks of \citet{deletang2023neural} span the Chomsky hierarchy (Appendix~\ref{app:tasks}). On each task the machine of Appendix~\ref{app:methods}, with three clocks of rings $2,3,5,7,11$, is trained with three seeds on lengths $1$ to $40$, with up to six attempts along the restart ladder (Table~\ref{tab:ladder}). Each run is tested on $32$ random inputs at every length from $41$ to $500$ and scored with the measures above. LSTMs and Transformers are trained on the same data and selected on the same validation lengths, with up to three attempts (Appendix~\ref{app:baselines}). Table~\ref{tab:bench} also quotes the best published results of other networks.

\begin{figure}[t]
\centering
\includegraphics[width=\textwidth]{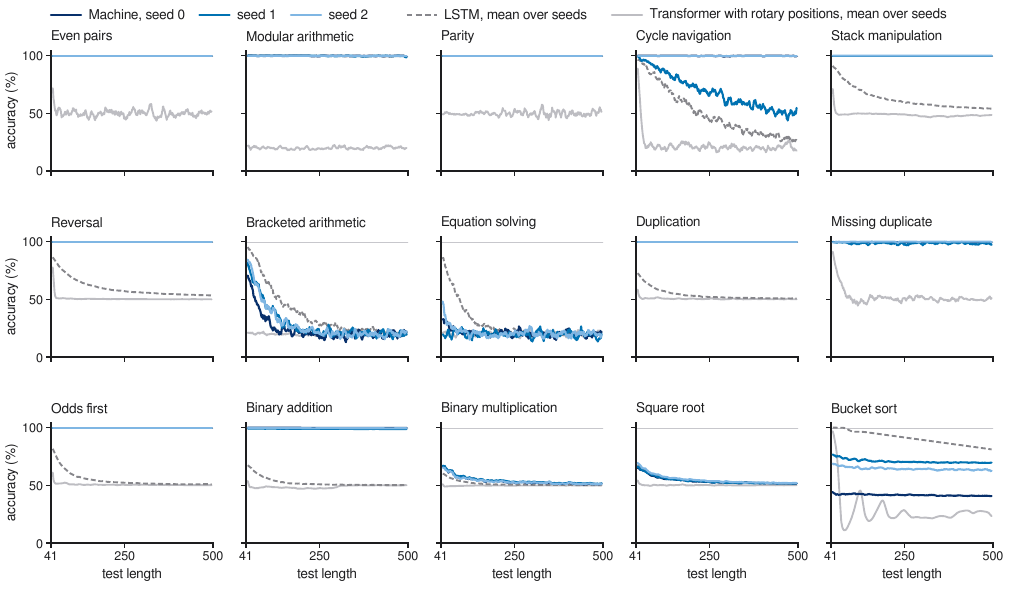}
\caption{\textbf{Accuracy at every test length} (mean over the $32$ inputs of each length, smoothed over ten consecutive lengths). Blue: the three seeds of the machine; dashed gray: LSTM; light gray: Transformer with rotary positions; thin line: $99\%$.}
\label{fig:lengths}
\end{figure}

\begin{table}[ht]
\caption{\textbf{Fifteen tasks, trained on lengths $1$ to $40$ and tested on $41$ to $500$} (\% of correct output symbols), grouped as regular, context free and context sensitive. LSTM, Stack-RNN and Tape-RNN: best of $30$ runs of \citet{deletang2023neural}; the Tape-RNN's head can also jump by the input length. Transformer: best of eleven positional encodings of \citet{ruoss2023randomized}, whose randomized encodings draw training positions from $1$ to $2{,}048$, so that every test position is seen. This work: mean $\pm$ s.d.\ and maximum over three seeds; bold: solved. Clocks: the number our procedure of Appendix~\ref{app:reference} uses; $\times$: no procedure in step with the input is known. Over a binary alphabet a running parity finds the missing duplicate without a clock, which is why the LSTM and the Transformer solve it.}
\label{tab:bench}
\centering\footnotesize
\setlength{\tabcolsep}{5pt}
\begin{tabular}{@{}llrrrrrcc@{}}
\toprule
 & Task & LSTM & Stack-RNN & Tape-RNN & Transformer & \multicolumn{1}{c}{This work} & max & Clocks\\
\midrule
Regular & Even pairs & 100.0 & 100.0 & 100.0 & 100.0 & \textbf{\rest{E1.even_pairs.mean}} & \rest{E1.even_pairs.max} & 0\\
 & Modular arithmetic & 100.0 & 100.0 & 100.0 & 28.1 & \textbf{\rest{E1.modular_arithmetic.mean}} & \rest{E1.modular_arithmetic.max} & 0\\
 & Parity & 100.0 & 100.0 & 100.0 & 52.6 & \textbf{\rest{E1.parity_check.mean}} & \rest{E1.parity_check.max} & 0\\
 & Cycle navigation & 100.0 & 100.0 & 100.0 & 73.6 & \rest{E1.cycle_navigation.mean} & \rest{E1.cycle_navigation.max} & 0\\
\midrule
Context free & Stack manipulation & 59.1 & 100.0 & 100.0 & 77.9 & \textbf{\rest{E1.stack_manipulation.mean}} & \rest{E1.stack_manipulation.max} & 1\\
 & Reverse string & 60.9 & 100.0 & 100.0 & 95.1 & \textbf{\rest{E1.reverse_string.mean}} & \rest{E1.reverse_string.max} & 1\\
 & Bracketed arithmetic & 59.2 & 96.1 & 95.4 & 34.9 & \rest{E1.modular_arithmetic_brackets.mean} & \rest{E1.modular_arithmetic_brackets.max} & 1\\
 & Solve equation & 67.8 & 56.2 & 64.4 & 28.1 & \rest{E1.solve_equation.mean} & \rest{E1.solve_equation.max} & 1\\
\midrule
Context & Duplicate string & 57.6 & 52.8 & 100.0 & 75.1 & \textbf{\rest{E1.duplicate_string.mean}} & \rest{E1.duplicate_string.max} & 1\\
sensitive & Missing duplicate & 54.3 & 55.2 & 100.0 & 100.0 & \textbf{\rest{E1.missing_duplicate_string.mean}} & \rest{E1.missing_duplicate_string.max} & 3\\
 & Odds first & 55.6 & 51.9 & 100.0 & 69.3 & \textbf{\rest{E1.odds_first.mean}} & \rest{E1.odds_first.max} & 2\\
 & Binary addition & 55.5 & 52.7 & 100.0 & 64.5 & \textbf{\rest{E1.binary_addition.mean}} & \rest{E1.binary_addition.max} & 3\\
 & Binary multiplication & 53.1 & 52.7 & 58.5 & 52.2 & \rest{E1.binary_multiplication.mean} & \rest{E1.binary_multiplication.max} & $\times$\\
 & Square root & 57.5 & 56.5 & 57.8 & 53.3 & \rest{E1.compute_sqrt.mean} & \rest{E1.compute_sqrt.max} & $\times$\\
 & Bucket sort & 99.3 & 78.1 & 70.7 & 100.0 & \rest{E1.bucket_sort.mean} & \rest{E1.bucket_sort.max} & 4\\
\bottomrule
\end{tabular}
\end{table}

\restable{tab:perseed}
\restable{tab:baselines}

\textbf{Results.} Table~\ref{tab:bench} gives the scores with the published results of other networks, Table~\ref{tab:perseed} lists every benchmark run of the machine, and Figure~\ref{fig:lengths} shows the accuracy at every test length. The machine solves \res{E1.nexact} tasks, among them duplication, odds first and binary addition, on which LSTMs and Transformers reach at most \res{E1.prev.duplicate_string}\%, \res{E1.prev.odds_first}\% and \res{E1.prev.binary_addition}\% (Figure~\ref{fig:tests}a; Table~\ref{tab:bench}). Of the \res{E1.fit.m} runs that fit, \res{E1.fit.k} extrapolate, and none of these falls below \res{E1.fit.min}\% at any length; their accuracy at length $500$ is their accuracy at length $41$. Of the other three, two keep at every length the few errors they make at the training lengths, \res{E1.near.detail}, and the third, on \res{E1.drift.task} with seed \res{E1.drift.seed}, loses accuracy steadily with length (Figure~\ref{fig:lengths}), from \res{E1.drift.a}\% at lengths $41$ to $90$ to \res{E1.drift.b}\% at lengths $441$ to $490$, as a graded quantity carried from step to step does, which is why a controller has to sit in discrete states (Section~\ref{sec:forbids}). No run fits multiplication, square root or bucket sort, the tasks without a procedure in three clocks, nor bracketed arithmetic or equation solving. Under the same protocol (Table~\ref{tab:baselines}), \res{E1.lstm.k} of the \res{E1.lstm.m} LSTM runs that fit their training lengths extrapolate, against \res{E1.tf.k} of the \res{E1.tf.m} Transformer runs that fit. The LSTM fits the four regular tasks, missing duplicate and bucket sort with every seed and bracketed arithmetic with one, and it extrapolates on all of them except cycle navigation, bucket sort and bracketed arithmetic. The Transformers fit their training lengths on even pairs, cycle navigation, reversal, missing duplicate, odds first and bucket sort, and extrapolate on none. A Transformer that reverses strings of length $40$ without error falls to chance within ten symbols, by length $50$ (Figure~\ref{fig:lengths}), the hole of Theorem~\ref{thm:hole} in its plainest form.

\textbf{What the machines read.} Removing the catalogue, so that every head returns nothing, shows what each machine reads its past through (Table~\ref{tab:bare}). On reversal, duplication, stack manipulation, odds first and binary addition, every machine needs it: without the catalogue the machines answer at most \res{B.past.hi}\% of the symbols of their training inputs correctly, at chance. On the regular tasks, where a finite controller suffices, some machines use the catalogue and some answer as well without it. On bracketed arithmetic, which no run fits, two of the three machines answer nearly as well without the catalogue as with it (\res{B.br.bare}\% against \res{B.br.full}\% for seeds $0$, $1$ and $2$): training found no procedure on the clocks for this task, and what a controller of $256$ units holds of an input by itself (Section~\ref{sec:forbids}) does not extend beyond the training lengths.

\textbf{Reading.} A rule that the machine fits on inputs of up to $40$ symbols holds up to length $500$ in \res{E1.fit.k} of the \res{E1.fit.m} runs that fit: training finds the procedure, and by Lemma~\ref{lem:closure} the procedure then holds at every length within range. Every machine that extrapolates on reversal, duplication, stack manipulation, odds first or binary addition reads its past through the catalogue; over a binary alphabet a running parity finds the missing duplicate, and one such machine answers without it. Networks that read positions fit the same inputs and fail beyond them, which is the hole of Theorem~\ref{thm:hole}(i).

\restable{tab:bare}

\subsection{The range belongs to the clocks}
\label{app:range_exp}
\textbf{Why.} Theorem~\ref{thm:range} says that the range belongs to the clocks (Section~\ref{sec:range}). On every input whose span is below $L$ the machine runs exactly as the same controller with unbounded clocks. The first departure from that run comes only after the span has reached $L$, and new cycles with product $L'$ move the range to $L'$ while the controller stays the same. For reversal the aliasing rule of Appendix~\ref{app:range}, derived after the theorem from the reversal procedure on random bits, predicts the accuracy beyond the range, $(L+\tfrac12(n-L))/n$ at length $n>L$, with no free parameter.

\textbf{Design.} Each trained machine of eight tasks (reversal, duplication, stack manipulation, odds first, missing duplicate, binary addition, bracketed arithmetic and bucket sort) is run without retraining. Its clocks are made of the cycles $\{3,4,5\}$, $\{2,3,5,7\}$, $\{2,3,5,7,11\}$ and $\{3,5,7,11,13\}$, with ranges $L=60$, $210$, $2{,}310$ and $15{,}015$, or they are unbounded integer clocks. The lengths are $L$ times $0.25$, $0.5$, $0.75$, $0.9$, $0.95$, $0.98$, $0.99$, $1$, $1.01$, $1.02$, $1.05$, $1.1$, $1.25$, $1.5$ and $2$, rounded, with $8$ random inputs at each. On every input we record the output with modular and with unbounded clocks, and the span of the readings each head compares. These are the query readings shifted by the head's offset together with the key readings of the records its gate accepts, over the whole run of the unbounded machine. We also record the step at which this span first reaches $L$, and the first step at which some head returns a different record with modular clocks than with unbounded ones (the first misread). Each machine is also run with each of its five cycles removed in turn, which shortens the range to the product of the other four, at lengths $0.5$, $0.95$, $1.05$ and $1.5$ times the shortened range, with $8$ inputs at each. The reversal machines trained on lengths $1$ to $40$, and those trained on lengths $1$ to $3$, $1$ to $5$ and $1$ to $10$ (Appendix~\ref{app:teach}), three of each, are also run with the cycles $11$, $13$, $17$, $19$ and $23$ ($L=\res{E2.mega.L}$) on one random input of each of the lengths $10^4$, $10^5$ and $10^6$.

\textbf{Results.} Every input agrees with Theorem~\ref{thm:range}. In \res{E2.in.phrase} inputs whose span is below $L$ the outputs agree with the unbounded clocks, and in \res{E2.alias.phrase} inputs with a misread the first misread comes after the span has reached $L$. With the cycles $3,5,7,11,13$, \res{E2.big.rev.k} of \res{E2.big.rev.m} reversal machines and \res{E2.big.dup.k} of \res{E2.big.dup.m} duplication machines, trained on at most $40$ symbols, are correct on every sampled input up to length \res{E2.ext.revdup}, \res{E2.ext.factor} times their longest training input. Table~\ref{tab:exchange} gives, for every machine that fits and extrapolates, the shortest sampled length with an output that differs from the unbounded clocks. For the \res{E2.nmach} machines that err at some sampled length this onset moves out with every enlargement of $L$, at ratios $n/L$ between \res{E2.ratio.lo} and \res{E2.ratio.hi} (Figure~\ref{fig:tests}d). The onset comes where the span first reaches $L$: at length $L$ for a clock that advances one reading per input symbol, and earlier for a machine whose readings run ahead of the input, for example a clock that advances two readings per answer step. One machine, on \res{E2.never}, never errs. The readings its heads compare span at most \res{E2.never.span} values even at length \res{E2.never.n}, so it never reaches far into its past. The task allows this, since over a binary alphabet the missing symbol is the parity of the visible ones. For reversal, the accuracy beyond the range follows $(L+\tfrac12(n-L))/n$ to within \res{E2.rev.err} percentage points on average over \res{E2.rev.n} combinations of machine, $L$ and length (Figure~\ref{fig:tests}c). With one cycle removed, the outputs agree with the unbounded clocks in \res{E2.abl.phrase} inputs whose compared readings span fewer than the shortened range. With the cycles $11$ to $23$, the \res{E2.mega.k} benchmark reversal machines reverse each of these inputs without error, the longest of a million symbols, and so do the three trained up to $5$ and \res{E2.mega.t10.k} of the three trained up to $10$; the heads of the third trained up to $10$ compare readings up to \res{E2.mega.t10.span} times the input length apart, so that the input of a million symbols lies beyond its range, and there it errs on \res{E2.mega.t10.err} of its answers; with unbounded clocks it reverses the same input without error. The machines trained up to $3$, which do not extrapolate, score the same at every length with these cycles as with unbounded clocks.

\textbf{Reading.} The controller is learned, and the range belongs to the clocks. What the controller learned at the readings of short inputs holds at every reading of the circle, so its range can be extended or shortened after learning, as a longer tape extends a Turing machine, to a million symbols for a rule learned from five, and beyond the range the failure of reversal follows the aliasing rule of Appendix~\ref{app:range}, a curve with no free parameter. Removing a grid module should shorten the range of exact memory in the same way (P1, Appendix~\ref{app:predictions}).

\restable{tab:exchange}

\subsection{What may not be read}
\label{app:controls}
\textbf{Why.} Section~\ref{sec:forbids} names what lets a learner fit its data without the data fixing its rule: a blurred match, which reports a distance; a controller that holds whole training inputs, as the machines on bracketed arithmetic do (Appendix~\ref{app:bench}); and the readings themselves, which keep taking new values up to the range. By Theorem~\ref{thm:hole}(i), a coordinate, or a coincidence blurred into a distance, leaves the rule open beyond the training lengths. Three variants of this experiment give the machine a blurred match, the phases of its clocks, and a coordinate of another kind, the step index (Section~\ref{sec:exp_forbid}); a fourth removes the noise added to the controller in training, to ask whether the noise or the reading fixes the rule. Leibniz's argument (Section~\ref{sec:forbids}) gives a test of reading that needs only inputs of the training lengths: a common shift of every reading, or of the step index, changes no answer of a learner that reads only coincidences.

\textbf{Design.} Four variants of the machine are retrained on reversal, duplication, stack manipulation, binary addition and modular arithmetic, three seeds each, with the full protocol and restart ladder, and tested on $256$ inputs at each of the lengths $100$, $200$ and $500$ (Table~\ref{tab:controls}). \emph{Graded clocks}: the moves stay mixtures at temperature $1$ throughout training and at test, so that a clock is a blend of states and a read is a weighted average over near matches. \emph{Step index}: the controller also receives $t/40$, a coordinate whose values beyond the training lengths are new. \emph{Phases}: the controller also receives the bumps of all clocks, the phase of every cycle; each phase takes finitely many values, and together they are the reading, a coordinate within the range. \emph{No noise}: the controller's state is never perturbed. Leibniz's test runs on every run given the step index or the phases, on $64$ inputs at each of the lengths $31$, $34$, $37$ and $40$: the step index is shifted by $1$, $10$, $40$ and $460$ steps, the last giving an input of length $40$ the index values of an input of length $500$, and every clock, with the reading \home{} returns to, by $1$, $7$, $100$ and $1{,}155$ readings. The share of answers moved is the largest, over the four shifts, of the fraction of output symbols that change (Table~\ref{tab:shift}).

\textbf{Results.} With graded clocks \res{C.graded.fit} runs fit, yet at length $500$ they fail on every task that reads its past, falling to \res{C.graded.lo} to \res{C.graded.hi}\%; modular arithmetic, which needs no clock, is unaffected (Table~\ref{tab:controls}). Given the step index or the phases, every run fits. The runs on reversal, duplication and stack manipulation change \res{L.clk.phrase} of their answers under Leibniz's shift, and all are exact at length $500$. Every run on modular arithmetic given the step index changes \res{L.mod.lo} to \res{L.mod.hi}\% of its answers when the index is moved to the values of length $500$, and falls there to \res{C.coord.mod.lo} to \res{C.coord.mod.hi}\%. The runs on binary addition, and on modular arithmetic given the phases, change between none and \res{L.mid.hi}\% of their answers and score \res{L.mid.lo} to \res{L.mid.top}\% at length $500$ (Table~\ref{tab:shift}; Figure~\ref{fig:tests}e). Without noise, \res{C.cap.fit} runs fit, and every run that fits stays above $99\%$.

\textbf{Reading.} A blurred coincidence reads a distance, so graded clocks fit the training lengths and leave the rule open beyond them. A coordinate offered to the controller leaves the rule open, and whether a run reads it shows on inputs of the training lengths: the runs whose answers Leibniz's shift moves lose accuracy beyond the training lengths, and the runs that leave the coordinate unread keep their rule. What fixes the rule is what the machine reads: free of the noise added in training, \res{C.cap.fit} runs find a procedure that holds at length $500$.

\restable{tab:controls}
\restable{tab:shift}

\subsection{How much experience fixes a rule}
\label{app:teach}
\textbf{Why.} By Lemma~\ref{lem:closure}, a learner that acts as a procedure on every input up to the closure size $N^\ast$, beyond which the procedure meets no new situation, acts as it on every input within range (Section~\ref{sec:reads}), and experience shorter than $N^\ast$ leaves some situation of the procedure unmet. For the six tasks of this experiment $N^\ast$ is \res{E0.N.reverse_string} for reversal, \res{E0.N.duplicate_string} for duplication, \res{E0.N.stack_manipulation} for stack manipulation, \res{E0.N.odds_first} for odds first, \res{E0.N.binary_addition} for binary addition and \res{E0.N.modular_arithmetic} for modular arithmetic (Table~\ref{tab:reference}). A learner can also fit its experience with another procedure, one that gives the task's answers on every input it has met and departs from them beyond; Theorem~\ref{thm:hole}(ii) rules such procedures out only from the larger size $N_{\rm teach}$ on, and a controller of $256$ units, with states to spare, leaves room for them (Lemma~\ref{lem:room}).

\begin{figure}[t]
\centering
\resultfigure{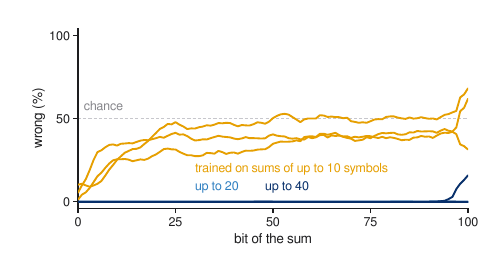}{3.3in}{code/figures/figS_quus.py}{1.75in}
\caption{\textbf{A quus found by training.} Machines trained on binary addition with sums of up to $10$ (orange), $20$ (light blue) and $40$ symbols (dark blue) add two random numbers of $100$ bits each, the highest bit one; one line per seed gives the share of the $256$ sums whose bit is wrong, averaged over five neighbouring bits.}
\label{fig:quus}
\end{figure}

\textbf{Design.} Machines are retrained on reversal, duplication, stack manipulation, odds first, binary addition and modular arithmetic with inputs of lengths $1$ to $N$, $N=3$, $5$, $10$ or $20$, three seeds each, with the restart ladder of the benchmark. They are selected on validation inputs of lengths $N-3$ to $N$ and tested on lengths $41$ to $500$ like the benchmark (Table~\ref{tab:teach}). The reversal machines trained up to $3$, $5$ and $10$ are also run with the cycles $11$ to $23$ (Appendix~\ref{app:range_exp}), and every machine is run on inputs of its training lengths with the catalogue removed. The binary addition machines trained up to $10$ and up to $20$, and the benchmark machines trained up to $40$, add two random numbers of \res{Q.bits} bits each, the highest bit one, in \res{Q.n} sums, and the errors are counted at every bit of the sum (Figure~\ref{fig:quus}).

\textbf{Results.} All \res{T.n} runs fit. None of the \res{T.below.n} runs trained on fewer symbols than its task's $N^\ast$ extrapolates, and the best of them scores \res{T.below.max}\% at lengths $41$ to $500$. Trained up to length $5$, every reversal machine extrapolates, and with the cycles $11$ to $23$ each of them reverses every input without error, the longest of a million symbols. Trained up to $3$, one more symbol than reversal's $N^\ast$, the reversal machines score \res{T.3.rev.lo} to \res{T.3.rev.hi}\%. Of the \res{T.near.m} runs trained on at most three or five symbols that reach their task's $N^\ast$, \res{T.near.k} extrapolate, all trained up to $5$: the three reversal machines and one machine each on duplication and stack manipulation. Trained up to length $10$, every run on reversal, duplication and stack manipulation extrapolates, as do \res{T.10.odds.k} of the three runs on odds first, and with the cycles $11$ to $23$ \res{E2.mega.t10.k} of the three reversal machines reverse a string of a million symbols without error (Appendix~\ref{app:range_exp}). On modular arithmetic, whose $N^\ast$ is \res{E0.N.modular_arithmetic}, every run trained up to $10$ fails and every run trained up to $20$ extrapolates. On binary addition the machines trained up to $10$ give the sum on the inputs of their training lengths and score \res{T.10.add.lo} to \res{T.10.add.hi}\% at lengths $41$ to $500$; they read their past through the catalogue, and without it they answer \res{B.add.lo} to \res{B.add.hi}\% of the symbols of their training inputs (Table~\ref{tab:bare}). On the sums of two numbers of \res{Q.bits} bits they get \res{Q.10.low.lo} to \res{Q.10.low.hi}\% of the four lowest bits wrong and \res{Q.10.high.lo} to \res{Q.10.high.hi}\% of the bits beyond the twentieth, near the $50\%$ of chance, and no sum right (Figure~\ref{fig:quus}). Trained up to $20$, they add \res{Q.20.exact.lo} to $100\%$ of these sums exactly and score \res{T.20.add.lo} to \res{T.20.add.hi}\% at lengths $41$ to $500$. The benchmark machines, trained up to $40$, add \res{Q.40.exact.lo} to \res{Q.40.exact.hi}\% of them exactly; the lowest is the machine that keeps its few errors at every length (Appendix~\ref{app:bench}), here at the highest bits and the end mark.

\textbf{Reading.} Experience fixes a procedure only once it contains the situations of that procedure: no run trained below $N^\ast$ extrapolates, and modular arithmetic needs inputs longer than its $N^\ast$. From $N^\ast$ on, very little experience can fix a rule for every length the clocks reach: five symbols fix reversal for a million. Near $N^\ast$, training also finds other procedures that agree with the task on the inputs met, which the spare states of the controller allow below $N_{\rm teach}$ (Lemma~\ref{lem:room}), and longer experience removes them: inputs of up to $10$ symbols fix reversal, duplication and stack manipulation in every run. On binary addition, ten symbols leave room for another rule on the clocks, a quus: it agrees with addition on the sums met and departs from it beyond, the more the higher the bit. Twenty symbols all but remove it, and the rule they fix adds numbers of a hundred bits.

\restable{tab:teach}

\subsection{How many clocks a procedure needs}
\label{app:clocks_exp}
\textbf{Why.} Our procedures keep one clock for each independent count (Section~\ref{sec:exp_teach}). To sort $k$ kinds of symbol in step with the input, a learner needs the $k-1$ boundaries between kinds when the output begins. These counts vary independently, and the procedure of Appendix~\ref{app:programs} keeps each in a clock of its own. The experiment asks whether training finds a machine with fewer clocks that keeps pace. For Turing machines the corresponding statement is a theorem: for every $C$ some language is recognized in real time with $C+1$ tapes and not with $C$ \citep{aanderaa1974k}. Through Theorem~\ref{thm:tapes} the same holds for clocks moved one reading per step and read at offset $0$ (Section~\ref{sec:computation}, Appendix~\ref{app:tapes}).

\textbf{Design.} Our procedure for bucket sort over five symbols keeps four counts in four clocks (Appendix~\ref{app:programs}). Machines with one, two and four clocks are trained on it with the full protocol, three seeds each, and tested at every fourth length from $41$ to $497$ (Table~\ref{tab:clocks}); the benchmark machines have three clocks. Bucket sort over $k=3$ and $4$ kinds of symbol is trained with the full protocol with $k-2$ and $k-1$ clocks, three seeds each, and tested at every fourth length from $41$ to $497$ (Table~\ref{tab:hier}).

\textbf{Results.} With five kinds of symbol no machine with fewer than four clocks fits, and the best validation accuracies are \res{K.val.1}\%, \res{K.val.2}\% and \res{K.val.3}\% with one, two and three clocks. The mean test scores are \res{K.mean.1}\%, \res{K.mean.2}\%, \res{K.mean.3}\% and \res{K.mean.4}\% with one to four clocks, rising with the number of boundaries the clocks can place (Table~\ref{tab:clocks}). With four clocks \res{K.c4.k} of the \res{K.c4.n} runs fits, and it sorts exactly at every tested length. Over $k=3$, $4$ and $5$ kinds of symbol, \res{H.fail.fit} of the \res{H.fail.n} machines with fewer than $k-1$ clocks fits its training lengths (best validation accuracy \res{H.fail.maxval}\%). With $k-1$ clocks \res{H.exact.phrase} sorts exactly at every tested length, and the others reach test scores between \res{H.other.lo}\% and \res{H.other.hi}\% (Table~\ref{tab:hier}). Figure~\ref{fig:neural}f collects all three values of $k$.

\textbf{Reading.} In these runs each independent count takes a clock of its own: with fewer clocks a controller of $256$ units does not fit even the training lengths, and with $k-1$ clocks training can find the exact procedure. Clocks are a resource that no controller replaces, as the hierarchy of tapes says for Turing machines.

\restable{tab:clocks}
\restable{tab:hier}

\subsection{What the trained machines do}
\label{app:dynamics}
\textbf{Why.} The theory fixes what the controller may read and leaves free how it moves its clocks, so the movements show which procedure a trained machine has found. Proposition~\ref{prop:disciplines} and Section~\ref{sec:disciplines} describe the movements that carry procedures: a stack is a clock moved forward by each push and back by each pop, and a copy is a clock sent \home{} and run again over the same readings. In the brain these movements are the forms of hippocampal replay (Section~\ref{sec:anatomy}).

\begin{figure}[ht]
\centering
\includegraphics[width=\textwidth]{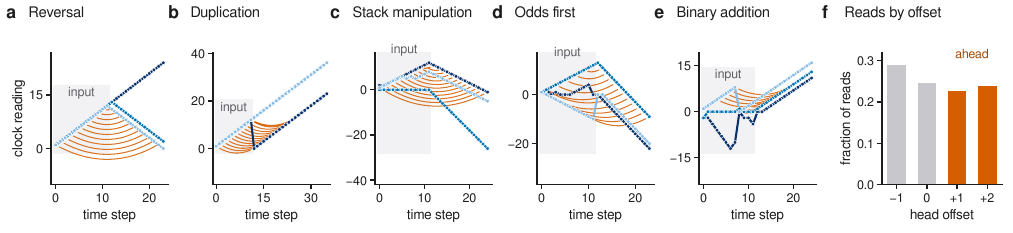}
\caption{\textbf{Clock movements of trained machines} (seed $0$, one input each; shaded: input). Lines: clock readings; arcs: reads during the answer, from the present step to the record found. \textbf{(a)} Reversal: a clock walks back. \textbf{(b)} Duplication: a clock returns \home{} and runs forward again. \textbf{(c)} Stack manipulation: clocks follow the depth and walk down to answer. \textbf{(d)} Odds first: clocks move two readings per step. \textbf{(e)} Binary addition: a clock returns \home{} and advances with the digits of the answer. \textbf{(f)} Fraction of the records found that are returned by heads at each offset, over all tasks and seeds.}
\label{fig:dynamicsS}
\end{figure}

\textbf{Design.} For every benchmark machine we record, on $16$ random inputs of length $100$, the move of every clock at every step and every record a head returns, with the offset of that head. On two example inputs of lengths $12$ and $24$ we record the complete run (\texttt{code/experiments/e3\_dynamics.py}). Figure~\ref{fig:dynamicsS} shows the clock readings and reads of five machines on one input each. Table~\ref{tab:moves} summarizes how the clocks of all machines on six tasks move while the input arrives, at the first answer step of the example inputs, and while the answer is given.

\textbf{Results.} The movements are the procedures of Appendix~\ref{app:reference}, found by training in different arrangements (Table~\ref{tab:moves}; Figure~\ref{fig:tests}b). Each reversal machine has a clock that advances while the input arrives and walks back while the answer is given (Figure~\ref{fig:dynamicsS}a). Each duplication machine sends one clock \home{} at the first answer step and runs it forward again (Figure~\ref{fig:dynamicsS}b). Each stack machine walks a clock down while answering (Figure~\ref{fig:dynamicsS}c). Each odds first machine moves its clocks by two readings per answer step, forward in two seeds and backward in one, so that every answer reads every second input (Figure~\ref{fig:dynamicsS}d). Over all machines, heads at positive offsets, which read ahead of the query clock, return \res{D.pos}\% of the records found (Figure~\ref{fig:dynamicsS}f).

\textbf{Reading.} Trained on inputs and outputs alone, the machines find the programs of Section~\ref{sec:disciplines}: reversal walks a clock back as a stack does, and duplication sends a clock \home{} and runs it again as a copy does. These movements have the forms of hippocampal replay, reverse replay in reversal and forward replay in duplication.

\restable{tab:moves}

\subsection{The machines in neural dynamics}
\label{app:neural_exp}
\textbf{Why.} By Theorem~\ref{thm:neural}, the attractor network of Eq.~\eqref{eq:cann} is the machine on its stable states (Section~\ref{sec:cann}). On those states reading the network after a move gives the move applied to the reading (part ii), so a controller trained on symbolic clocks should run unchanged on neurons (Section~\ref{sec:exp_neural}). A place cell that fires when all but $j$ rings agree shrinks the range to the product of the $M-j$ smallest periods (part iii). Wells protect the reading against noise, and without them the position of a bump is graded and its errors accumulate; adaptation shortens the lag of a bump behind a moving input, above $\tau/\tau_v$ the bump leads a slow input, and without input it travels by itself once it overcomes the wells (the two further properties of Appendix~\ref{app:neural_proof}). Replay, the firing of place cells in sequence, is on the theory a clock moved over stored readings (Section~\ref{sec:anatomy}); the controller can drive it while it answers, and adaptation can drive it when input is absent.

\textbf{Design.} \emph{Trained machines.} The benchmark controllers of ten tasks, three seeds each, run on neural clocks and place cells (Appendix~\ref{app:neural_executor}) on $16$ random inputs at lengths $41$, $100$, $200$ and $500$. Their outputs are compared symbol by symbol with the symbolic machine on the same inputs (Table~\ref{tab:neural}). Five tasks are also run in variants: noise $\sigma=0.6$ and $0.8$; place cells that fire when all but one or two rings agree, at lengths around the predicted ranges $210$ and $30$; no wells, with $\sigma=0.3$; and a recurrent kernel twice as wide (Table~\ref{tab:neuralvar}). \emph{Clocks alone.} Sixteen clocks of rings $2,3,5,7,11$ are driven for $2{,}000$ steps by moves drawn uniformly from $\{-2,\dots,2\}$, with a reset at a step with probability $0.02$, with wells ($\epsilon=1$) and without ($\epsilon=0$), at noise $\sigma$ from $0$ to $1.2$. A slip is a step after which the nearest well of some ring differs from the symbolic clock driven by the same moves. After a slip the symbolic clock takes the neural reading, so that slips are counted one by one (Table~\ref{tab:clockslips}). \emph{Replay.} Each reversal and duplication machine answers one input of length $40$. For every pair of clocks, each answer step at which the first clock shows a reading that the second showed during the input is matched with the last input step at that reading, as a head that compares the two clocks does. The pair that revisits the most input moments carries the replay. The rank correlation between answer steps and matched input steps is negative for reverse replay and positive for forward replay. A clock walked through $40$ readings is then released without input, with wells of depth $0.5$ and adaptation from $0$ to $1.2$. \emph{Adaptation.} In one ring of $11$ wells a bump is released with a trail of adaptation behind it, for $m$ from $0$ to $1.2$ and $\epsilon\in\{0,0.5,1\}$, and its speed is measured over the second half of $600\tau$. A bump without wells follows an input of strength $0.3$ or $1$ that moves at $0.05$ or $0.1$ neurons per $\tau$, and its lead is measured (Table~\ref{tab:adapt}). The code is \texttt{code/experiments/e4\_neural.py}.

\restable{tab:neural}
\restable{tab:neuralvar}
\restable{tab:clockslips}
\restable{tab:adapt}

\textbf{Results.} \emph{Trained machines.} Without noise the neural machines give the symbolic output on \res{N.exact.phrase} test inputs of \res{N.exact.ntasks} tasks, and no clock leaves its well in any step of any run (Table~\ref{tab:neural}; Figure~\ref{fig:neural}a). Noise of strength $0.6$ on every neuron leaves the output of \res{N.noise06.agree} inputs unchanged; at $0.8$ a neural reading differs from the symbolic one at \res{N.noise08.slips} steps and \res{N.noise08.agree}\% of output symbols agree. Here a slip counts at every later step until a reset brings the two clocks together again, whereas the clocks alone below count each slip once. Place cells that tolerate one or two disagreeing rings stop being recruited at \res{N.partial1.cells} and \res{N.partial2.cells} cells (Figure~\ref{fig:brain}d). With these place cells the machines agree with the symbolic machine on \res{N.partial1.below} and \res{N.partial2.below} inputs whose span is below $210$ and $30$ (Figure~\ref{fig:neural}c). Without wells, at noise $0.3$, the clocks slip at most steps and agreement falls with length, from \res{N.nowells.hi}\% at length \res{N.nowells.n0} to \res{N.nowells.lo}\% at \res{N.nowells.n1}. A recurrent kernel twice as wide leaves the output of \res{N.broad.agree} inputs unchanged (Table~\ref{tab:neuralvar}). \emph{Clocks alone.} With wells the clocks never slip in \res{N.clock.steps} steps at any noise up to $\sigma=\res{N.clock.maxsig}$. At $\sigma=\res{N.clock.firstsig}$ slips appear, in \res{N.clock.firstrate}\% of steps, on the ring of \res{N.clock.firstring} wells, the largest. Without wells the clocks slip at every noise level, in \res{N.clock.nowell08}\% of steps at $\sigma=0.8$ and in \res{N.clock.nowell0}\% even without noise (Table~\ref{tab:clockslips}; Figure~\ref{fig:neural}b). The position of a free bump is graded, and the small error of every move accumulates, which is why Section~\ref{sec:forbids} requires clocks to sit in discrete attractor states. \emph{Replay.} The answers of reversal machines revisit the input in reverse order and those of duplication machines in the same order, with rank correlations \res{N.replay.rev} and \res{N.replay.dup} on average (Figure~\ref{fig:neural}d). Two of the three duplication machines replay across clocks: a clock that rested during the input runs forward during the answer over the readings that another clock recorded. In one of them a third clock revisits its own readings in reverse order at the same time. A clock walked through $40$ readings and released without input, with wells of depth $0.5$, rests for adaptation below \res{N.release.m}. At \res{N.release.m} it runs \res{N.release.dir} through the stored readings by itself, \res{N.release.seq}, resting about \res{N.release.dwell}$\tau$ at each. \emph{Adaptation.} Without wells a released bump travels from $m=\res{N.adapt.onset}$, just above $\tau/\tau_v=\res{N.adapt.threshold}$, and faster as $m$ grows (Figure~\ref{fig:brain}e). With wells of depth $0.5$ travel starts between $m=0.2$ and $0.3$, and with depth $1$ adaptation strong enough to cross the wells extinguishes the bump. A bump that follows a moving input lags it less as adaptation grows. It leads an input moving at $0.05$ neurons per $\tau$ from $m=0.05$, and at $m=0.08$, the largest value tested, it still trails an input twice as fast (Table~\ref{tab:adapt}).

\textbf{Reading.} The controller cannot tell the network from the symbolic machine: on its stable states the network is the machine, as Theorem~\ref{thm:neural} says. Wells keep the network on those states, and place cells that require every ring to agree keep the full range; without wells the errors of the reading accumulate, and tolerant place cells shorten the range to the product of the periods that must agree. Replay takes the two forms the theory gives it: the controller drives it while it answers, in reverse order in reversal and in the same order in duplication, and adaptation drives it when input is absent.

\FloatBarrier
\subsection{Turing machines and time}
\label{app:tm_exp}
\textbf{Why.} By Theorem~\ref{thm:tapes}, a Turing machine with $C$ work tapes that reads one input symbol per step is, step for step, a machine of Eq.~\eqref{eq:machine} with $C$ unbounded clocks. By Theorem~\ref{thm:range}, clocks of range $L$ run it in the same way as long as its head range stays below $L$. No machine of Eq.~\eqref{eq:machine} multiplies online one step per input symbol. Given steps without input, one clock runs any Turing machine whose tape fits in its range, so a machine given time to think should multiply (Section~\ref{sec:computation}, Appendix~\ref{app:tapes}).

\textbf{Design.} \emph{Compiled machines} (\texttt{code/experiments/e5\_computation.py}). A transducer with $C$ tapes that runs in real time is compiled into a controller as in the proof of Theorem~\ref{thm:tapes}(i) (\texttt{code/orrery/tm.py}). Random machines have $C\in\{1,2,3\}$ tapes, $8$ states, binary input and output, and tape symbols blank, $1$ and $2$. Every entry of the table is drawn once, uniformly, with each head moving left, right or not at all with probabilities $0.25$, $0.25$ and $0.5$. Twenty machines for each $C$ run on four random inputs at each of the lengths $10$, $30$, $100$, $300$ and $1{,}000$, with unbounded clocks and with clocks of the cycles $3,4,5$, $2,3,5,7$ and $2,3,5,7,11$. The output of every step is compared with a direct simulation of the machine. The head range of a run is the largest extent of the squares visited by one head (Table~\ref{tab:tmexact}). \emph{Multiplication with time to think.} A machine with three tapes copies the factors $a$ and $b$, lowest bit first, onto tapes $0$ and $1$ as they arrive. During the steps without input it adds $a$, shifted, onto an accumulator on tape $2$ for every bit of $b$ that is one, propagating carries. It then rewinds, marks the highest one of the product and waits for the answer cues, and at the cues it emits the product, lowest bit first, and an end mark. On the benchmark inputs of length $n$ it is given $(n-1)^2/2+8n+16$ steps to think, which suffice at every length, and it is compared with the same machine given none (Table~\ref{tab:think}). \emph{On neural dynamics}, five compiled random machines with one and with two tapes run on four inputs of length $100$, and the multiplier on two inputs at each of the lengths $5$, $8$ and $12$, with neural clocks and place cells.

\begin{figure}[t]
\centering
\resultfigure{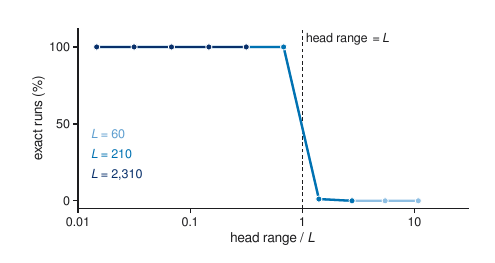}{3.3in}{code/figures/figS_results.py}{1.75in}
\caption{\textbf{Compiled Turing machines with bounded clocks.} Fraction of runs of compiled random Turing machines that agree with direct simulation, against the head range divided by $L$, for $L=60$, $210$ and $2{,}310$ (the three lines coincide); every run whose head range stays below $L$ is exact.}
\label{fig:results}
\end{figure}

\textbf{Results.} The compiled machines agree with direct simulation in \res{C.tm.phrase} runs with unbounded clocks, and with bounded clocks in \res{C.tm.within} runs whose head range is below $L$; beyond the range they agree in \res{C.tm.beyond.k} of \res{C.tm.beyond.m}, as Theorem~\ref{thm:range} allows (Table~\ref{tab:tmexact}; Figure~\ref{fig:results}). Given time, the multiplier is exact on every input at every tested length up to \res{C.think.maxn}, with up to \res{C.think.steps} steps to think. Without time it scores at most \res{C.think.acc0}\% (Table~\ref{tab:think}; Figure~\ref{fig:neural}e). On neural dynamics the compiled random machines agree with direct simulation in \res{C.neural.tm} runs, and the multiplier is exact up to length \res{C.neural.mult.maxn}.

\textbf{Reading.} A tape is a clock and its cells are records: a compiled Turing machine runs exactly on clocks, symbolic or neural, while its heads stay within the range. The trained machines, given no steps without input, fail at multiplication; given such steps, a compiled machine of the same kind multiplies exactly at every tested length.

\restable{tab:tmexact}
\restable{tab:think}

%% file: app_evidence.tex
\section{Neural Evidence}
\label{app:evidence}

This appendix collects the findings behind Section~\ref{sec:anatomy}, part by part of the machine, and supports the paragraph \emph{What the theory explains} of that section. Each paragraph ends with what the theory makes of the findings, and Appendix~\ref{app:predictions} states the predictions of Section~\ref{sec:anatomy} in full, P1 to P8.

\textbf{The clocks are grid modules.} Grid cells fire at the vertices of a triangular lattice that tiles the environment \citep{hafting2005microstructure}. They cluster into a few discrete modules that differ in scale, orientation and theta modulation and can respond independently to changes in the geometry of the environment; the scales of adjacent modules stand in a ratio of about $1.4$ to $1.7$ \citep{stensola2012entorhinal,barry2007experience}. The joint activity of each module lies on a torus, and each cell keeps its position on the torus across environments and into sleep \citep{gardner2022toroidal}. Modules stay coordinated when their activity is dissociated from sensory cues \citep{waaga2022grid}, as one move that turns every cycle requires. The stellate cells of layer II, where many grid cells lie, are coupled almost only through inhibitory interneurons, and this recurrent inhibition is enough for an attractor network to form grids \citep{couey2013recurrent}. As a residue code the modules have a range set by the least common multiple of their periods, exponential in their number, while the update of position is addition within each module \citep{fiete2008what,sreenivasan2011grid}. Economy arguments for a nested code, whose range is the largest period, predict a geometric series of scales with the observed ratio \citep{mathis2012optimal,wei2015principle}. Proposition~\ref{prop:cycles} reaches cycles from a different requirement, that a learned procedure reach its past through coincidences only, and the coprime periods of Section~\ref{sec:representation} from the economy of the range.

\textbf{The form is given before the map.} Toroidal structure appears in the entorhinal cortex of rat pups at postnatal day $10$, before eye opening and before the pups explore, in up to three modules from the start \citep{guardamagna2026toroidal}. Head direction cells keep the couplings of an adult network before eye opening \citep{bassett2018self}. Spatially periodic fields come later: when pups first leave the nest, head direction cells are adult in form while grid cells mature last \citep{langston2010development,wills2010development}. On the theory the group of the clocks is given, and experience aligns it with the world.

\textbf{The moves are signals of movement and copies of commands.} Speed cells in the medial entorhinal cortex fire at rates that rise linearly with running speed, whatever the context \citep{kropff2015speed}. Continuous attractor networks driven by velocity and heading integrate path accurately over distances of metres \citep{burak2009accurate}, and the entorhinal cortex has been proposed to integrate the animal's movements through such a periodic network \citep{mcnaughton2006path}. Desert ants on lengthened legs overestimate the distance home and ants on shortened legs underestimate it \citep{wittlinger2006ant}: their moves are steps. Copies of motor commands routed to sensory structures, corollary discharge, are found from nematodes to primates \citep{crapse2008corollary}; in the machine every move is such a copy.

\textbf{\home{} is a reset at a boundary.} The error of grid cells grows with the distance travelled since the last encounter with a boundary, and an encounter with a wall corrects the error in the direction perpendicular to the wall and leaves the error along it \citep{hardcastle2015environmental}. In a hairpin maze the grid map breaks into fragments that repeat from arm to arm, and the map starts again at every turn \citep{derdikman2009fragmentation}. In people watching a film, the hippocampus responds at the boundaries between events, the size of its response predicts how well the event just ended is later reinstated, and the boundaries found in cortical activity match the events people report \citep{baldassano2017discovering}; boundary cells of the human medial temporal lobe mark the same moments \citep{zheng2022neurons}. In the machine \home{} returns a clock to a fixed reading: a wall resets the ring perpendicular to it, a turn of the maze resets the module, and an event boundary resets the clock under which the episode is filed.

\textbf{Coincidence detectors.} In models a place field arises from a sum of grid cells with similar phases, and a greater spread of phases gives several irregular fields \citep{solstad2006grid}; random grid inputs followed by competition through inhibition give place fields of realistic size and number without learning \citep{dealmeida2009input,monaco2011modular}. Place maps also form when grid periodicity is abolished \citep{brandon2014new}, and place cells appear in development before stable grid fields \citep{langston2010development,wills2010development}. On the theory a place cell detects a coincidence of whatever clocks reach it, of which the grid modules are the best known.

\textbf{Large spaces give many fields} (Figure~\ref{fig:brain}d). On a $48$\,m track each CA1 cell has its own propensity to form fields, and the number of fields per cell is broadly distributed \citep{rich2014large}. In bats flying through a $200$\,m tunnel only $12\%$ of place cells have a single field, and a cell has on average $4.9$ fields per direction \citep{eliav2021multiscale}. The fields of a cell are spaced without period: the gaps between them are distributed exponentially, and the authors found the data inconsistent with a strong periodic grid input. The fields also differ in size, from $0.6$ to $32$\,m across cells and twentyfold within one cell. In enclosures a few times larger than usual, CA1, CA3 and dentate cells have several irregularly arranged fields \citep{fenton2008unmasking,park2011ensemble}. On the theory a place cell that tolerates $j$ disagreeing modules (Theorem~\ref{thm:neural}(iii)) fires wherever any $M-j$ of its modules return to the readings recorded when it was recruited; different subsets return at different places, and with the incommensurate periods of real modules these places are spread irregularly along the tunnel, so the fields fall without period, as observed, while a cell that needs every module to agree fires once within the range. The variation of field sizes within one cell is left unexplained by the theory.

\textbf{Identical compartments repeat.} In two nearly identical boxes joined by a corridor many place cells have similar fields in both \citep{skaggs1998spatial}, and in four parallel compartments fields repeat and keep repeating with experience \citep{spiers2015place}, as the grid map of a hairpin maze repeats from arm to arm \citep{derdikman2009fragmentation}. Fields repeat only in compartments that face the same way, and in any arrangement after lesions of the head direction system \citep{harland2017lesions}. Grids that first repeat between two compartments merge into one pattern with prolonged experience \citep{carpenter2015grid}. On the theory a doorway resets the clocks \home{}, heading is one more ring, and a reset that stops once the layout is learned joins the two compartments on one clock.

\textbf{Remapping is a change of relative phase.} Global remapping of place cells comes with a realignment of the grid, a shift and rotation of the whole grid map \citep{fyhn2007hippocampal}, while the phase relations among the cells of one module stay fixed \citep{yoon2013specific}. In the model of \citet{monaco2011modular} a small number of modules, each shifted by its own amount, is enough to produce global remapping. In a preprint, \citet{lykken2025functional} report that in rats moving between familiar rooms each module shifted its phase by a different amount while the place cells remapped, and that smaller differences between the shifts of the modules gave partial remapping. Remapping also follows changes in the firing rates of grid fields that keep their positions \citep{kanter2017novel}. A relative shift of the cycles changes every coincidence and so opens a new catalogue on the same clocks; a change of rates changes the patterns that place cells store.

\textbf{The catalogue.} Indexing theory holds that the hippocampus stores an index of the cortical activity of each episode and that a partial cue reactivating the index reinstates the episode \citep{teyler1986hippocampal,teyler2007hippocampal}; hippocampal engrams act as such indices \citep{goode2020integrated}. Vector-HaSH builds such an index on a scaffold of grid modules, with hippocampal cells as conjunctions of grid states and sensory content bound to them \citep{chandra2025episodic}; Section~\ref{sec:anatomy} builds the catalogue in the same way, and Appendix~\ref{app:related} says where the two accounts part. In the theory the cue is the present pattern of the clocks, and among the records it reactivates the most recent is returned. This is a memory of keys and values whose keys are clock readings \citep{gershman2025key}, and a memory written and read by address as computation requires \citep{gallistel2009memory}, here built from attractors and synapses, an implementation Gallistel and King considered unlikely.

\textbf{Replay and reading ahead are clocks in motion} (Figure~\ref{fig:brain}e). Sequences of place cells replay in reverse at the end of a run \citep{foster2006reverse} and forward before it \citep{diba2007forward}, and before a goal directed run they lead from the present location to a remembered goal \citep{pfeiffer2013hippocampal}. During sleep they replay compressed in time \citep{nadasdy1999replay}. Replay also runs along paths never travelled \citep{gupta2010hippocampal} and toward unvisited but cued places \citep{olafsdottir2015hippocampal,olafsdottir2018role}, and in humans it follows the learned abstract order of a task and reverses after reward, with a code for position in the sequence leading the code for the item by $50$\,ms \citep{liu2019human}. Replay moves in discrete hops: the reactivated position sharpens at one place for several milliseconds and then jumps to the next, locked to slow gamma \citep{pfeiffer2015autoassociative}. Grid cells replay together with place cells during rest, in register with them \citep{olafsdottir2016coordinated}, and the superficial layers of the medial entorhinal cortex also replay on their own, independently of hippocampal replay \citep{oneill2017superficial}. In a $200$\,m tunnel replay is fragmented, each event covering a small part of the space \citep{eliav2025fragmented}. Within each theta cycle place cells fire at progressively earlier phases \citep{okeefe1993phase}, and ensembles sweep through the path ahead in compressed form \citep{skaggs1996theta,foster2007hippocampal}, alternating between possible futures \citep{kay2020constant}. Grid and place ensembles sweep outward from the animal, alternating left and right, and the length of a sweep is about a fifth of the grid spacing in every module \citep{vollan2025left}. In the machine a replay is a clock moved over stored readings, hopping from well to well, and a sweep is a head at a positive offset; the position code that leads the item code is a reading that retrieves its record; a clock that runs on its own, with nothing read, is entorhinal replay without the hippocampus.

\textbf{Adaptation, anticipation and acetylcholine.} Head direction cells of the anterior thalamus anticipate the future direction of the head by about $23$\,ms, while those of the postsubiculum encode its present direction \citep{blair1995anticipatory,taube1998comparisons}; attractor models with adaptation reproduce a lead time independent of turning speed \citep{mi2014spike}, and adaptation in the vestibular afferents also accounts for much of its variability \citep{vandermeer2007anticipation}. In attractor models of place and grid cells, adaptation produces theta sweeps, phase precession and procession \citep{chu2024firing}, and sweeps whose length scales with grid spacing and running speed \citep{ji2025systems}, as observed \citep{vollan2025left}; the strength of adaptation sets the statistics of replay \citep{ji2026dynamical}. Acetylcholine blocks the slow afterhyperpolarization and the accommodation of hippocampal pyramidal cells through muscarinic receptors \citep{cole1983acetylcholine,cole1984characterization}, and noradrenaline does so through $\beta$ receptors \citep{madison1986actions}. Hippocampal acetylcholine is high during walking and REM sleep and minimal during sharp wave ripples \citep{zhang2021cholinergic}, and optogenetic activation of septal cholinergic neurons suppresses ripples \citep{vandecasteele2014optogenetic}. High cholinergic tone favours encoding and low tone consolidation, in part by suppressing recurrent excitation \citep{hasselmo1999neuromodulation,hasselmo2006role}; P5 separates this route from adaptation, because only adaptation sets how fast the released sequences run. Within each theta cycle, phases that favour encoding and retrieval alternate \citep{hasselmo2002proposed}, as writes keyed by one clock and reads queried by another alternate in the machine.

\textbf{Recruitment in one trial.} A plateau potential in a CA1 cell creates a place field in one trial, potentiating inputs active seconds before and after it \citep{bittner2015conjunctive,bittner2017behavioral}, and the same plasticity shifts existing fields \citep{milstein2021bidirectional}. In the machine a place cell is recruited when no cell answers the present pattern.

\textbf{Records and rules.} The hippocampus stores an episode in one exposure and the cortex learns regularities over many, the two speeds that complementary learning systems were proposed to explain \citep{marr1971simple,mcclelland1995why}. Patient H.M., without a hippocampus, learned a motor skill over days of practice and kept no memory of the sessions in which he learned it \citep{corkin1968acquisition}. In the machine a record is data, written once and never trained, and the rule is the finite table of the controller, learned slowly and unable to depend on how many records there are; a learner without the catalogue keeps its rules.

\textbf{Clocks beyond space.} Hippocampal time cells fire at successive moments of a delay \citep{pastalkova2008internally,macdonald2011hippocampal,eichenbaum2014time}. On repeated laps of one track many CA1 cells fire at the same place on every lap with a rate that marks the lap, and the lap code carries over to a track of another shape: cells that count laps \citep{sun2020hippocampal}. Grid cells integrate elapsed time and distance on a treadmill \citep{kraus2015during}, a subcircuit of the medial entorhinal cortex represents elapsed time while the animal stands still \citep{heys2018evidence}, and the lateral entorhinal cortex carries a code of time that drifts over seconds to hours and is read as how long ago an experience was \citep{tsao2018integrating}; the medial cortex carries path integration and global reference frames, the lateral objects and local frames \citep{knierim2014functional}. Entorhinal neurons also map sound frequency \citep{aronov2017mapping}. Signals with the sixfold symmetry of grids appear in virtual navigation \citep{doeller2010evidence} and in human single neurons \citep{jacobs2013direct}, and also in conceptual and odour spaces \citep{constantinescu2016organizing,bao2019grid,bellmund2018navigating,behrens2018what}. On a linear track the fields of grid cells of one module are parallel slices through one lattice \citep{yoon2016grid}, as moves along a line walk a clock of two rings along a straight path on its torus. During mental navigation between landmarks many neurons of the macaque entorhinal cortex are periodic, with signatures of a continuous attractor, and the variability of the monkeys' timing grows more slowly than the distance, as resets at remembered landmarks predict \citep{neupane2024mental}. Medial entorhinal activity forms ultraslow periodic sequences lasting minutes \citep{gonzalo2024minute}, and medial premotor populations trace a loop that restarts with every produced interval \citep{betancourt2023amodal}. In mice that learn looping sequences of goals, most medial frontal neurons tile progress to the next goal, and position in the sequence is held by cells that fire at fixed lags from an anchor; these ring buffers persist in sleep and support inference on new sequences \citep{el2024cellular,whittington2025tale}. The left entorhinal cortex of humans represents position in nested sequences from coarse to fine along its long axis \citep{shpektor2024hierarchical}, and ordinal position within a chunk oscillates at the chunk rate \citep{al2021mental}. In macaque prefrontal cortex the items of a sequence occupy subspaces specific to their rank, each holding a ring that mirrors the circular layout of the items \citep{xie2022geometry}, and they are moved between subspaces for backward recall \citep{tian2024mental}. Mental rotation takes time proportional to the angle \citep{shepard1971mental}, as a clock moved by bounded steps does. In each of these cases an order is held by a periodic code, as it is by the clocks of the machine, and the drifting code of the lateral entorhinal cortex is a coordinate, which P7 separates from the periodic code of the medial.

\textbf{Phase arithmetic in neurons.} In \emph{Drosophila} heading is a bump on a ring attractor \citep{seelig2015neural,kim2017ring}, and travelling direction is computed by adding sinusoidal activity patterns whose phases and amplitudes are those of vectors \citep{lyu2022building,lu2022transforming}: neurons add on cycles. Head direction states of mice lie on one ring, the same in waking and in REM sleep \citep{chaudhuri2019intrinsic}, and the case for continuous attractors in the brain is now strong \citep{khona2022attractor}.

\textbf{Serial order.} Lashley argued that the serial order of skilled behaviour cannot come from chaining, in which each element triggers the next, and requires a plan prepared before execution \citep{lashley1951problem}. Models of serial recall code position relative to the start and end of a list \citep{henson1998short}, bind items to a timing signal \citep{burgess1999memory} or to oscillators of different frequencies that are reset and replayed at recall \citep{brown2000oscillator}, or let a gradually drifting context retrieve items in order \citep{howard2002distributed}. Recall errors exchange items that held the same position in different groups, and intrusions from an earlier list keep their position \citep{henson1998short}: a record made at the same reading is found, as P2 expects of a periodic clock. Judgements of order need a slow serial search whose speed depends on recency, while items are recognized in one step \citep{mcelree1993serial}; lesions of the hippocampus impair memory for the order of a series of odours and spare memory for which odours occurred \citep{fortin2002critical}. Event boundaries improve later recognition and impair memory for the order of events across them, and boundary and event cells of the human medial temporal lobe mark these moments \citep{zheng2022neurons}, as a reset \home{} at a boundary leaves no reading that orders the two sides. Backward recall has also been read as repeated forward search \citep{thomas2003forward}; P3 distinguishes this account from a clock walked back. Skilled sequences in humans are prepared as a queue organized by an ordinal template that generalizes across actions \citep{kornysheva2019neural}.

\textbf{Working memory and control.} Prefrontal neurons hold information across delays \citep{goldmanrakic1995cellular}; prefrontal activity patterns represent goals and bias the pathways from input to output \citep{miller2001integrative}, and the basal ganglia gate when prefrontal working memory is updated \citep{frank2001interactions,oreilly2006making}, as human imaging shows in the filtering of distractors \citep{mcnab2008prefrontal}. Working memory holds about four chunks \citep{cowan2001magical}, or a continuous resource shared among items \citep{ma2014changing}, and preparatory activity in motor cortex settles into discrete attractors \citep{inagaki2019discrete}. A grid-like signal appears in the medial prefrontal cortex as well as in the entorhinal cortex when people move through a conceptual space \citep{constantinescu2016organizing}. These are the properties the theory asks of a controller: few states, discrete and gated; a grid-like signal in prefrontal cortex is, on the theory, a clock the controller moves and does not read (P8).

\textbf{Recursion and its limits.} Monkeys, children and adults generate sequences with centre embedding from new items, monkeys after extra exposure \citep{ferrigno2020recursive}; crows do so on many transfer trials \citep{liao2022recursive}; macaques extend mirror sequences to greater lengths \citep{jiang2018production}. Tamarins learn the regular pattern $(AB)^n$ and not the nested $A^nB^n$, which humans learn \citep{fitch2004computational}, though humans can learn it without using its nested structure \citep{perruchet2005does}. In written language centre embedding stops at depth three, and in speech multiple centre embedding is practically absent \citep{karlsson2007constraints}. Dutch crossed dependencies are rated easier than German nested ones \citep{bach1986crossed}, as copying with a clock sent \home{} is easier than walking a clock back over a nest; a recent replication finds the difference in ratings and not in comprehension \citep{yadav2025revisiting}. Intracranial activity rises with each word and falls when words merge into a phrase \citep{nelson2017neurophysiological}, and cortical activity tracks phrases and sentences at their own rates \citep{ding2016cortical}.

\textbf{Counting and estimating} (Figure~\ref{fig:brain}f). When adults press a key a target number of times, two laws of error appear in one experiment \citep{cordes2001variability}. Adults who estimate the number without counting, their speech occupied by a repeated word, show a coefficient of variation that stays constant as the number grows, the scalar variability of a magnitude; adults who count aloud show a coefficient of variation that falls as the inverse square root of the number, as independent errors of single steps give. Neurons of the monkey prefrontal cortex tuned to numerosity have tuning curves that are logarithmically compressed, and the neural responses obey Weber's law as the behaviour does \citep{nieder2003coding,nieder2016neuronal}; nonverbal number is a magnitude with scalar variability, like duration \citep{gibbon1977scalar,gallistel2000nonverbal,feigenson2004core}. Small sets of up to about four items are apprehended at once, and larger sets are counted at a few hundred milliseconds per item \citep{trick1994why}; neurons of the human medial temporal lobe show a coding boundary near four \citep{kutter2023distinct}. The Munduruku have number words up to five; they compare and add large approximate numbers as well as speakers of languages with a full count list, and their exact arithmetic fails beyond their number words \citep{pica2004exact}. The Pirah\~a, with words for one, two and many, fail at exact tasks beyond three \citep{gordon2004numerical}, yet they match large sets exactly while the sets stay in view and fail when memory is required \citep{frank2008number}, so number words serve as records. Children learn the meanings of number words one at a time before the counting principle extends to the whole count list \citep{sarnecka2008cardinal}, and their counts most often end at a number ending in nine \citep{siegler1982development}; among about $800$ kindergarteners after one school term the most frequent stopping points were $39$ and $29$ \citep{gould2016hurdles,gould2017mapping}. On the theory, estimating reads a graded magnitude, a coordinate whose errors grow with size, while counting moves a clock and reads coincidences with a learned list, exact up to the range of the clock.

\subsection{Predictions}
\label{app:predictions}
The predictions of Section~\ref{sec:anatomy}, stated in full. Each names the measurement that would test it and, where the main text gives one, the result that would refute the theory.

\textbf{P1. Aliasing at the least common multiple.} Exact sequence memory that relies on the grid modules fails where the joint phase of the modules in play repeats, the least common multiple of their periods; silencing one module moves the failure to the least common multiple of the rest, and restoring it moves the failure back. The feasible form uses the anatomy: grid scale grows from the dorsal to the ventral end of the medial entorhinal cortex \citep{hafting2005microstructure,stensola2012entorhinal}, so the ventral part holds the modules of largest period and the dorsal part the most grid cells. Lesions of ventral medial entorhinal cortex should shorten exact sequence memory most, since they remove the largest factor of the range, and dorsal lesions, which remove the most grid cells, least. An account in which capacity is set by the number of cells or synapses predicts the reverse order.

\textbf{P2. Periodic codes of serial position.} In long sequences the neural similarity of positions $i$ and $i+p$ is raised for every engaged period $p$, above the similarity of positions at other distances. A code that represents position by a monotone magnitude predicts a similarity that falls with distance and nothing else.

\textbf{P3. Backward recall is reverse replay.} Serial position decoded from neural activity during backward recall sweeps down through the positions written during encoding, one position per recalled item, as the clock walks back; in forward recall it sweeps up. Backward recall by repeated forward search \citep{thomas2003forward} predicts a sweep up before each recalled item.

\textbf{P4. Partial coincidence shortens the range.} Weakening the inhibition that sharpens place fields \citep{grienberger2017inhibitory} lets a place cell fire when $j$ of the modules that drive it disagree, and caps exact memory at the product of the $M-j$ smallest periods (Theorem~\ref{thm:neural}(iii); Figure~\ref{fig:brain}d). The experiment of Section~\ref{sec:exp_neural} runs this manipulation on the model, where place cells that tolerate one or two disagreeing rings shorten the range from $2{,}310$ to $210$ and $30$.

\textbf{P5. Adaptation releases replay.} Acetylcholine suppresses the slow afterhyperpolarization that produces adaptation \citep{cole1983acetylcholine} and suppresses replay \citep{vandecasteele2014optogenetic}; low cholinergic tone, which \citet{hasselmo1999neuromodulation} tied to consolidation, releases sequences that run without input. The new part of the prediction is the speed: released sequences run faster as adaptation grows, as the travelling bump of Appendix~\ref{app:neural_proof} does, and a route through recurrent excitation alone leaves the speed unchanged.

\textbf{P6. Discrete resting phases and grid-place register during replay.} During a hippocampal replay the decoded phase of each grid module moves in discrete steps, resting in one well and jumping to the next as the decoded place cell position hops \citep{pfeiffer2015autoassociative}, and it stays in register with the place cells: the phases decoded at each hop are the phases under which the place cells firing there were recruited, as the coordinated grid and place cell replay of \citet{olafsdottir2016coordinated} shows for rest. Entorhinal replay without the hippocampus \citep{oneill2017superficial} is a clock run with nothing read, which the theory allows. A hippocampal replay during which the grid stands still would refute the identification of the grid modules with the clocks.

\textbf{P7. Order versus elapsed time.} Memory for the order of events rests on the periodic time code of the medial entorhinal cortex \citep{kraus2015during,heys2018evidence}, and memory for how long ago an event happened on the drifting code of the lateral \citep{tsao2018integrating}, in line with the division of labour between the two \citep{knierim2014functional}. Silencing the medial cortex should impair judgements of order and spare judgements of elapsed time, and silencing the lateral the reverse, because order is read as a coincidence on a clock and elapsed time as a coordinate.

\textbf{P8. Shift invariance of the prefrontal controller.} The prefrontal code of the pending items of a procedure is unchanged when a new room shifts every grid module and the place cells remap, because the controller receives what the heads find and never the readings themselves. A grid-like signal in prefrontal cortex \citep{constantinescu2016organizing} is a clock the controller moves and does not read: it should shift with the entorhinal modules while the code of the pending items stays. Retrieved records are decodable from the prefrontal neurons that hold working memory, and the phase of the grid code is not.

%% file: app_related.tex
\section{Related Work}
\label{app:related}

This appendix places the theory among earlier work: length generalization in networks, the cycles that networks find when they generalize, neural theories of structure, computation and expressivity, learnability, automata in neural dynamics, continuous attractor networks, symmetry in learning, and recursion, rules and symbols. It supports Sections~\ref{sec:computation}, \ref{sec:brain} and~\ref{sec:discussion}, and Table~\ref{tab:diagnosis} lists what common architectures read.

\textbf{Length generalization in networks.} Memory-augmented networks gave a controller a tape, a stack or a matrix of slots \citep{graves2014neural,graves2016hybrid,weston2015memory,joulin2015inferring,grefenstette2015learning}. In the benchmark of \citet{deletang2023neural} networks order roughly by the Chomsky hierarchy, no architecture solves binary multiplication or square root, and the networks given a stack or a tape are the ones that extrapolate on the five tasks of Section~\ref{sec:exp_bench} that reach their past through a clock: the Stack-RNN extrapolates on stack manipulation and reversal, and the Tape-RNN is the one network that extrapolates on duplication, odds first and binary addition. The Tape-RNN's best scores use a head that can also jump by the input length, a move supplied by hand. These networks read their memory through graded attention, and their structure, like ours, is built by hand; the theory says which structure a learner needs, why a stack and a tape are such structures, and that the read has to be a coincidence. Length generalization in Transformers has been pursued through positional encodings \citep{press2022train,su2024roformer,kazemnejad2023impact,peng2024yarn}: randomized encodings \citep{ruoss2023randomized} and encodings adapted to the data \citep{zheng2024dape} improve Transformers on parts of the benchmark, as do stack attention and recurrent variants \citep{li2024transformer,beck2024xlstm,grazzi2025unlocking}, and the best of them reach $75.1$, $69.3$ and $64.5\%$ on duplication, odds first and binary addition. Decoders trained on short inputs generalize best with no positional encoding at all \citep{kazemnejad2023impact}, and a network trained to decide when to stop learns to read its position \citep{newman2020eos}. Small Transformers trained on additions of up to seven digits fail at eight, and trained on one and three digits fail at two \citep{lee2024teaching}. Systematic generalization can be learned by meta learning, yet the same models fail on longer outputs \citep{lake2023human}, and longer inputs are the hardest test of compositionality \citep{lake2018generalization,hupkes2020compositionality,keysers2020measuring}; processors of graphs trained on small graphs reach about three quarters of their accuracy on larger ones \citep{velickovic2021neural,velickovic2022clrs,ibarz2022generalist}.

Many networks do extrapolate. Recurrent and convolutional networks extend from small to large instances of addition, multiplication, mazes and prefix sums when trained with curricula, noise and many restarts \citep{kaiser2016neural,price2016extensions,freivalds2017improving,schwarzschild2021can,bansal2022end}, especially when their fixed points do not depend on the path taken \citep{anil2022path}. A Transformer copies strings twenty times longer than its training strings by finding the previous occurrence of the present $n$-gram \citep{jelassi2024repeat}. Transformers generalize in length on arithmetic with index hints, coupled positions or embeddings that number the digits \citep{zhou2024what,cho2024position,mcleish2024transformers,jelassi2023length,shen2023positional}, with tasks chosen among the programs a Transformer can express \citep{zhou2024transformers,weiss2021thinking}, with looping and adaptive depth \citep{dehghani2019universal,graves2016adaptive,fan2025looped,giannou2023looped}, or with chain of thought formatted as the tape of a Turing machine \citep{hou2025universal}; language models tuned on short instances do not extrapolate, while scratchpad prompting helps \citep{anil2022exploring}. Each of these methods either bounds the coordinate so that training shows all its values (random positions up to a chosen maximum, offsets of digit significance up to $100$, coupled positions up to a largest identifier, a hundred index hints), and a hyperparameter then sets the range, or it replaces a read of position by a match of content. The loop count of looped models is itself a coordinate, and breaking its correlation with the length of the input reduces their failures \citep{kuo2026stabilizing}. In language models, the dimensions of rotary codes whose wavelength exceeds the training context never complete a rotation and carry absolute position \citep{peng2024yarn,liu2024scaling}, and extending the context rescales positions back into the trained range \citep{chen2023extending}: a cycle longer than experience is a coordinate. One property, whether what a decision reads takes finitely many values, sorts all of these results, and the machine of Eq.~\eqref{eq:machine} has the property by construction. Table~\ref{tab:diagnosis} lists what common architectures read.

\begin{table}[ht]
\caption{\textbf{What architectures read.} Coordinate: a quantity in the situation whose values grow with the input. Discrete: states that sit in wells. Range: how the reach of addressing is set.}
\label{tab:diagnosis}
\centering\footnotesize
\begin{tabular}{@{}llll@{}}
\toprule
Architecture & Coordinate read & Discrete & Range\\
\midrule
RNN, LSTM, state space model & a graded state that drifts with length & no & none\\
Transformer & positions, or distances by a graded similarity & no & the trained lengths\\
Stack-RNN, Tape-RNN & a blended head; jumps by the input length & no & the memory size\\
This work & none & yes & product of coprime periods\\
\bottomrule
\end{tabular}
\end{table}

\textbf{Networks that generalize find cycles.} Transformers that learn modular addition after long training \citep{power2022grokking} embed numbers on a few Fourier frequencies and compute with trigonometric identities \citep{nanda2023progress,zhong2023clock}; networks learn composition in finite groups through representations of the group \citep{chughtai2023toy}, and networks of several kinds implement an approximate Chinese remainder computation with few features \citep{mccracken2025uncovering}. Pretrained language models represent numbers with Fourier components of periods $2$, $5$ and $10$ \citep{zhou2024pretrained} and add them on a helix with periods $2$, $5$, $10$ and $100$ and one linear direction for magnitude \citep{kantamneni2025language}; days and months lie on circles \citep{engels2025not}; heads that give the successor of a number, day or month use features of the residue modulo $10$ \citep{gould2024successor}; a deployed language model tracks a character count on a rippled helix tiled by features that resemble place cells \citep{gurnee2025manifolds}. Without a cycle there is no exact count: linear recurrent networks and state space models whose transitions have nonnegative or positive eigenvalues read a graded state that drifts with length and cannot count modulo any number \citep{merrill2024illusion,sarrof2024expressive,grazzi2025unlocking}. Sinusoidal and rotary position codes are sets of cycles \citep{vaswani2017attention,su2024roformer}, and rounding every rotary wavelength to an integer improves extrapolation \citep{wang2024resonance}. Coprime periods have served as positional codes: a conversational encoder indexes positions by their residues modulo $47$ and $11$ to reach $517$ positions \citep{henderson2020convert}, and Transformers given rotations of periods $5$, $7$, $11$ and $13$ copy strings up to length $5{,}005$, the product of the periods \citep{guo2026copying}, the range Theorem~\ref{thm:range} assigns to such clocks. The principle that generalization needs locality, symmetry and discrete attractors has also been argued with a neural cellular automaton that adds numbers far longer than its training numbers \citep{wei2026spatiotemporal}. Here the cycles are derived from what a decision may read, the controller that uses them is learned, and the same machine is the dynamics of an attractor network.

\textbf{Neural theories of structure.} Grid cells have been modelled as continuous attractors \citep{burak2009accurate,mcnaughton2006path,khona2022attractor}, as the interference of velocity-controlled oscillators whose phase difference reads position \citep{burgess2007oscillatory}, as a modular code of high capacity \citep{fiete2008what,sreenivasan2011grid}, as the economical \citep{mathis2012optimal,wei2015principle} or predictive \citep{stachenfeld2017hippocampus} code of space, as the product of networks trained to integrate paths \citep{banino2018vector,cueva2018emergence}, hexagonal under simple biological constraints \citep{sorscher2023unified}, and as representations of the group of the animal's movements \citep{gao2021path,dorrell2023actionable}; here the group is derived from generalization, and the controller never reads it. Grid codes have been analysed as residue number systems and codes that correct errors \citep{fiete2008what,sreenivasan2011grid,mathis2012optimal,wei2015principle}, as residues bound into one vector by modular attractor networks \citep{kymn2024binding}, and as residue numbers that support computation in vector-symbolic representations of high dimension, which bind content to structure by algebra \citep{kymn2024computing}; multiscale codes inspired by grid cells encode locations for machine learning \citep{mai2020multiscale}. The Tolman--Eichenbaum machine separates a structural code, learned by path integration, from sensory content bound in the hippocampus, and generalizes structure across environments \citep{whittington2020tolman}; it is close to a Transformer with recurrent position codes \citep{whittington2022relating}, and structured slots in prefrontal cortex extend it to sequence memory \citep{whittington2025tale}. Graphs that clone each observation into several hidden states learn higher order structure from aliased observations \citep{george2021clone}, and key-value accounts of the hippocampus treat the entorhinal code as the key \citep{gershman2025key}. Vector-HaSH stores content on a scaffold of grid modules, with hippocampal cells as their conjunctions, and recalls a sequence by a velocity read from the hippocampal state \citep{chandra2025episodic}; that velocity is a function of the grid state, a coordinate, so the rule it learns is tied to the states met in training (Theorem~\ref{thm:hole}(i)). The hippocampus has been described as a generator of sequences independent of space and time \citep{buzsaki2018space}, and models of serial recall bind items to oscillators or drifting context \citep{brown2000oscillator,burgess1999memory,howard2002distributed}. Here the controller moves the clocks and reads only records, and the cycles, the coincidence detectors, the catalogue and the blind controller follow from one requirement, that finite experience fix the rule, a requirement that owes nothing to navigation or capacity. The same requirement gives the hippocampal index \citep{teyler1986hippocampal}, the two learning speeds \citep{mcclelland1995why}, replay, time cells \citep{eichenbaum2014time} and the cognitive map \citep{tolman1948cognitive,okeefe1978hippocampus}, and it adds a condition these models leave open: the controller reads coincidences and never the code.

\textbf{Computation and expressivity.} Computation in real time on multitape machines \citep{yamada1962real,rabin1963real,hartmanis1965computational,rosenberg1967real} has a strict hierarchy in the number of tapes \citep{aanderaa1974k,paul1981information,paul1982online,paturi1990milking}, and tapes with several heads, with or without jumps between heads, add no power in real time \citep{fischer1968limited,fischer1972real,kosaraju1979real,savitch1984power}; Theorem~\ref{thm:tapes} places the machine of Eq.~\eqref{eq:machine} in this landscape. Universality results for recurrent networks and Transformers keep the unbounded part of a computation in a precision: recurrent networks are universal with unbounded precision \citep{siegelmann1995computational}, and with bounded precision when they grow stacks of neurons read only at the top \citep{chung2021turing}, and Transformers are universal with positional encodings of unbounded precision \citep{perez2021attention}. Constructions with bounded precision keep the unbounded part in a memory addressed by position, as the neural Turing machine does \citep{graves2014neural} and as a looped Transformer does that emulates a computer with binary addresses of fixed width \citep{giannou2023looped}. Chain of thought and looped computation buy time to think, and extend what Transformers compute with the number of steps \citep{merrill2024expressive,feng2023towards,li2024chain,giannou2023looped,fan2025looped}. Section~\ref{sec:computation} keeps the unbounded part in clocks the controller never reads, and the multiplier of Section~\ref{sec:exp_tm} thinks in the same sense. Recurrent networks recognize counter languages with saturated activations, a count held in a state that is a coordinate \citep{weiss2018practical}; self attention of bounded size cannot track parity or nesting at every length \citep{hahn2020theoretical} and lies in small circuit classes \citep{merrill2023parallelism}; Transformers that simulate automata learn shortcuts that fail at new lengths \citep{liu2023transformers}; and state space models share these limits \citep{merrill2024illusion}.

\textbf{Learnability beyond the training support.} Identification in the limit fails for every superfinite class of languages from positive examples \citep{gold1967language}; regular languages are learned from queries and counterexamples \citep{angluin1987learning}, and deterministic finite state transducers are identified from input and output pairs once the sample contains a characteristic set \citep{oncina1993learning}, the classical relative of Theorem~\ref{thm:hole}(ii). For Transformers whose positional information is periodic or local, queries and keys fall into a finite partition independent of the length, and length generalization follows in the limit; copying and addition cannot be written this way \citep{huang2025formal}. The machine of Eq.~\eqref{eq:machine} writes them because a head may test two clock readings for equality, a relation invariant under translation and not local, valid within the range of the clocks. Length generalization also holds when each output depends on a bounded number of earlier tokens \citep{golowich2025role}. On inputs outside the support, networks trained by gradient descent follow the bias of their architecture: networks of rectified linear units become linear along every direction \citep{xu2021how}, and several models learn the interpolator of least degree \citep{abbe2023generalization}. Theorem~\ref{thm:hole} asks which rules a finite set of inputs fixes on all longer inputs, and answers it by what the learner reads.

\textbf{Automata in neural dynamics.} Coupled recurrent networks with soft competition implement finite state machines \citep{rutishauser2009state,neftci2013synthesizing}, attractor networks with mixed selectivity hold the states of tasks governed by rules \citep{rigotti2010internal}, and constructions with symbolic vectors place finite state machines in Hopfield networks \citep{cotteret2024vector,hopfield1982neural}. Turing machines have been compiled into neural fields and piecewise linear recurrent networks through G\"odel encodings \citep{beimgraben2014universal,carmantini2017modular}, where the tape is again stored in the precision of a state; large models build symbol processing from vector binding \citep{smolensky1990tensor,eliasmith2012large}. Accounts of hippocampal memory by keys and values or by content \citep{sharma2022content,gershman2025key,chandra2025episodic} treat the entorhinal cortex and hippocampus as a system of memory addresses, and the differentiable neural computer reads an external memory by content and by the order of writing \citep{graves2016hybrid}. In the present machine the tape is the orbit of a bump on rings of a continuous attractor network, and the head is the bump.

\textbf{Continuous attractor networks.} Bumps of activity in networks with translation invariant connections \citep{amari1977dynamics,benyishai1995theory,zhang1996representation} form a continuous family whose position integrates velocity \citep{zhang1996representation,burak2009accurate}; their dynamics reduce to the motion of the bump along the neutral direction \citep{wu2008dynamics,fung2010moving,wu2016continuous}. Firing rate adaptation lets the bump travel above $m=\tau/\tau_v$ and shortens its lag behind a moving input, which above that value becomes a lead for slow inputs \citep{mi2014spike,li2025dynamics}; noisy adaptation gives L\'evy flights \citep{dong2021noisy}, oscillatory tracking gives theta sweeps and phase precession \citep{chu2024firing,ji2025systems}, and the strength of adaptation sets whether replay is stationary, diffusive or superdiffusive \citep{ji2026dynamical}. Theorem~\ref{thm:neural} uses these results with the symmetry of the network reduced to the group of the clock by weak wells.

\textbf{Symmetry in learning.} Geometric deep learning builds the symmetries of the input into a network \citep{bronstein2021geometric,cohen2016group}. Here the symmetry belongs to the learner's own memory, and its invariants are what the learner may read.

\textbf{Recursion, rules and symbols.} Accounts of recursion as a dedicated operation or a stack \citep{hauser2002faculty}, as serial programs run by a global workspace \citep{zylberberg2011human}, as one of several mechanisms of sequence coding \citep{dehaene2015neural}, or as compression in languages of thought \citep{dehaene2022symbols,amalric2017language} describe what recursive procedures do; here recursion is a way of moving through a group. Fodor and Pylyshyn argued that systematicity and productivity require a classical architecture at the cognitive level \citep{fodor1988connectionism}; Smolensky answered that symbols describe a network's dynamics only approximately \citep{smolensky1988proper}, and van Gelder that cognition is dynamics and computation a misdescription of it \citep{vangelder1998dynamical}. Multilayer perceptrons trained by backpropagation do not extend the identity function and similar mappings beyond their training space \citep{marcus2001algebraic}, and recurrent networks fail when generalization requires systematic composition \citep{lake2018generalization}. Section~\ref{sec:discussion} gives the case in which symbols and dynamics describe one system exactly: the symbols are the invariants of a symmetry of the dynamics. Wittgenstein's and Kripke's problem of following a rule \citep{wittgenstein1953philosophical,kripke1982wittgenstein} and Goodman's new riddle of induction \citep{goodman1955fact} ask what fixes a rule; the hole argument and the point coincidence argument of general relativity \citep{einstein1916foundation,earman1987what,stachel1989einstein,norton1993general} supply the form of the answer.

%% file: app_code.tex
\section{Code and Data}
\label{app:code}
This appendix lists the code, results and trained machines that accompany the paper and that Section~\ref{sec:experiments} refers to, one line per module, with the section or appendix each serves. Every number in the text and in the tables is written by \texttt{code/make\_numbers.py} from \texttt{code/results/}, and every figure by a script in \texttt{code/figures/}, which also writes the numbers drawn in each panel to \texttt{paper/figs/source\_data/}. Every figure and table can be redrawn from the results without running an experiment.

\begin{center}
\small
\begin{tabular}{@{}p{0.34\textwidth}p{0.60\textwidth}@{}}
\toprule
\texttt{orrery/model.py} & the machine of Eq.~\eqref{eq:machine} and its relaxation; Leibniz's shift of every reading or of the step index\\
\texttt{orrery/fast.py} & the discrete machine at test; exchanged, shortened and unbounded clocks; the catalogue removed\\
\texttt{orrery/train.py} & the training protocol and the restart ladder\\
\texttt{orrery/reference.py} & the reference controllers of Appendix~\ref{app:reference}\\
\texttt{orrery/tasks.py}, \texttt{evaluate.py} & the fifteen tasks and the benchmark score\\
\texttt{orrery/baselines.py} & LSTM and Transformer baselines\\
\texttt{orrery/neural.py} & the network of Eq.~\eqref{eq:cann}, the maps $\Pi$, $Q$ and the Chinese remainder theorem, place cells and the neural executor (Appendix~\ref{app:neural})\\
\texttt{orrery/tm.py} & Turing machines compiled into controllers (Theorem~\ref{thm:tapes}); random machines and the schoolbook multiplier\\
\texttt{experiments/e0\_reference.py} & reference controllers: exactness, situations and $N^\ast$\\
\texttt{experiments/e1\_benchmark.py} & the benchmark, for the machine and the baselines (Section~\ref{sec:exp_bench})\\
\texttt{experiments/e2\_theory.py} & clock exchange and cycles of a million readings, cycle removal, forbidden readings and Leibniz's test, short experience and binary addition bit by bit, the catalogue removed, and number of clocks (Sections~\ref{sec:exp_range} to~\ref{sec:exp_teach})\\
\texttt{experiments/e3\_dynamics.py} & clock movements and reads of trained machines (Section~\ref{sec:exp_dynamics})\\
\texttt{experiments/e4\_neural.py} & the machines in neural dynamics: clocks, adaptation, trained controllers, replay (Section~\ref{sec:exp_neural})\\
\texttt{experiments/e5\_computation.py} & compiled Turing machines, time to think, and the number of clocks (Sections~\ref{sec:exp_teach} and~\ref{sec:exp_tm})\\
\texttt{make\_numbers.py} & results to \texttt{paper/numbers.tex} and \texttt{paper/results\_tables.tex}\\
\texttt{figures/} & one script per figure, with the source data of its panels\\
\texttt{tests/} & exactness of the executor and a run of the whole pipeline at small size\\
\bottomrule
\end{tabular}
\end{center}

The constants of training are set in \texttt{orrery/train.py} and are the same for every task (Appendix~\ref{app:methods}). A single job runs with \texttt{python experiments/e2\_theory.py --job K}, and \texttt{experiments/parallel.py} runs all jobs of a script on several devices. A benchmark run of the machine took a median of \res{E1.compute.median} hours with its restarts (Appendix~\ref{app:compute}); the exact executor runs on a CPU.